\documentclass[10pt,twocolumn,letterpaper]{article}

\usepackage[pagenumbers]{cvpr} 

\usepackage[dvipsnames]{xcolor}

\definecolor{cvprblue}{rgb}{0.21,0.49,0.74}
\usepackage[pagebackref,breaklinks,colorlinks,citecolor=cvprblue]{hyperref}

\usepackage[accsupp]{axessibility} 

\def\paperID{4334} 
\def\confName{CVPR}
\def\confYear{2024}

\title{Dr.Hair: Reconstructing Scalp-Connected Hair Strands without Pre-Training\\via Differentiable Rendering of Line Segments}
\author{
Yusuke Takimoto$^{1*}$~~~~~~
Hikari Takehara$^{1*}$~~~~~~
Hiroyuki Sato$^{1*}$~~~~~~
Zihao Zhu$^{1,2\dagger}$~~~~~~
Bo Zheng$^{1}$
\vspace{0.2cm} \\
$^{1}$Huawei Technologies Japan K.K.~~~~~~~~~~~~~~~~~~~~~~~~
$^{2}$Keio University~~~~~~~~~~~~~~~~~~~~
}

\usepackage{mathrsfs}

\usepackage {mwe}

\usepackage{mathbbol}
\usepackage{amssymb}             
\DeclareSymbolFontAlphabet{\amsmathbb}{AMSb}%

\usepackage{tikz}

\usepackage[export]{adjustbox}
\usepackage{float}
\usepackage{wrapfig}

\usepackage{tabularx}
\usepackage{array}

\newcommand{\dataroot}{./data/}

\begin{document}

\newcommand\extralabel[2]{{\edef\@currentlabel{\@currentlabel#2}\label{#1}}}
\newlength{\mywidth}
\newlength{\myheight}

\newcommand{\CenterRow}[2]{
  \dimen0=\ht\strutbox%
  \advance\dimen0\dp\strutbox%
  \multiply\dimen0 by#1%
  \divide\dimen0 by2%
  \advance\dimen0 by-.5\normalbaselineskip%
  \raisebox{-\dimen0}[0pt][0pt]{#2}}
\twocolumn[{%
\renewcommand\twocolumn[1][]{#1}%
\maketitle
\begin{center}
    \centering
    \captionsetup{type=figure}
    \begin{tabular}{ccccc}
        \includegraphics[width=.172\textwidth]{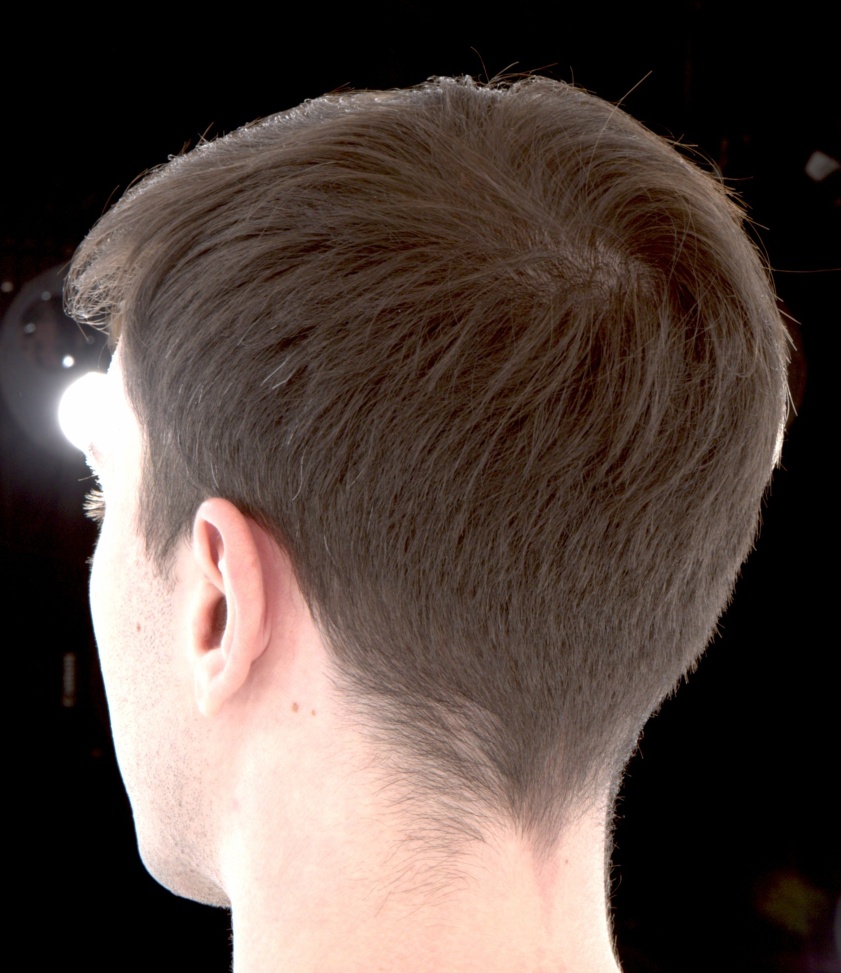} &
        \includegraphics[width=.172\textwidth]{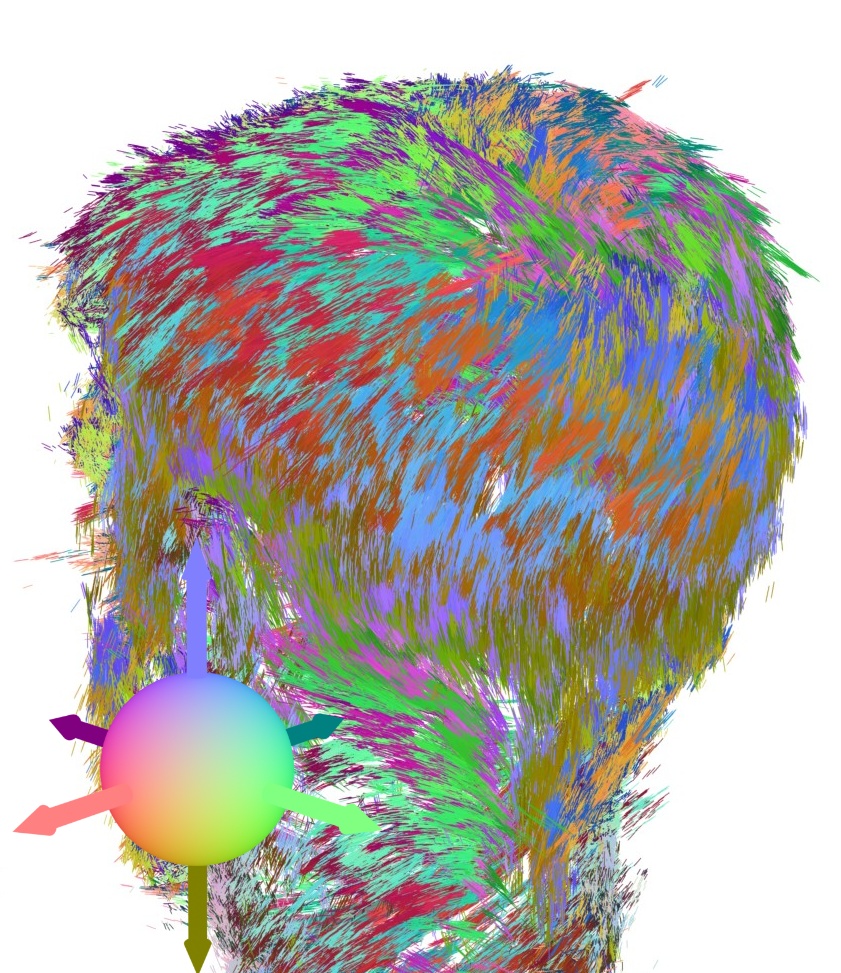} &
        \includegraphics[width=.172\textwidth]{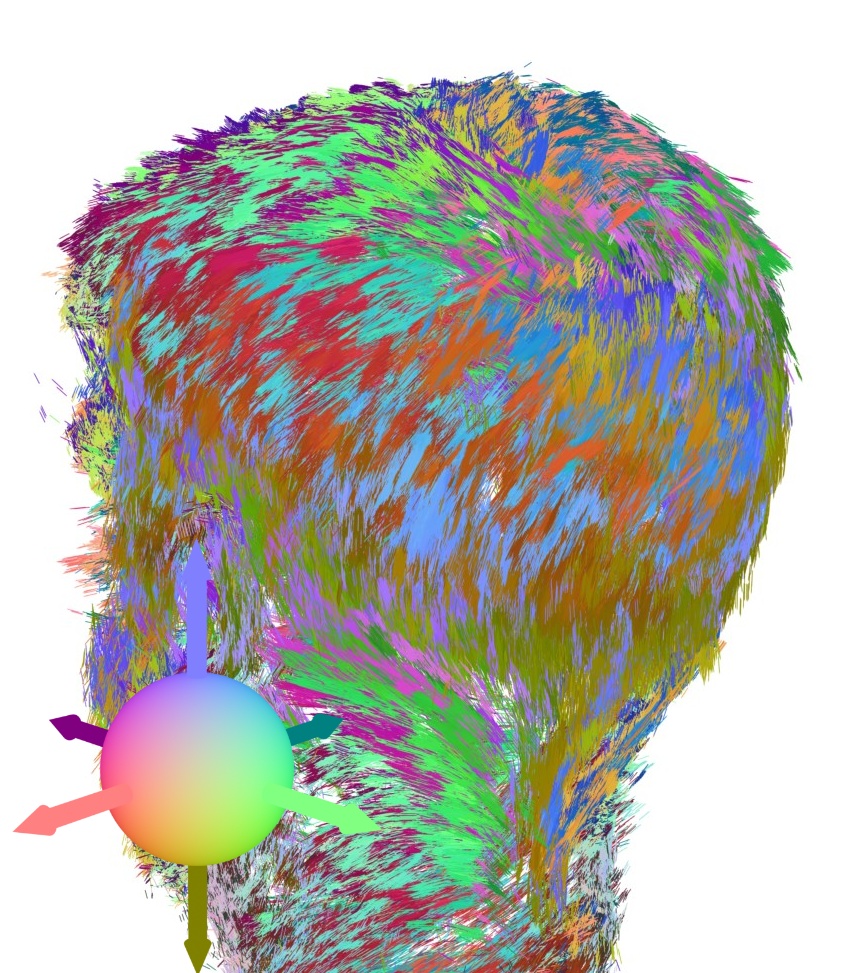} &
        \includegraphics[width=.172\textwidth]{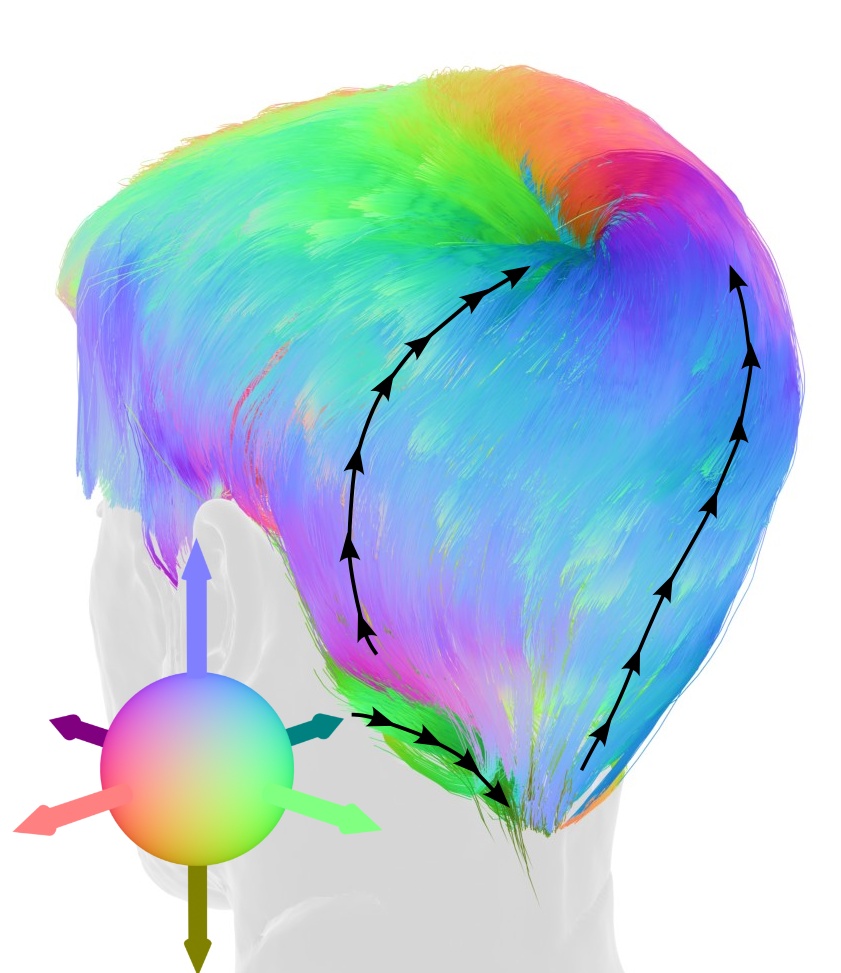} &
        \includegraphics[width=.172\textwidth]{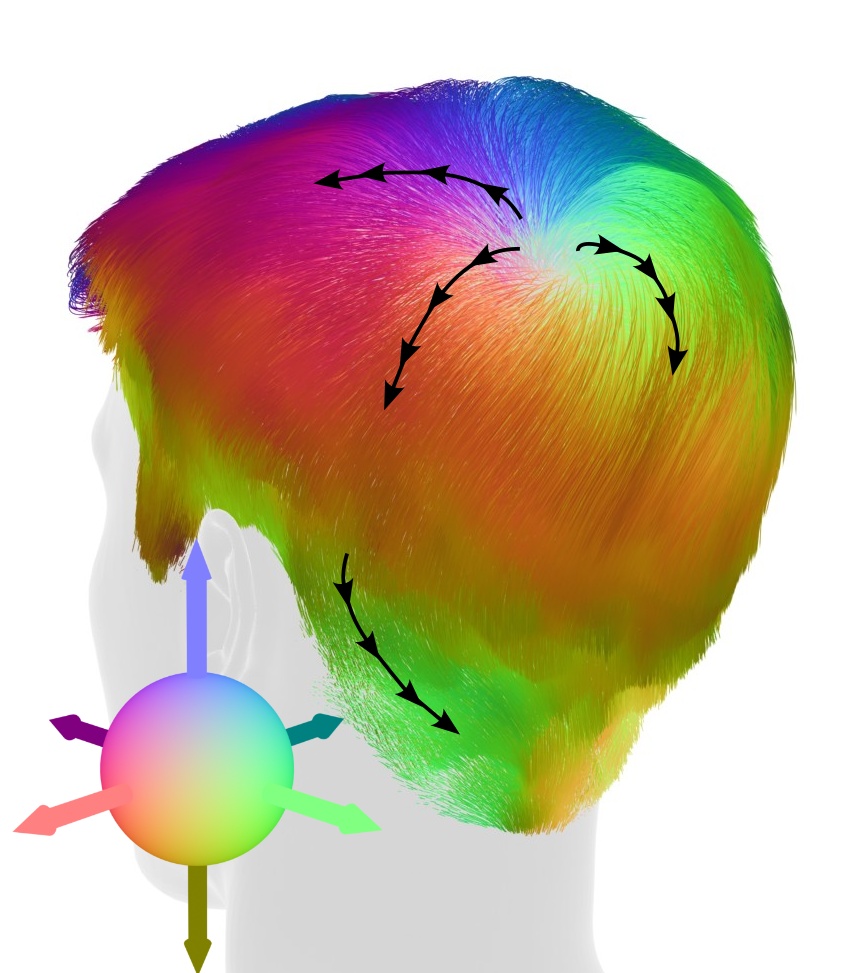} \\
        \includegraphics[width=.172\textwidth]{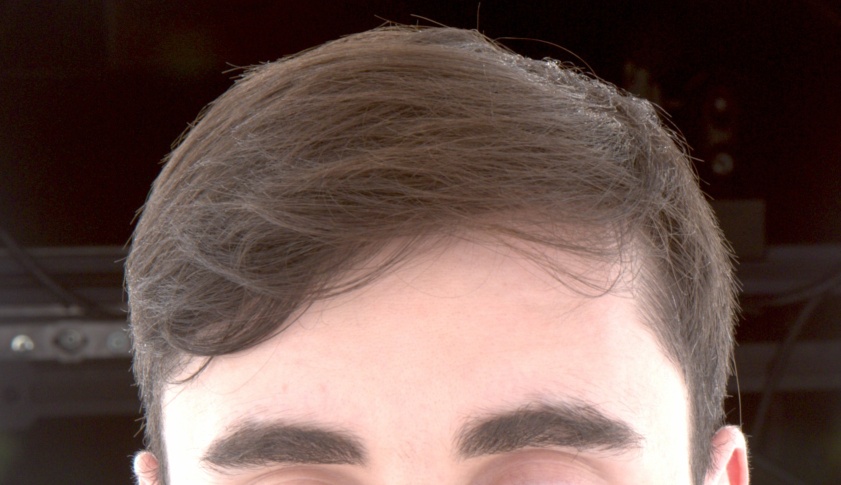} &
        \includegraphics[width=.172\textwidth]{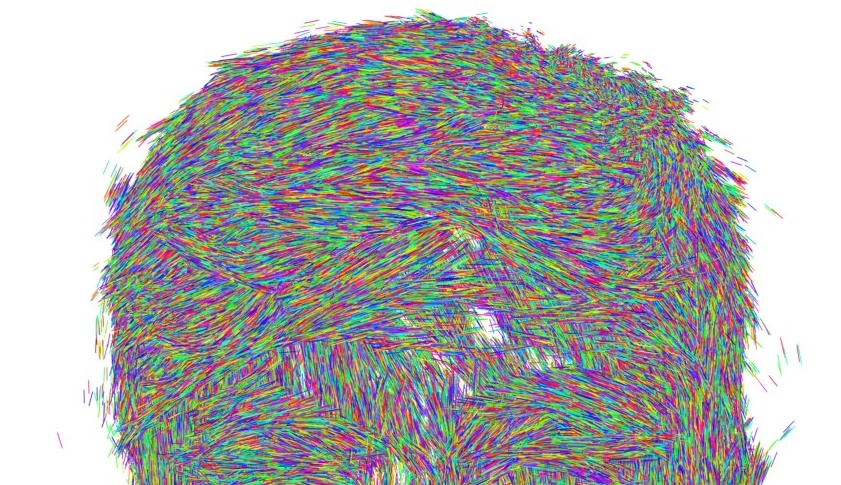} &
        \includegraphics[width=.172\textwidth]{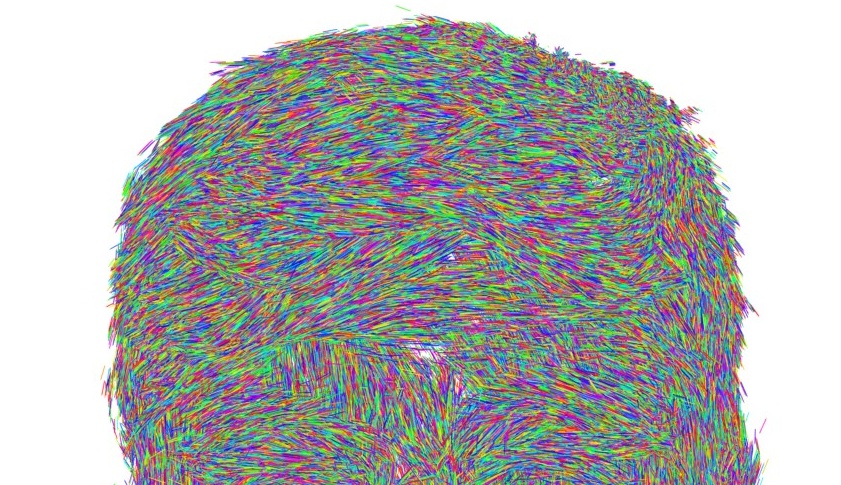} &
        \includegraphics[width=.172\textwidth]{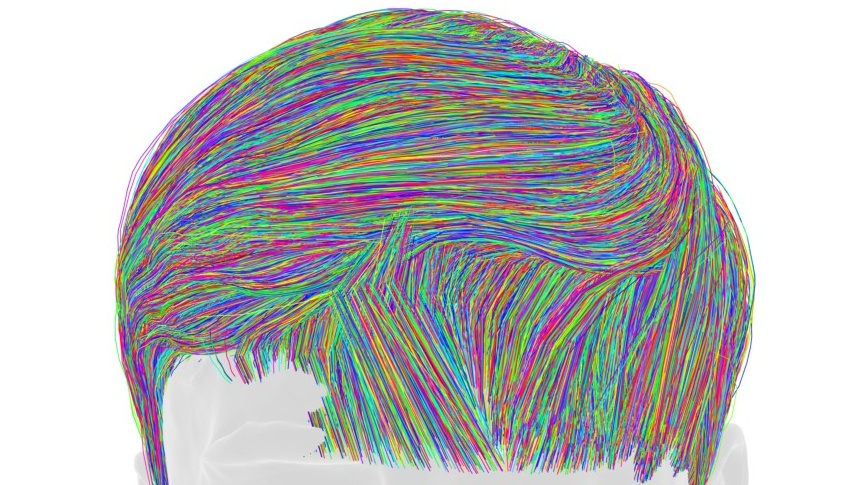} &
        \includegraphics[width=.172\textwidth]{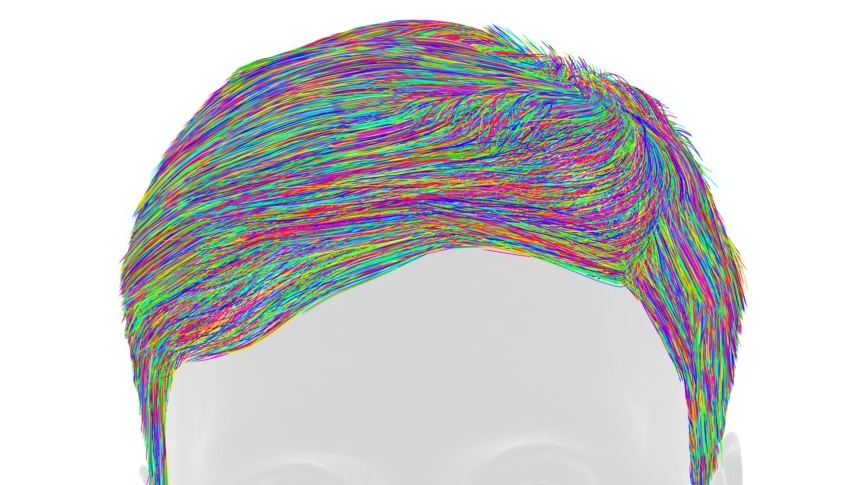} \\
        \textbf{Image}  & \textbf{LPMVS} \cite{Nam_2019_CVPR} & \textbf{Strand Integration} \cite{maeda2023refinement} & \textbf{NeuralHaircut} \cite{Sklyarova_2023_ICCV} & \textbf{Ours}\\[6pt]
    \end{tabular}
    \captionof{figure}{Results of existing strand-based 3D reconstruction methods and our method tested with the data captured by a multi-camera system.
                       In the upper row, color and colored arrows represent 3D orientation of hair strands.
                       The overlaid black arrows were drawn manually to visualize rough orientations.
                       The lower row shows individual strands with random color.
                       LPMVS and Strand Integration failed to estimate consistent direction, and their strands are too short not to connect to the scalp.
                       The absolute orientation of strands estimated by NeuralHaircut is mostly the opposite of the actual hair orientation.
                       Our method demonstrates better precision in reconstructing the directional flow of scalp-connected hair.}
    \label{fig:exp_lc}
\end{center}%
}]
\def\thefootnote{*}\footnotetext{Core authors}
\def\thefootnote{$\dagger$}\footnotetext{Work done during an internship at Huawei Technologies Japan K.K.}

\def\thefootnote{\arabic{footnote}}

\begin{abstract}
In the film and gaming industries, achieving a realistic hair appearance typically involves the use of strands originating from the scalp.
However, reconstructing these strands from observed surface images of hair presents significant challenges.
The difficulty in acquiring Ground Truth (GT) data has led state-of-the-art learning-based methods to rely on pre-training with manually prepared synthetic CG data.
This process is not only labor-intensive and costly but also introduces complications due to the domain gap when compared to real-world data.
In this study, we propose an optimization-based approach that eliminates the need for pre-training.
Our method represents hair strands as line segments growing from the scalp and optimizes them using a novel differentiable rendering algorithm.
To robustly optimize a substantial number of slender explicit geometries, we introduce 3D orientation estimation utilizing global optimization, strand initialization based on Laplace's equation, and reparameterization that leverages geometric connectivity and spatial proximity.
Unlike existing optimization-based methods, our method is capable of reconstructing internal hair flow in an absolute direction.
Our method exhibits robust and accurate inverse rendering, surpassing the quality of existing methods and significantly improving processing speed.
\end{abstract}

\section{Introduction}
\label{sec:intro}
High-quality 3D hair data is essential for depicting realistic human figures in movies, games, and metaverse.
However, capturing real hair is notoriously difficult due to its intricate properties, including its elongated shape, overlapping strands, transparency, reflectivity, and uniformity, which challenge even the most advanced computer vision stereo techniques at a sub-pixel resolution.
As a result, hair processing is one of the most formidable tasks in image-based 3D human modeling.

Hair-specific reconstructions methods have been studied for many years \cite{paris2004capture, wei_2005_multiple, paris_2009_hair,jakov_2009_fiber, Nam_2019_CVPR, Bousseau2021Photometric, maeda2023refinement}.
These approaches are characterized by applying a Gabor filter to the hair image to calculate the 2D orientation, which is then combined with 3D measurements for optimization.
However, these methods require accurate calibration of the illumination and cameras, making it difficult to scale them to casual shooting environments.
In addition, since only the hair's surface can be measured, it is not easy to estimate the form of the hair connected to the scalp, which is commonly used in industry.

Data-driven reconstructions, particularly those using volumetric representations via neural networks, have recently gained traction in 3D hair modeling, and many studies on humans with hair \cite{lombardi2021mixture,cao_2022_authentic} are conducted.
Among them, hair-specific methods using one or a few views \cite{chai2016autohair, hu_2017_avator,zhou2018hairnet,saito_2018_3d,zhang2019hair,Wu_2022_CVPR,Zheng_2023_CVPR, zhang_2017_fourview,DeepMVSHair_2022} have been actively studied.
The recently proposed methods \cite{rosu2022neuralstrands, Sklyarova_2023_ICCV} use pre-trained priors and perform strand fitting through differentiable rendering \cite{dss2019yifan, adop2022ruckert, liu2019softras, ravi2020pytorch3d} against the multi-view images at runtime.
However, creating the CG data used for pre-training is not only costly and requires manual work by artists but also has the problem of domain gaps.

In response, we propose Dr.Hair, an optimization-based pipeline that recovers individual strands connected to the scalp from multi-view images to address the above problems.
We start with conventional hair representation used in real-time rendering and standard CG tools.
After fitting a scalp to the raw hair mesh, we compute consistent 3D orientations from 2D orientation images.
From the results, the guide strands are initialized based on a differential equation.
Finally, a hierarchical relationship called {\it{guide-child}} is utilized for optimization based on differentiable rendering.
To summarize, our contributions are:
\begin{itemize}
\item 3D orientation estimation using global optimization, estimating consistent surface orientation;
\item Laplace's equation-based strand initialization, filling interior hair flow smoothly from surface observations;
\item Rasterization-based differentiable rendering algorithm for line segments, generating smooth gradient in image space while maintaining high-frequency detail;
\item Reparameterization of strand shapes, propagating dense gradients throughout the geometry;
\item An optimization framework using hierarchical relations of guide and child hair.
\end{itemize}
Finally, we validate the effectiveness of our method on synthetic and real data.
\begin{figure*}[t]\centering
\includegraphics[width=\textwidth]{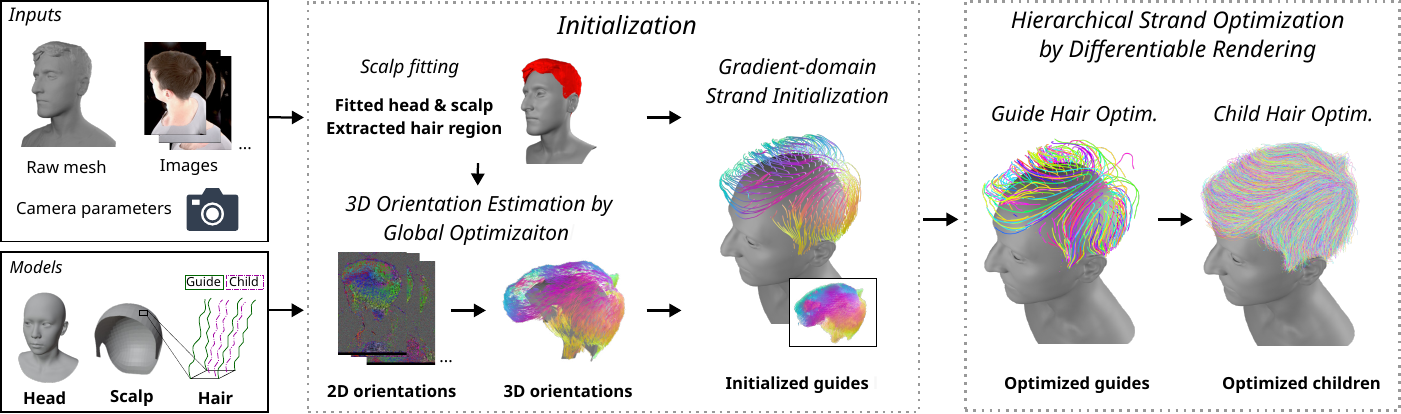}
\caption{The overview of our pipeline.
Our approach combines traditional real-time rendering techniques with recent advances in differentiable rendering.
First, we fit a template to a raw mesh. Next, we compute consistent 3D orientations from 2D orientation images and initialize guide strands based on a differential equation.
Finally, optimization based on differentiable rendering is applied by leveraging the hierarchical relationship between guides and children.
}
\label{fig:pipeline}
\end{figure*}

\section{Related Work}
\label{sec:relatedwork}

\subsection{Optimization-based methods}
Since the early 2000s, to measure the strands on hair surface, optimization methods combining 3D geometry with 2D orientation extracted from multiple images have been explored \cite{paris2004capture, wei_2005_multiple, paris_2009_hair, jakov_2009_fiber}.
These methods utilized Gabor filters to address the high specularity arising from the cylindrical shape and semi-transparent material and have been extended to dynamic scenarios  \cite{luo2011dynamic, luo2012multi}.
Subsequently, Line-based PatchMatch MVS (LPMVS) \cite{Nam_2019_CVPR} have enabled more accurate strand geometry acquisition.
Strand Integration \cite{maeda2023refinement} further refines the strands of LPMVS.
Joint measurement of material properties has been achieved \cite{Bousseau2021Photometric}.
However, these methods require controlled lighting conditions and struggle with the 180\textdegree~ambiguity of 2D/3D orientation.

Some methods have addressed this ambiguity issue using user stroke input \cite{chai_2013_dynamic}, Markov Random Field optimization \cite{luo_2013_structure}, and pre-trained models \cite{zhang_2018_modeling} to estimate long strands connected to the scalp.
Specialized techniques for braided hair \cite{hu_2014_braided} and for estimating simulation parameters \cite{10.1111:cgf.13126} also exist.
Although handling various hairstyles is challenging for these approaches, our global optimization robustly estimates consistent 3D orientations.
To extrapolate internal hair flow from surface observations, second order differential equations \cite{paris_2009_hair, jakov_2009_fiber} and 3D PatchMatch \cite{zhang_2018_modeling} have been used.
These interpolation techniques are valuable for seamlessly integrating hair strands onto the scalp.

Most existing methods stray from the conventional hair modeling process conducted by human artists employing tools such as XGen \cite{maya}, Ornatrix \cite{ornatrix}, and Blender Hair Curves \cite{blender}.
In these tools, hair manipulation commonly involves a hierarchical arrangement of guide and child strands.
Artists primarily manipulate guides, while children are generated through interpolation.
EnergyHair \cite{zhang2022energyhair} facilitates image-based interactive hair modeling utilizing this hierarchical structure.
Our pipeline brings this guide-child hierarchy into a fully automatic hair strand reconstruction.

Differentiable rendering (DR) has been attracting attention as a method for reconstructing 3D scenes.
Not only implicit functions \cite{Lombardi:2019, mildenhall2020nerf, wang2021neus}, but also methods for explicit geometric primitives such as meshes \cite{NimierDavidVicini2019Mitsuba2, Mitsuba3, liu2019softras, ravi2020pytorch3d, Laine2020diffrast, Nicolet2021Large, cole2021differentiable, takimoto_2022_dressi} and point clouds \cite{dss2019yifan, adop2022ruckert} have been studied extensively.
However, DR of line primitives used for hair has not been well studied.
We propose a DR framework for line segments to robustly optimize hair strands.
\subsection{Learning-based methods}
Data-driven hair strand reconstruction has been widely studied.
An early work used simulated examples \cite{hu_2014_robust}, and the field has gained popularity after the release of a synthetic dataset created by hand at significant cost, USC-HairSalon \cite{hu2015database}.

\begin{figure}[t]\centering
\begin{minipage}[t]{0.24\linewidth}
\centering
\includegraphics[width=2.5cm,trim={1cm 3cm 0 0},clip]{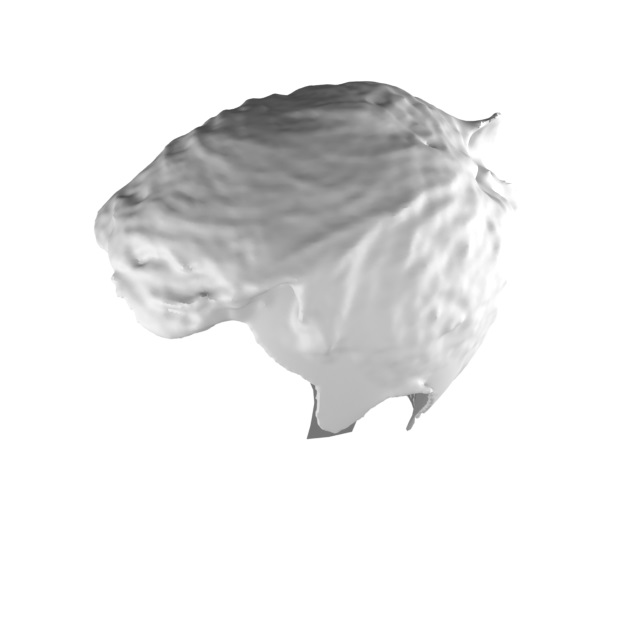}
\subcaption{Raw mesh}
\end{minipage}
\begin{minipage}[t]{0.24\linewidth}
\centering
\includegraphics[width=2.5cm,trim={1cm 3cm 0 0},clip]{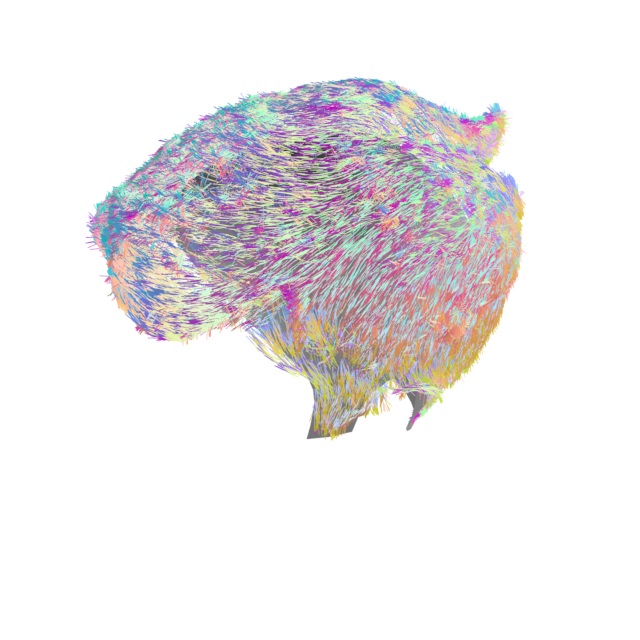}
\subcaption{Before optim.}
\end{minipage}
\begin{minipage}[t]{0.24\linewidth}
\centering
\includegraphics[width=2.5cm,trim={1cm 3cm 0 0},clip]{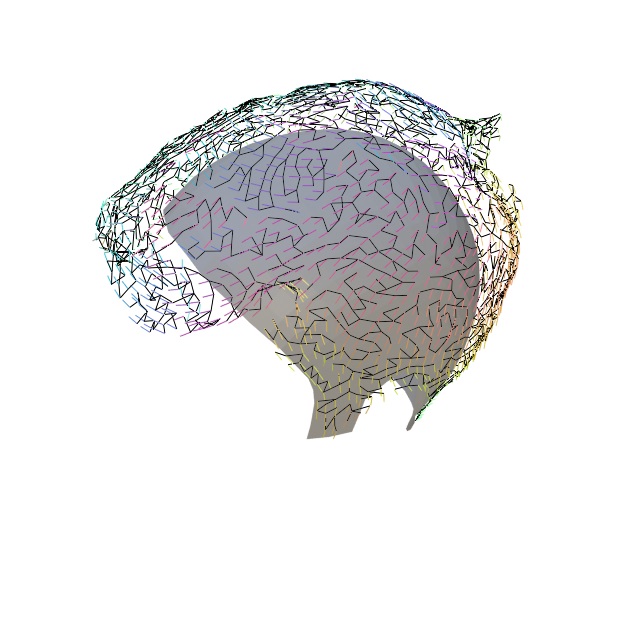}
\subcaption{MST}
\end{minipage}
\begin{minipage}[t]{0.24\linewidth}
\centering
\includegraphics[width=2.5cm,trim={1cm 3cm 0 0},clip]{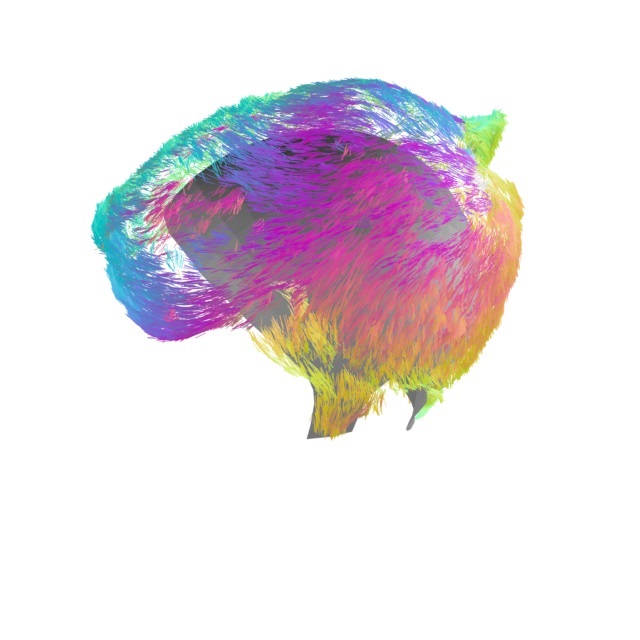}
\subcaption{After optim.}
\end{minipage}
\caption{Surface 3D orientation estimation by global optimization:
Hair regions are visualized on a gray scalp.
(a) A raw mesh.
(b) Surface 3D orientations before our optimization. The color stands for 3D orientation. Because of 180\textdegree~ambiguity, orientations flip intermittently.
(c) Minimum Spanning Tree (MST). The nodes are downsampled points, which are visualized as colored lines after propagation. Black lines represent edges.
(d) Surface 3D orientations after our optimization. Consistent hair flow from the whorl to the ends is established.
}
\label{fig:3d_orien}
\end{figure}

Many methods attempt to overcome the ambiguity of thin hair strands through volumetric representation, particularly in simple setups like single-view \cite{chai2016autohair, hu_2017_avator, zhou2018hairnet, saito_2018_3d, zhang2019hair, Wu_2022_CVPR, Zheng_2023_CVPR}, sparse views \cite{zhang_2017_fourview, DeepMVSHair_2022}, and sketches \cite{shen2020deepsketchhair}.
However, due to the small number of views, their 3D consistency is limited.
Fine-tuning to deform strands against views \cite{hu2015database, DeepMVSHair_2022} is utilized to enhance high-frequency details.
Neural-based volumetric representations \cite{mildenhall2020nerf} have been applied to head and hair reconstruction \cite{Lombardi:2019, lombardi2021mixture, cao_2022_authentic}.
CT2Hair \cite{shen2023CT2Hair} reconstructs high-quality strands from wig CT data.
Dynamic scenes are handled as well \cite{yang_2019_dynamic, Wang_2022_CVPR, wang2023neuwigs}.
Generative models for hair strands have been recently proposed \cite{10.1145/3618309, sklyarova2023haar}.

In a multi-view setup, some methods train priors such as strand generators and perform the geometry texture \cite{wang_2009_example} optimization against input images.
NeuralStrands \cite{rosu2022neuralstrands} uses surface 3D orientation \cite{Nam_2019_CVPR} as a constraint and rasterizes strands generated from neural shape texture using point-based DR \cite{dss2019yifan, adop2022ruckert}.
It achieves photorealistic hair appearance using neural appearance texture and a UNet-based neural renderer.
However, manual annotation is needed to resolve the 180\textdegree~orientation ambiguity.
NeuralHaircut \cite{Sklyarova_2023_ICCV} uses volumetric reconstruction \cite{wang2021neus} as the first stage.
Then, regularizing the geometry texture with a pre-trained diffusion model, broad gradient propagation by mesh-based soft rasterization \cite{liu2019softras, ravi2020pytorch3d} is performed for sparse strands.
It finally generates realistic images using UNet.
These methods offer excellent rendering quality, but strand geometry accuracy is limited by blurred images of soft rasterization and domain gaps.
Moreover, both pre-training and optimization are time-consuming.
Recently proposed GaussianHair \cite{luo2024gaussianhair} utilizes gaussian splatting \cite{kerbl3Dgaussians} along with a pre-trained strand decoder.

\section{Method}
\label{sec:method}
\figurename~\ref{fig:pipeline} illustrates the overview of the proposed method.

\subsection{Initialization}
In this step, the scalp and the hair strands connected to it are initialized.
We use a head mesh and a separate scalp mesh with a different topology as templates as shown in \figurename~\ref{fig:pipeline}.
A separate scalp with uniform vertex distribution and clear sideburn shapes is convenient for hair growth.
The scalp mesh has a correspondence with the scalp region of the head mesh, and their vertex positions can be mutually transformed by linear interpolation.

The input comprises multi-view images, camera parameters, and a raw mesh.
As a preprocessing step, the head is automatically fitted to the face region of the raw mesh by non-rigid ICP utilizing landmarks and segmentation.
The scalp is then optimized to lay inside the hair region of the raw mesh, and the hair region is extracted.
Details of this scalp fitting are provided in the supplementary material.

\subsubsection*{3D Orientation Estimation by Global Optimization}
First, using Gabor filters \cite{paris2004capture}, the 2D orientation and confidence are calculated from the color images of each view.
Next, 3D-oriented points are reconstructed by using a modified version of LPMVS \cite{Nam_2019_CVPR}.
For fast computation, depth values were fixed by rendering the raw mesh from each viewpoint, and only the 3D orientation was optimized.
Then, to reduce noise, mean-shift is applied to the points by following \cite{Nam_2019_CVPR}.
The resulting orientation contains a 180\textdegree~ambiguity in the Euclidean space, so global consistency from the roots to the tips should be sought.

We make a graph structure that connects neighboring points using edge weights determined by the absolute value of the inner product of orientations.
On this graph, an MST \cite{prim1957shortest} is constructed.
The orientation is then sequentially propagated from the initial point until all points are reached on the MST; if the inner product of neighboring orientations is negative, the destination is rotated by 180\textdegree.
Since a single run may potentially lead to local optima, adding random perturbations to the edge weights of the graph, an MST creation and propagation are performed 100 times.
We define the score of the graph as the sum of inner products of adjacent points' orientations and choose the best in the trials as the global optimal solution.
At this stage, the orientation is globally consistent but uncertain in absolute terms.
In other words, opposite directions, from the tips to the roots, may be estimated.
Therefore, the heuristic that most hair strands should face the direction of gravity is used.
The above process is applied to downsampled points, and the orientation is reflected back to the original resolution.
Original points that differ much in orientation from the optimized points are removed as noise.
Our global optimization process is shown in \figurename~\ref{fig:3d_orien}.
\subsubsection*{Gradient-Domain Strand Initialization}
\begin{figure}[t]\centering
\begin{minipage}[t]{0.32\linewidth}
\centering
\includegraphics[width=2.8cm,trim={0 3cm 0 0},clip]{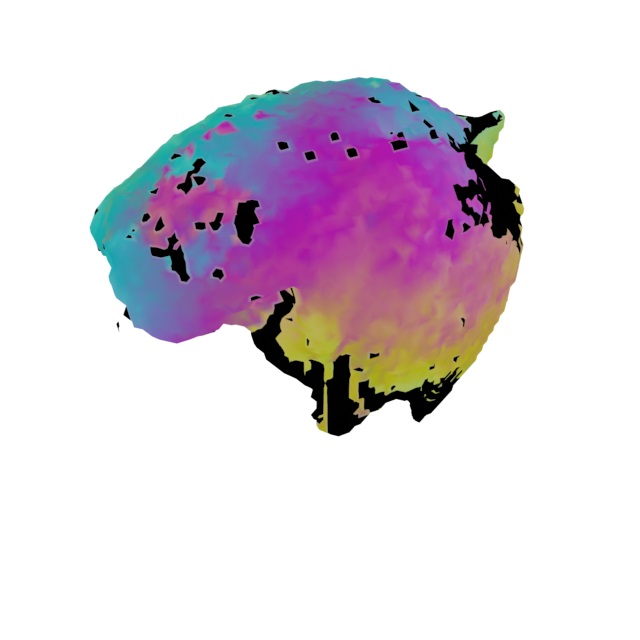}
\subcaption{Boundary cond.}
\end{minipage}
\begin{minipage}[t]{0.32\linewidth}
\centering
\includegraphics[width=2.8cm,trim={0 3cm 0 0},clip]{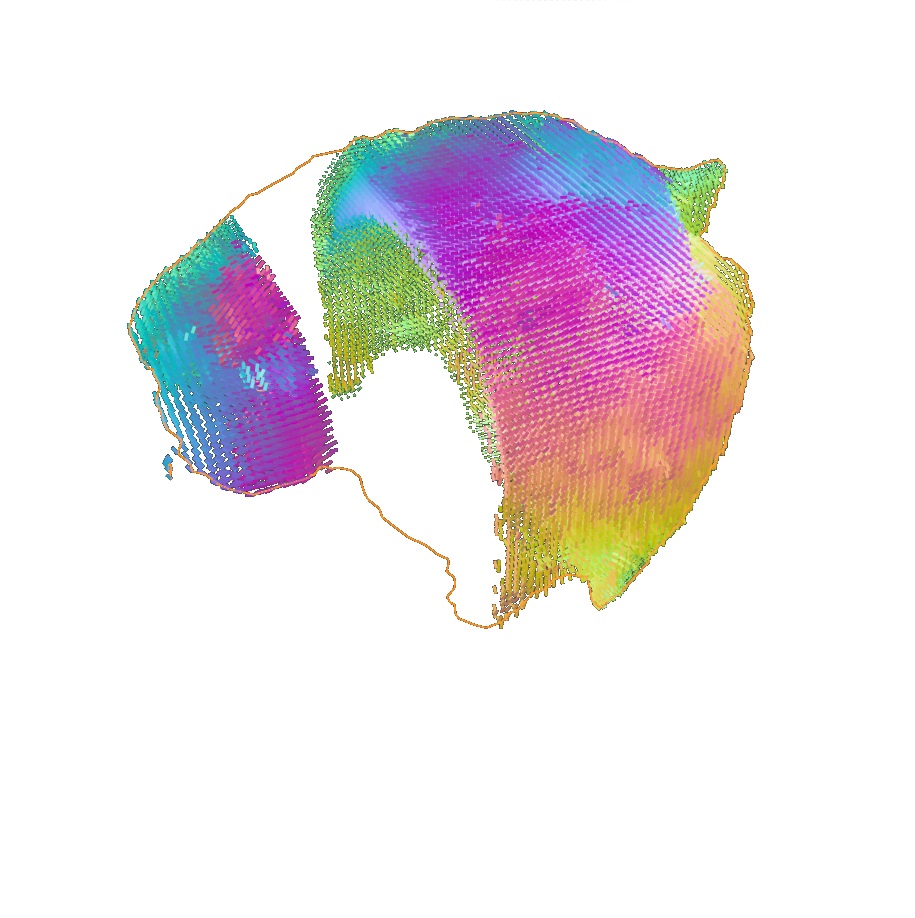}
\subcaption{Optimized volume}
\end{minipage}
\begin{minipage}[t]{0.32\linewidth}
\centering
\includegraphics[width=2.8cm,trim={0 3cm 0 0},clip]{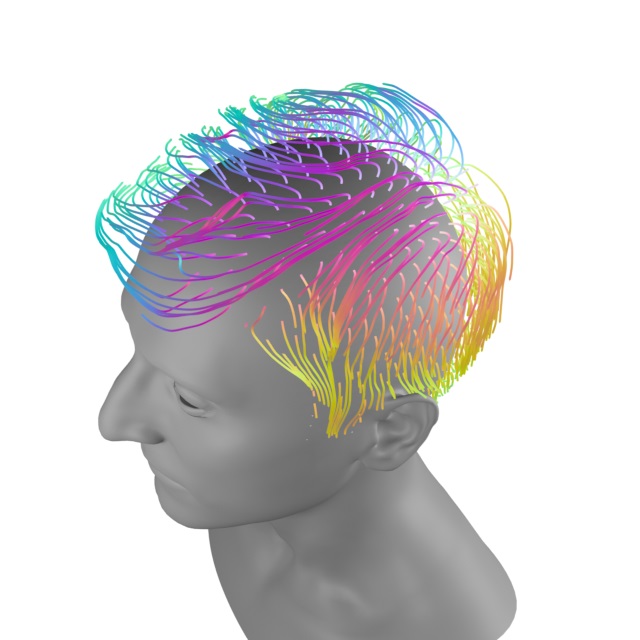}
\subcaption{Extracted strands}
\end{minipage}
\vspace*{-3mm}
\caption{Gradient-domain strand initialization:
(a) Boundary condition. Colored region represents $\mathbb{H}$ and $\mathbb{S}$, and black region represents $\mathbb{U}$.
(b) Optimized volume sliced by certain planes. The Interior is smoothly filled.
(c) Extracted strands from the optimized volume.}
\label{fig:laplace_eq}
\end{figure}

We leverage gradient domain processing \cite{perez2003poisson,kazhdan2006poisson} to estimate spatially smooth internal hair flow through volumetric representaiton \cite{zhang2019hair,DeepMVSHair_2022}.
A visualization of this process is given in \figurename~\ref{fig:laplace_eq}.

Boolean operators extract the space $\Omega$ filled with hair, enclosed by the boundary of hair regions in the raw mesh and the scalp mesh.
We consider filling $\Omega$ with a smooth hair flow field.
$f_o(p)=(n_x, n_y, n_z)$ denotes the orientation at a position $p=(x, y, z) \in \Omega$.
It should satisfy the following properties:
\begin{equation}
\begin{split}
{\nabla}^2 f_o(p) = 0 \quad \text{subject to} \quad \|f_o(p)\|_2 = 1, \\
f_o(p_h) = H(p_h), \; {p_h} \in \mathbb{H}, \; f_o(p_s) = S(p_s), \; {p_s} \in \mathbb{S}
\end{split}
\end{equation}
$f_o(p)$ follows a type of Laplace's equation, a special case of Poisson's equation, whose solution can be determined by boundary conditions.
There are Dirichlet boundary conditions with multiple types: $\mathbb{\Omega}=\{\mathbb{H}, \mathbb{S}, \mathbb{U}\}$.
$\mathbb{H}$ represents the hair surface boundary, and $H(p_h)$ is the estimated 3D orientation of the hair surface obtained in the previous step.
$\mathbb{S}$ and $S(p_s)$ are the scalp boundary and the orientation of the scalp surface, respectively.
Based on the observation that hair on the top of the head grows upwards, but the hair on the back and sides tends to point downwards due to biological characteristics of scalp pores and gravity, we define $S(p_s)$ heuristically with a down vector $d$ as follows:
\begin{equation}
S(p_s) = \text{normalize}(n_s(p_s) + d \min( n_s(p_s) \cdot d + 1, 1))
\end{equation}
$\mathbb{U}$ is an undefined boundary without specific conditions.
For example, it accompanies neck collision and no valid 3D orientation.
We discretize $f_o(p)$ on a regular grid, initialize it by filling the interior space with zeros, and iteratively solve it for each element of XYZ with a successive over-relaxation method.
To avoid instability, the norm constraint is enforced after the iterations.

Finally, we convert the 3D orientation field into guide strands.
Starting from root vertices of $\mathscr{V}$ on $\mathbb S$, we traverse the orientation of the voxels in sequence until reaching $\mathbb H$, generating guide strands.

\subsection{Hierarchical Strand Optimization}
We optimize hair line segments with a novel DR algorithm with reparameterization.
Guide-child hierarchy is incorporated into our optimization framework.
\subsubsection*{Hair Strands as Line Segments}
\begin{figure}[t]\centering
\begin{minipage}[t]{0.6\linewidth}
\centering
\includegraphics[height=2.7cm]{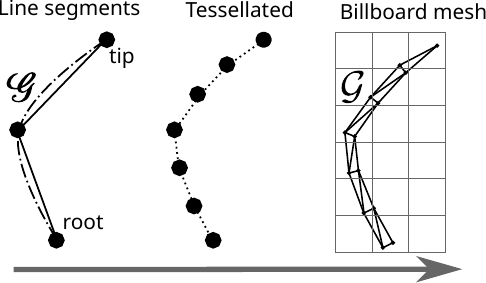}
\caption{Line segments $\mathscr{G}$ are subdivided and converted into billboard mesh $\mathcal{G}$ for rasterization.
The $\mathcal{G}$ thickness is typically less than one pixel.}
\label{fig:line_segments}
\end{minipage}
\hspace{0.04\columnwidth}
\begin{minipage}[t]{0.30\linewidth}
\centering
\includegraphics[height=2.7cm]{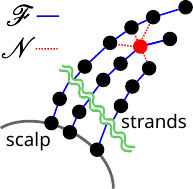}
\caption{Adjacency used in our Laplacian ($k=4$).}
\label{fig:reparameterization}
\end{minipage}
\end{figure}

The representation of hair follows the common practice in real-time rendering \cite{yuksel2010advanced}.
\figurename~\ref{fig:line_segments} displays our hair geometry.
The geometry of hair $\mathscr{G}=\{\mathscr{V}, \mathscr{F}\}$ is a collection of line segments.
Here, $\mathscr V$ refers to the vertex positions corresponding to the division points, and $\mathscr F$ represents the connectivity between upper and lower vertices, expressing spline curves \cite{catmull1974class}.
During rendering, after tessellation, $\mathscr{G}$ is further converted into camera-facing triangle strips $\mathcal{G}=\{\mathcal{V}, \mathcal{F}\}$ and rasterized.
The tip becomes a single triangle.
$\mathcal{V}$ and $\mathcal{F}$ denote the vertex positions and indices, respectively.

A two-stage structure of guide and child hair is employed to express hairstyles.
Guide hair grows from each vertex of the scalp mesh, while child hair grows from sampled positions by Sobol sequence \cite{sobol1967distribution}.
Child hair shape is linearly interpolated with the nearest four guide hairs.
653 guide strands and 50,000 child strands are used in this paper.
For children, the geometry $\mathscr{G}_{c}=\{\mathscr{V}_{c}, \mathscr{F}_{c}\}$ are defined in the same way, and the same rasterization is applied.
\subsubsection*{Differentiable Rendering of Line Segments}
After billboard mesh $\mathcal{G}$ is generated, those triangles are rasterized with the help of hardware, and anti-aliasing (AA) for hair is applied to the rasterized screen-space buffer.

We designate the rasterized color at pixel position $s^{}$ as $c(s^{})$.
$N_8$ and $s_n, n \in N_8 $ denote the 8 neighboring pixels and their positions, respectively.
${c}_{bl}(s^{}, s^{}_n, \mathcal{G})$ represents the blended color.
The function $\text{tri}(s^{})$ returns the triangle ID, while $0 \leq r(s^{}, s^{}_n, \mathcal{G}) \leq 1$ defines a function that returns the distance from pixel $s^{}$ to the edge of the triangle spanning $s^{}_n$.
If multiple edges cross a pixel boundary, the one with the closest depth is chosen.
The screen-space gradient that can move $\mathcal{V}$ is generated via this function that accesses $\mathcal{G}$.
The AA color ${c}_{aa}(s^{}, \mathcal{G})$ is the average of the blended colors.
\vspace{-0.5\baselineskip}
\begin{equation}
{c}_{aa}(s^{}, \mathcal{G}) = (c(s) + \sum_{ n \in N_8 }{{c}_{bl}(s^{}, s^{}_{n}, \mathcal{G})}) / (|N_8|+1)
\end{equation}
\vspace{-1.5\baselineskip}
\begin{equation}
{c}_{bl}(s^{}, s^{}_n, \mathcal{G}) = \\
\begin{cases}
c_{bl'}(s^{}, s^{}_n, \mathcal{G})  & \text{if} \; \text{tri}(s^{}) \neq \text{tri}(s^{}_n) \\
c(s)                       & \text{otherwise}
\end{cases}
\end{equation}
\vspace{-1.5\baselineskip}
\begin{equation}
c_{bl'}(s^{}, s^{}_n, \mathcal{G}) = r(s^{}, s^{}_n, \mathcal{G}) c(s) + (1-r(s^{}, s^{}_n, \mathcal{G})) c(s^{}_n)
\end{equation}
While we have referred to $c_*$ as {\it{colors}} for convenience, this approach can be extended to handle any rasterized vertex attributes, such as silhouette and depth.

Our approach draws inspiration from nvdiffrast \cite{Laine2020diffrast}, a DR for meshes.
The AA of nvdiffrast generates gradients only on the edge pixels where the occlusion actually occurs.
Instead, all pixels with different IDs of adjacent triangles are updated for smoother gradients.
AA can keep finer geometric details than soft rasterization \cite{liu2019softras, ravi2020pytorch3d} used in NeuralHaircut \cite{Sklyarova_2023_ICCV}.
As opposed to splatting \cite{cole2021differentiable}, we leverage the distance between pixels and geometric edges for stronger gradient propagation.
The comparison of AAs is available in the supplementary material.

\subsubsection*{Reparameterization}
Even if AA smoothes the gradients in the screen-space, severe occlusions and non-deterministic rasterization of strands with less than one pixel width result in sparse propagation of gradients into the geometry.
To address this issue, we propose a reparameterization of hair geometries as regularization, which is inspired by a mesh reparameterization \cite{Nicolet2021Large}.
We introduce a Laplacian matrix $\mathbf{L}$ for line segments, considering both geometric connectivity and spatial proximity, to transform the parameter space for optimization.
An example of the Laplacian we propose is shown in \figurename~\ref{fig:reparameterization}.
\begin{equation}
(\mathbf{L})_{ij} = \\
\begin{cases}
-w_{ij}  & \text{if} \; (i,j) \in \{\mathscr{F} \cup \mathscr{N}\}\\
\Sigma_{(i, k)\in \{\mathscr{F} \cup \mathscr{N}\}} w_{ik}     & \text{if} \; i = j\\
0 & \text{otherwise}
\end{cases}
\end{equation}
Here, $\mathscr{N}$ is a set of combinations of neighboring vertices obtained by searching for the $k$-Nearest Neighbors ($k$NN)
based on Euclidean distance for each line segment.
The number of neighbors $k$ allows us to control the effect of spatial proximity relations.
In all experiments presented in this paper, $w_{ij}=1$ is used.
Let $\mathbf x$ be a matrix assembled from $x \in \mathscr{V}$.
Using $\mathbf L$, we reparameterize the vertex positions $\mathbf x$ in Cartesian coordinate into the values $\mathbf u$ in the differential coordinate, where dense gradients are delivered.
$\mathbf{I}$ is the identity matrix, and the parameter $\lambda$ controls regularization effect.
\begin{equation}
\mathbf{u} = (\mathbf{I} + \lambda \mathbf{L})\mathbf{x}
\end{equation}
\subsubsection*{Guide Hair Optimization}
We efficiently perform strand optimization following the Coarse-to-Fine strategy.
First, the vertex positions of guide hair $\mathscr{V}$ are optimized.
Reparameterization with $k=K_{g}$ is applied for guide hair.
We rasterize the strands at a thickness close to the actual hair, 0.2 mm.
$L_{g}$ is minimized using the Adam optimizer \cite{kingma2014adam} in $I_{g}$ iterations.
\begin{equation}
R_{b} = w_{stick} * R_{stick} + w_{root} * R_{root} + w_{c} * R_{c}
\end{equation}
\vspace{-1.5\baselineskip}
\begin{equation}
L_{g} =  w_{d} * L_{d} + w_{m} * L_{m} + w_{t} * L_{t} + R_{b}
\end{equation}
Here, $R_{b}$ constitutes a base regularizer, and it consists of three parts.
$R_{stick}$ is an L1 regularizer to prevent the hair from penetrating the scalp, which is computed from the depth of the penetrated strands and that of the scalp.
$R_{root}$ is an L1 regularizer to ensure that the roots of the hair keep their initial positions as possible.
$R_{c}$ is a curvature regularizer for guide hair, representing the sum of the curvatures formed by adjacent line segments.
$L_{d}$ is an L1 depth loss calculated between the depth values rendered by the raw mesh and the strands.
$L_{m}$ is an L1 mask loss computed between the hair mask extracted from the input image and one of the rendered strands.
$L_{t}$ is a 3D tangent loss computed as the sum of cosine losses between the estimated 3D orientation and the strands both rendered in screen-space.
$w_{*}$ are the weights for each loss.

\subsubsection*{Child Hair Optimization}
In this step, we abandon the guide interpolation and optimize the vertex positions of child hair, $\mathscr{V}_{c}$, for finer alignment in the two stages.
First, We apply reparameterization with $k=K_{c}$ for child hair and perform optimization in $I^{0}_{c}$ iterations.
After the first stage, we relax the conditions by setting $k=0$ and execute the final alignment in $I^{1}_{c}$ iterations.
Both of these steps share a common loss term, $L_{c}$, which is minimized using the Adam optimizer.
\begin{equation}
L_{c} =  w_{d} * L_{d} + w_{m} * L_{m} +  w_{o} * L_{o} + R_{b}
\end{equation}
$L_{o}$ is a 2D orientation loss computed as the sum of absolute cosine losses between the 2D orientation extracted from the input image and that of the rendered strands.

\section{Experiments}
\label{sec:experiments}
Qualitative and quantitative comparisons were carried out with state-of-the-art strand reconstruction methods that use multi-view images: \textbf{LPMVS} \cite{Nam_2019_CVPR}, \textbf{Strand Integration} \cite{maeda2023refinement}, and \textbf{NeuralHaircut} \cite{Sklyarova_2023_ICCV}.
An open CPU implementation made by the authors of Strand Integration was used for LPMVS, while the other methods were tested with official implementations.
All methods require multi-view images with camera parameters as input.
NeuralHaircut was trained for 300k iterations in Stage 1 (surface) and 200k iterations in Stage 2 (strand).
The proposed method is implemented on the top of Blender, PyTorch, and Dressi-AD, a Vulkan-based DR framework with hardware rasterizer \cite{takimoto_2022_dressi}.
For Adam, learning rate was set to 0.001 with $\beta_1=0.9$ and $\beta_2 =0.999$.
Other hyperparameters were set as follows:
$\lambda = 50$,
$w_{root} = 1.0$,
$w_{stick} = 0.1$,
$w_{d} = 0.01$,
$w_{c} = 0.01$,
$w_{m} = 1$,
$w_{t} = 1$,
$w_{o} = 1$,
$K_{g} = 4$,
$K_{c} = 4$,
$I_{g} = 2000$,
$I^{0}_{c} = 2000$, and
$I^{1}_{c} = 1000$.
Our input raw mesh is reconstructed by OpenMVS \cite{openmvs2020} unless otherwise noted.
Detailed experimental settings and more results are available in the supplementary material.

\subsection{Synthetic data}
\begin{table}[t]\centering
\caption{Speed comparison on synthetic data.
We measured the time taken to generate strands starting from multiple image and camera parameter inputs on the same machine.
The time taken for our surface reconstruction by OpenMVS is set to 1, and relative times are shown for the others.
}
\label{tab:time}
\footnotesize
\begin{tabular}{lcccc} \hline
\scriptsize{Method}               &    LPMVS   & Strand Integ.  & NeuralHaircut & Ours    \\ \hline
{\scriptsize{Surface recon.}}   &    N/A      &   N/A       & 1080        & 1 \\
{\scriptsize{Strand recon.}} &   57  & 81    & 2160      & 25 \\ \hline
\end{tabular}
\end{table}
We numerically evaluated our method on hair models prepared by Yuksel et al. \cite{Yuksel2009}.
The 58 images were ray-traced in Blender Cycles using a virtual camera on a hemisphere under uniform lighting.
We did not use GTs as input, except for camera parameters.
We follow previous studies \cite{Nam_2019_CVPR, rosu2022neuralstrands, Sklyarova_2023_ICCV} that measure precision, recall, and F-score between reconstructed strands and GTs.
Nevertheless, we chose the 3D correspondence to evaluate the internal strands. 
3D space is searched per source sample to judge whether at least one destination sample is within the distance and angle error thresholds, and the quantities are computed.
Moreover, while angular error evaluation in the previous studies accepts an ambiguity of 180\textdegree, we evaluated it in a range of 360\textdegree~to assess absolute hair flow.

Quantitative comparison with existing methods and ablation study are shown in \tablename~\ref{tab:exp_cg}.
Our full pipeline shows better values than the other methods in all criteria.
In particular, the much higher recall values indicate that our method successfully recovers internal hair directions that were difficult to handle with existing methods.
The effectiveness of each component in the proposed pipeline is also validated.
For straight hair with simple internal flow, our strand initialization worked well and showed good scores even without DR.
For complex curly hair, the influence of other modules, global optimization, reparameterization, and guide-child hierarchy becomes more pronounced.

The performance of each method was also compared.
\tablename~\ref{tab:time} shows the time taken to process Curly Hair.
Typically, NeuralHaircut takes a couple of days, and LPMVS and Strand Integration require a few hours, but our pipeline finishes in less than one hour.

\subsection{Real data}
\begin{table*}[htbp]\centering
\caption{Quantitative comparison with existing methods and ablation study on synthetic data.
\textbf{P}, \textbf{R}, and \textbf{F1} denote precision, recall, and F1 score, respectively.
Higher is better.
The lower rows describe the values of our full pipeline and ours without individual modules.
w/o DR: DR optimization is not applied, and the initialized strands are evaluated.
w/o guide opt.: Child strands are optimized from the beginning of the DR step.
w/o reparam.: Reparameterization is disabled.
w/o $\mathscr{N}$: Only $\mathscr{N}$ is abandoned in the reparameterization.
w/o strand init.: Strands are initialized by straight lines parallel to the normal of the scalp.
w/o global opt.: Only gravity heuristic is applied to the initial 3D orientation, and 180\textdegree~ambiguity is accepted on the other steps.
}
\vspace*{-5mm}
\label{tab:exp_cg}
\small
\begin{tabular} {lwc{3.5mm}wc{3.5mm}wc{3.5mm}|wc{3.5mm}wc{3.5mm}wc{3.5mm}|wc{3.5mm}wc{3.5mm}wc{3.5mm}||wc{3.5mm}wc{3.5mm}wc{3.5mm}|wc{3.5mm}wc{3.5mm}wc{3.5mm}|wc{3.5mm}wc{3.5mm}wc{3.5mm}}\\
{}  & \multicolumn{9}{c||}{\textbf{Straight Hair}} & \multicolumn{9}{c}{\textbf{Curly Hair}}\\
\multicolumn{1}{r}{Threshold}  & \multicolumn{3}{c|}{1mm/10\textdegree} & \multicolumn{3}{c|}{2mm/20\textdegree} & \multicolumn{3}{c||}{3mm/30\textdegree} & \multicolumn{3}{c|}{1mm/10\textdegree} & \multicolumn{3}{c|}{2mm/20\textdegree} & \multicolumn{3}{c}{3mm/30\textdegree}\\
\multicolumn{1}{r}{Measure} & \textbf{P} & \textbf{R} & \textbf{F1} & \textbf{P} & \textbf{R} & \textbf{F1} & \textbf{P} & \textbf{R} & \textbf{F1} & \textbf{P} & \textbf{R} & \textbf{F1} & \textbf{P} & \textbf{R} & \textbf{F1} & \textbf{P} & \textbf{R} & \textbf{F1}\\
\hline
LPMVS \cite{Nam_2019_CVPR} & 29.9 & 27.5 & 28.6 & 39.7 & 52.2 & 45.1 & 42.9 & 66.8 & 52.2 & 18.4 & 6.1 & 9.1 & 32.8 & 15.3 & 20.9 & 37.2 & 23.9 & 29.1\\
Strand Integration \cite{maeda2023refinement} & 33.4 & 34.1 & 33.7 & 42.4 & 55.4 & 48.1 & 44.8 & 66.8 & 53.6 & 19.4 & 6.7 & 9.9 & 34.3 & 16.1 & 21.9 & 38.5 & 23.6 & 29.3\\
NeuralHaircut \cite{Sklyarova_2023_ICCV} & 50.2 & 14.9 & 23.0 & 76.1 & 29.2 & 42.2 & 85.6 & 38.4 & 53.1 & 20.9 & 3.9 & 6.6 & 58.1 & 14.5 & 23.2 & 80.0 & 27.3 & 40.7\\
\hline
Ours & 60.3 & 46.4 & 52.5 & 88.2 & 84.3 & 86.2 & \textbf{94.5} & 93.6 & \textbf{94.1} & \textbf{38.3} & 23.6 & \textbf{29.2} & \textbf{79.1} & 61.0 & \textbf{68.9} & \textbf{90.0} & 81.0 & \textbf{85.3}\\
Ours (w/o DR) & \textbf{65.4} & 41.8 & 51.0 & \textbf{88.6} & 78.8 & 83.4 & 93.3 & 88.6 & 90.9 & 22.1 & 15.7 & 18.4 & 59.2 & 56.1 & 57.6 & 75.8 & 82.0 & 78.8\\
Ours (w/o guide opt.) & 61.1 & 46.8 & \textbf{53.0} & 86.8 & 86.1 & \textbf{86.5} & 92.7 & 95.0 & 93.9 & 36.8 & 22.9 & 28.2 & 77.4 & 60.6 & 68.0 & 88.8 & 80.7 & 84.6\\
Ours (w/o reparam.) & 7.9 & 42.5 & 13.4 & 24.0 & \textbf{97.1} & 38.5 & 38.8 & \textbf{99.9} & 55.9 & 6.4 & \textbf{29.5} & 10.6 & 23.1 & \textbf{93.5} & 37.1 & 39.3 & \textbf{99.6} & 56.4\\
Ours (w/o $\mathscr{N}$) & 59.5 & \textbf{46.9} & 52.5 & 86.6 & 85.4 & 86.0 & 92.8 & 94.3 & 93.5 & 36.5 & 23.2 & 28.3 & 76.4 & 60.7 & 67.6 & 87.6 & 80.7 & 84.0\\
Ours (w/o strand init.) & 5.3 & 1.0 & 1.6 & 23.4 & 7.0 & 10.7 & 37.7 & 18.6 & 24.9 & 9.3 & 4.9 & 6.4 & 22.3 & 23.8 & 23.0 & 32.4 & 48.0 & 38.7\\
Ours (w/o global opt.) & 58.1 & 45.7 & 51.2 & 85.5 & 85.8 & 85.7 & 91.8 & 94.7 & 93.2 & 26.1 & 13.6 & 17.9 & 63.5 & 46.4 & 53.6 & 75.3 & 68.5 & 71.8\\
\end{tabular}
\end{table*}
\begin{figure*}[htbp]
\centering
\begin{tabular}{ccccc}
\includegraphics[width=.16\textwidth]{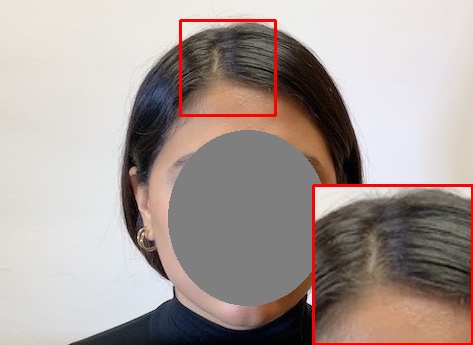} &
\includegraphics[width=.16\textwidth]{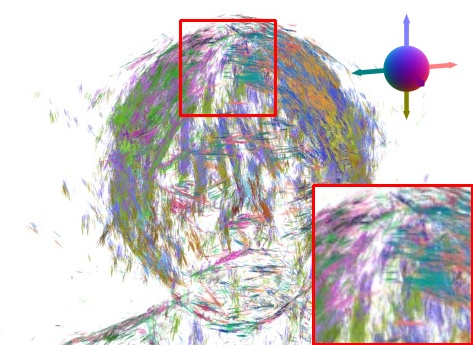} &
\includegraphics[width=.16\textwidth]{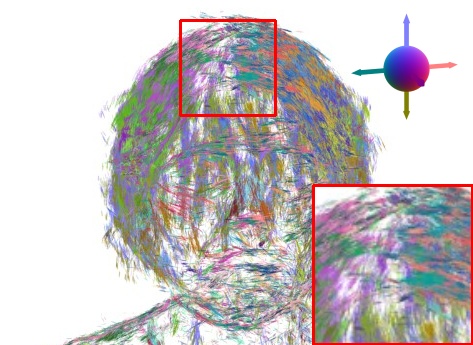} &
\includegraphics[width=.16\textwidth]{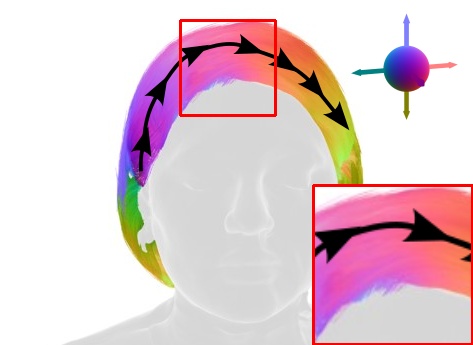} &
\includegraphics[width=.16\textwidth]{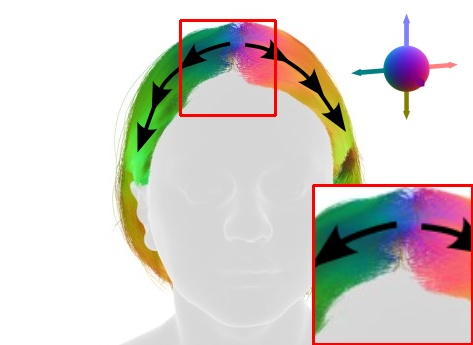} \\
\includegraphics[width=.16\textwidth]{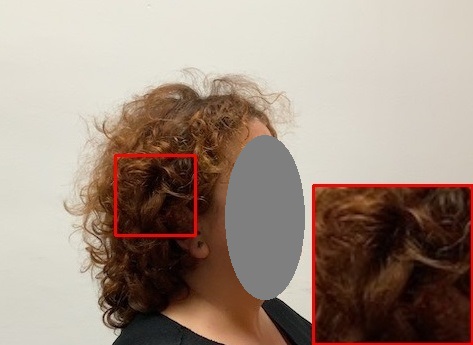} &
\includegraphics[width=.16\textwidth]{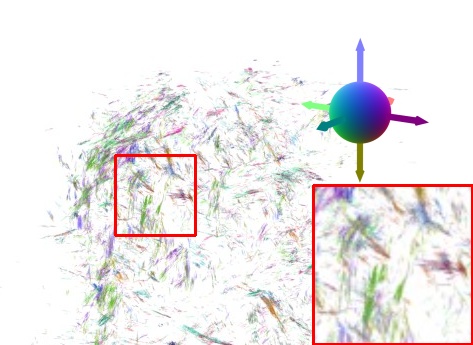} &
\includegraphics[width=.16\textwidth]{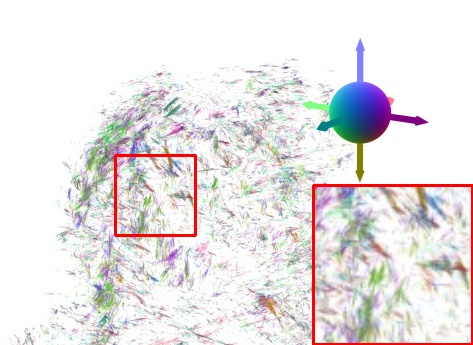} &
\includegraphics[width=.16\textwidth]{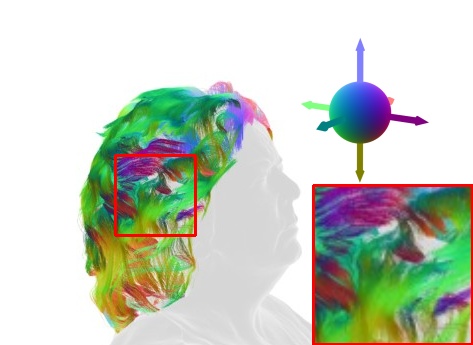} &
\includegraphics[width=.16\textwidth]{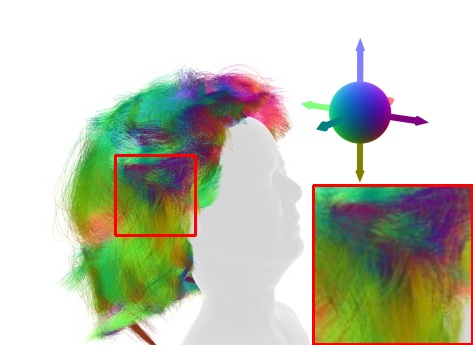} \\
\includegraphics[width=.16\textwidth]{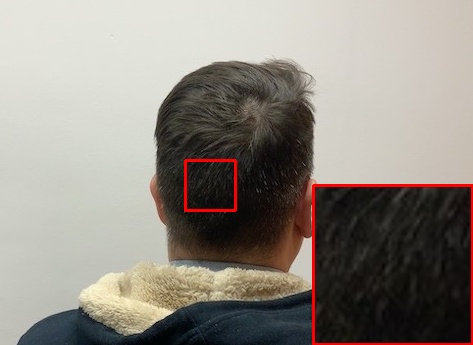} &
\includegraphics[width=.16\textwidth]{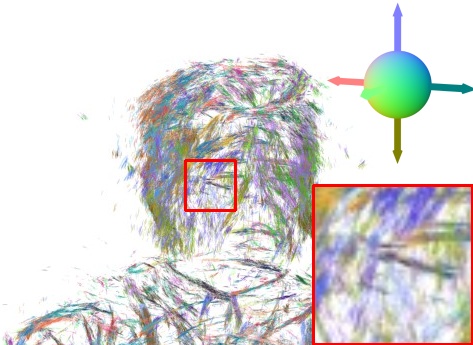} &
\includegraphics[width=.16\textwidth]{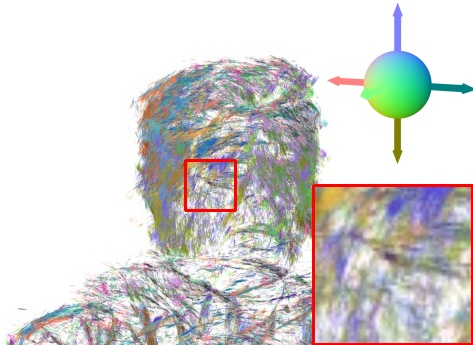} &
\includegraphics[width=.16\textwidth]{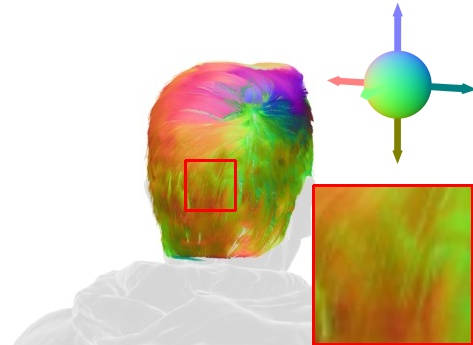} &
\includegraphics[width=.16\textwidth]{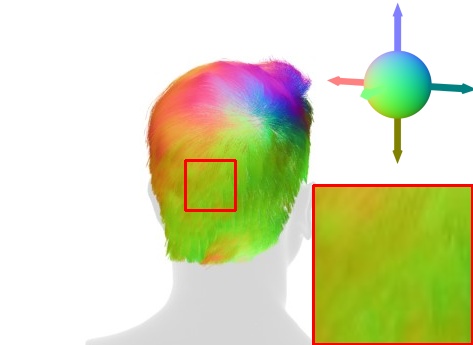} \\
\textbf{Image}  & \textbf{LPMVS} & \textbf{Strand Integration} & \textbf{NeuralHaircut} & \textbf{Ours}
\end{tabular}
\vspace*{-2mm}
\caption{Comparison on a real-world multi-view dataset, H3DS \cite{ramon2021h3d}\protect\footnotemark .
Color and arrows represent 3D orientation of each strand.
In the upper row, our method identifies the hair parting.
In the middle row, ours reconstructs dense hair with no visible white scalp.
In the bottom row, a smooth hair flow is estimated by ours.
NeuralHaircut struggles in all cases.
LPMVS and Strand Integration are prone to flying noise.}
\label{fig:exp_h3ds}
\end{figure*}
\begin{figure*}[htbp]
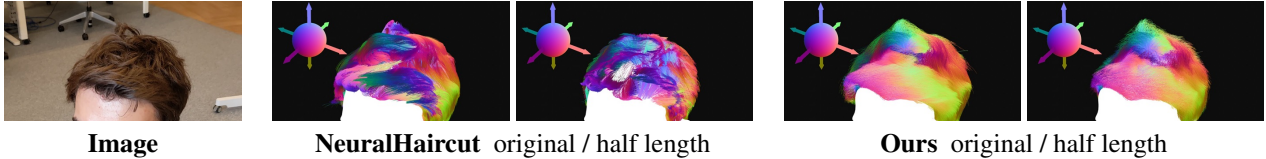

\begin{tabular}{ccc}
\adjincludegraphics[width=.18\textwidth, trim={0 {.5\height} 0 {.0\height}},clip]{{\dataroot}HairPaper2023/datasets/result_rendered/monocular/real.jpg} &
\adjincludegraphics[width=.18\textwidth, trim={0 {.5\height} 0 {.0\height}},clip]{{\dataroot}HairPaper2023/datasets/result_rendered/monocular/process/neural_100.jpg}
\adjincludegraphics[width=.18\textwidth, trim={0 {.5\height} 0 {.0\height}},clip]{{\dataroot}HairPaper2023/datasets/result_rendered/monocular/process/neural_050.jpg} &
\adjincludegraphics[width=.18\textwidth, trim={0 {.5\height} 0 {.0\height}},clip]{{\dataroot}HairPaper2023/datasets/result_rendered/monocular/process/ours_100.jpg}
\adjincludegraphics[width=.18\textwidth, trim={0 {.5\height} 0 {.0\height}},clip]{{\dataroot}HairPaper2023/datasets/result_rendered/monocular/process/ours_050.jpg}\\
\textbf{Image}  & \textbf{NeuralHaircut}~~original / half length & \textbf{Ours}~~original / half length
\end{tabular}
\vspace*{-2mm}
\caption{Comparison on a hand-held monocular video captured by a smartphone \cite{Sklyarova_2023_ICCV}.
On each method, the left image shows the original reconstruction, and the diagram on the right shows the hair length edited in half.
Our method demonstrates robust reconstruction even under a severe capturing condition.
Moreover, the half length image indicates our hair is editable thanks to the correct direction.
}
\label{fig:exp_mono}
\end{figure*}
We show comparisons on H3DS dataset \cite{ramon2021h3d} in \figurename~\ref{fig:exp_h3ds}.
GT raw meshes and camera parameters are used for all methods but they are not very accurate.
Our method is capable of reconstructing reasonable results despite the limited accuracy of the input data.

\figurename~\ref{fig:exp_mono} visualizes results on a monocular video sequence.
In even worse calibrations, our method robustly reconstructs the strands.
Hair length editing is also performed to prove that our method can reconstruct accurate hair flow.

\begin{figure}[htbp]
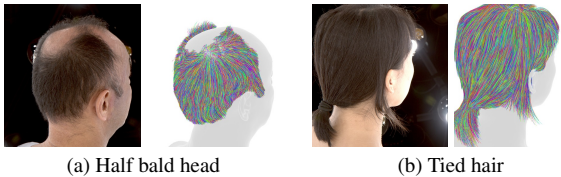
\centering
\begin{minipage}[t]{0.48\linewidth}
\centering
\adjincludegraphics[width=1.8cm, trim={0 {.20\height} 0 {.1\height}},clip]{{\dataroot}HairPaper2023/figures/fdc/subject3/hair/images/Cam45_1.jpg}
\adjincludegraphics[width=1.8cm, trim={0 {.20\height} 0 {.1\height}},clip]{{\dataroot}HairPaper2023/datasets/result_rendered/fdc/subject3/fdc_t00455682_ours__randcolor0000am_DrFmt_44.jpg}
\subcaption{Half bald head}
\end{minipage}
\begin{minipage}[t]{0.48\linewidth}
\centering
\adjincludegraphics[width=1.8cm, trim={0 {.05\height} 0 {.25\height}},clip]{{\dataroot}HairPaper2023/figures/fdc/subject4/hair/images/Cam45_1.jpg}
\adjincludegraphics[width=1.8cm, trim={0 {.05\height} 0 {.25\height}},clip]{{\dataroot}HairPaper2023/datasets/result_rendered/fdc/subject4/fdc_p84183835_ours__randcolor0000am_DrFmt_44.jpg}
\subcaption{Tied hair}
\end{minipage}
\vspace*{-3mm}
\caption{Robustness against challenging hairstyles. Our method can reconstruct (a) a half bald head and (b) tied hair.}
\label{fig:exp_lc2}
\end{figure}

\begin{figure}[htbp]
\begin{tabular}{lccc}
\raisebox{+5mm}{\rotatebox{90}{\textbf{NeuralHaircut}}}  &
\raisebox{-.0\height}{\includegraphics[width=.13\textwidth]{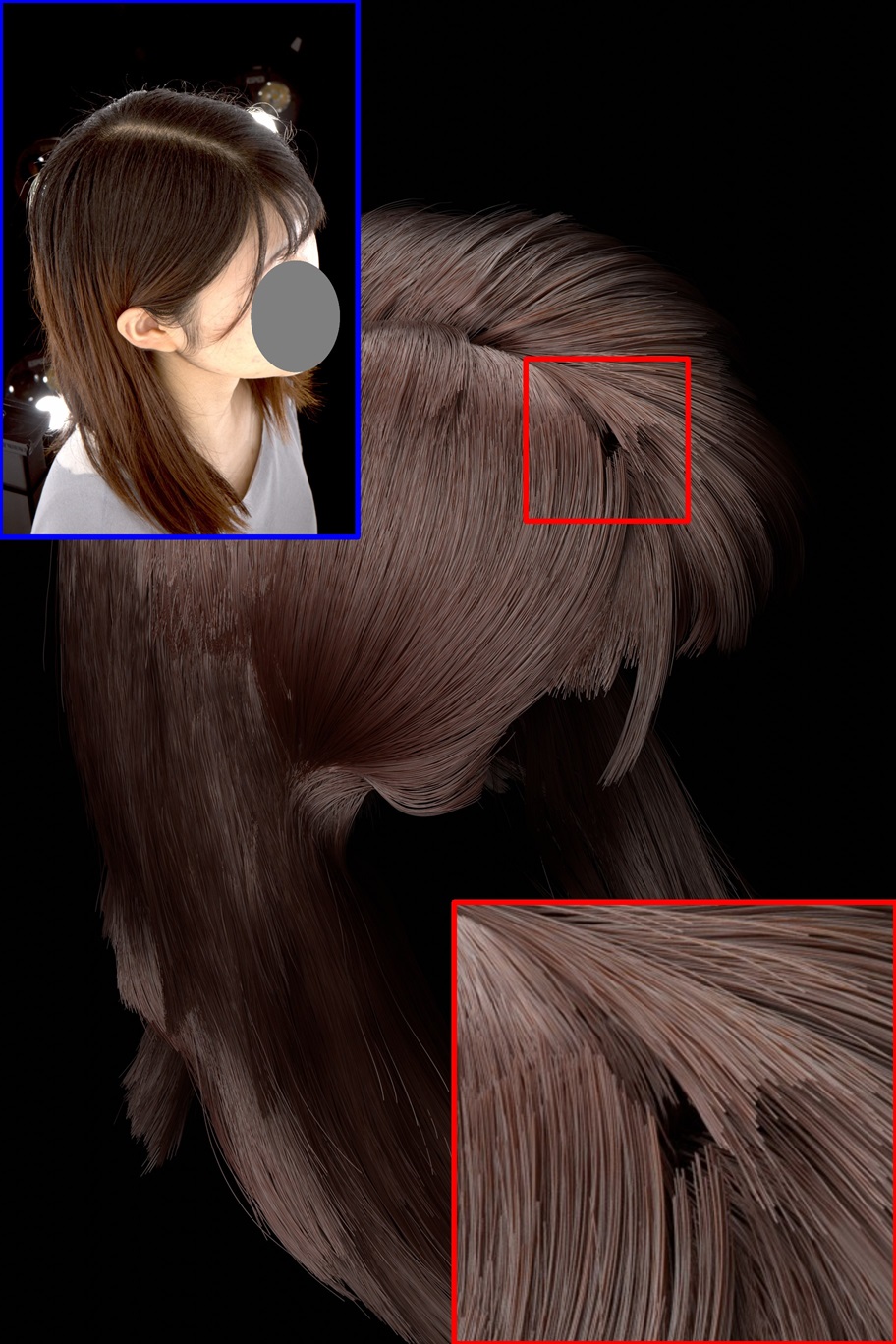}} &
\raisebox{-.0\height}{\includegraphics[width=.13\textwidth]{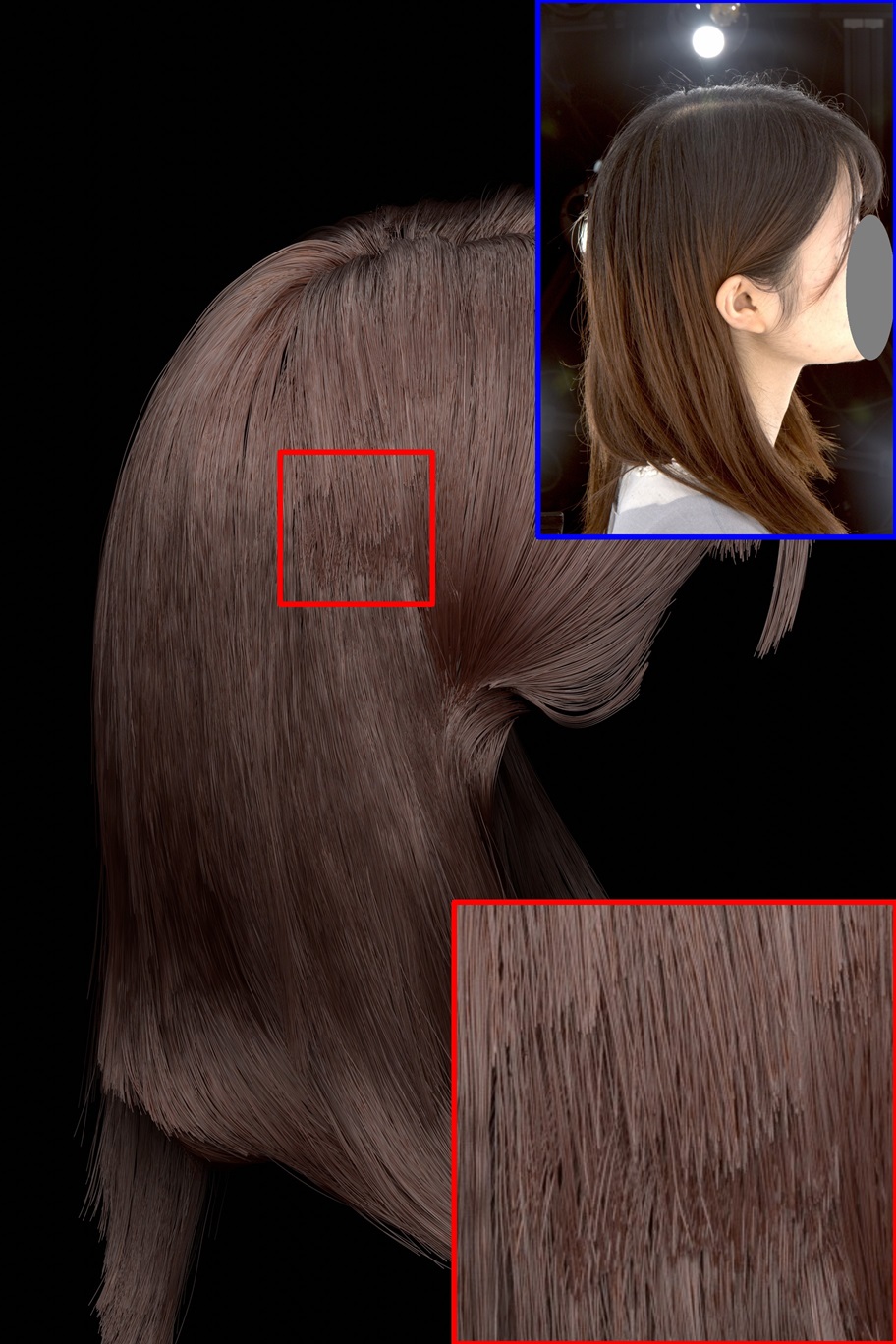}} &
\raisebox{-.0\height}{\includegraphics[width=.13\textwidth]{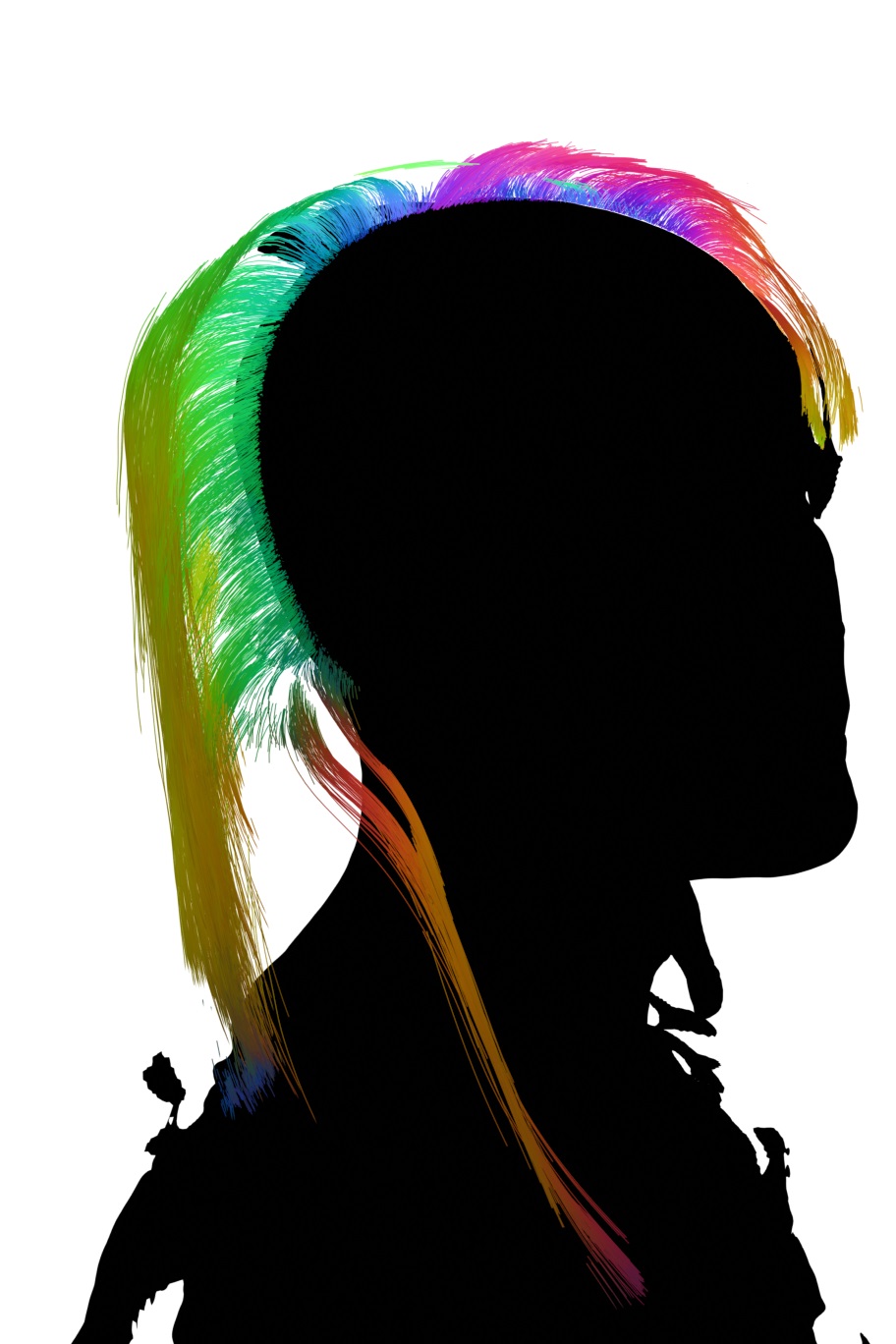}} \\
\raisebox{+10mm}{\rotatebox{90}{\textbf{Ours}}}  &
\raisebox{-.0\height}{\includegraphics[width=.13\textwidth]{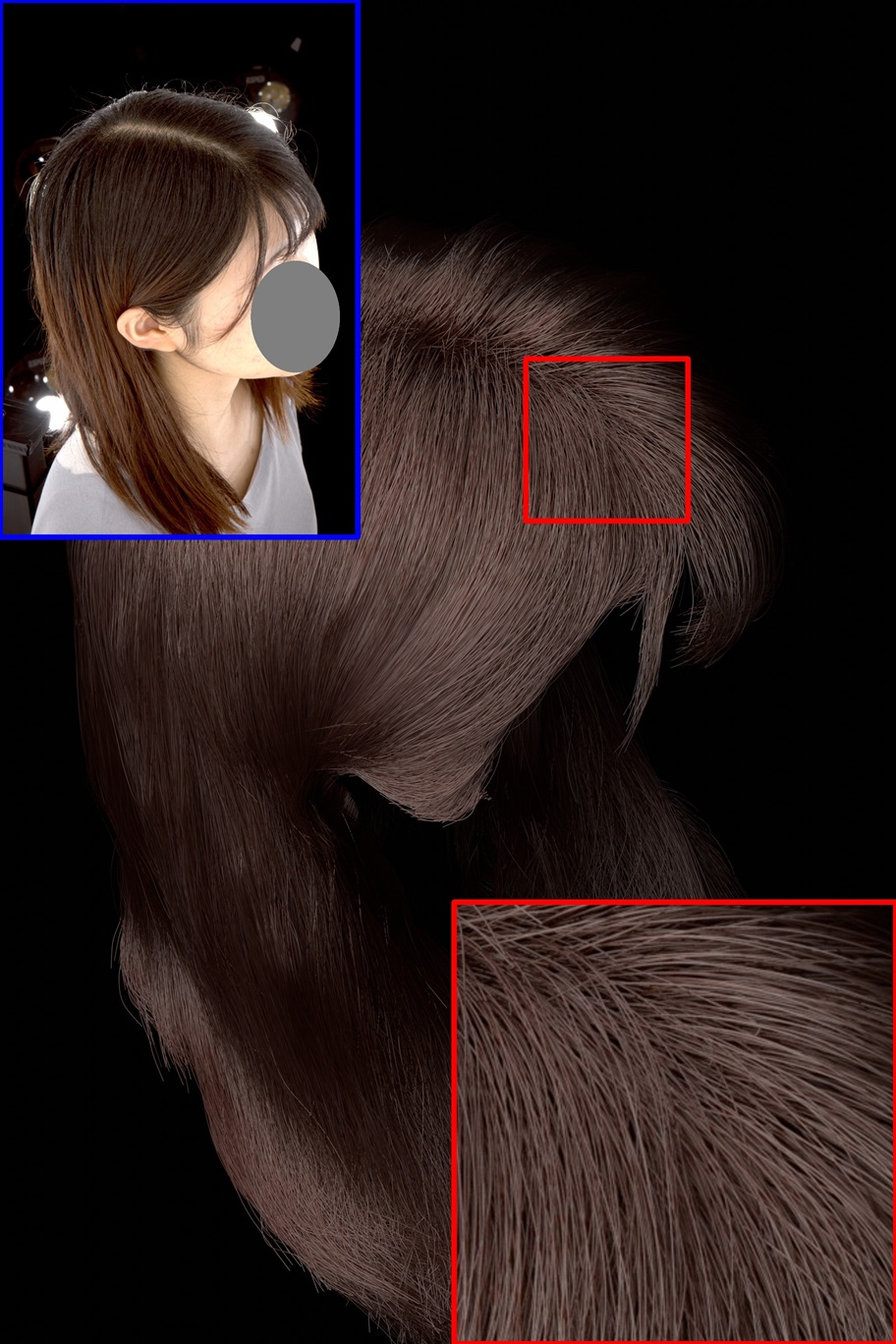}} &
\raisebox{-.0\height}{\includegraphics[width=.13\textwidth]{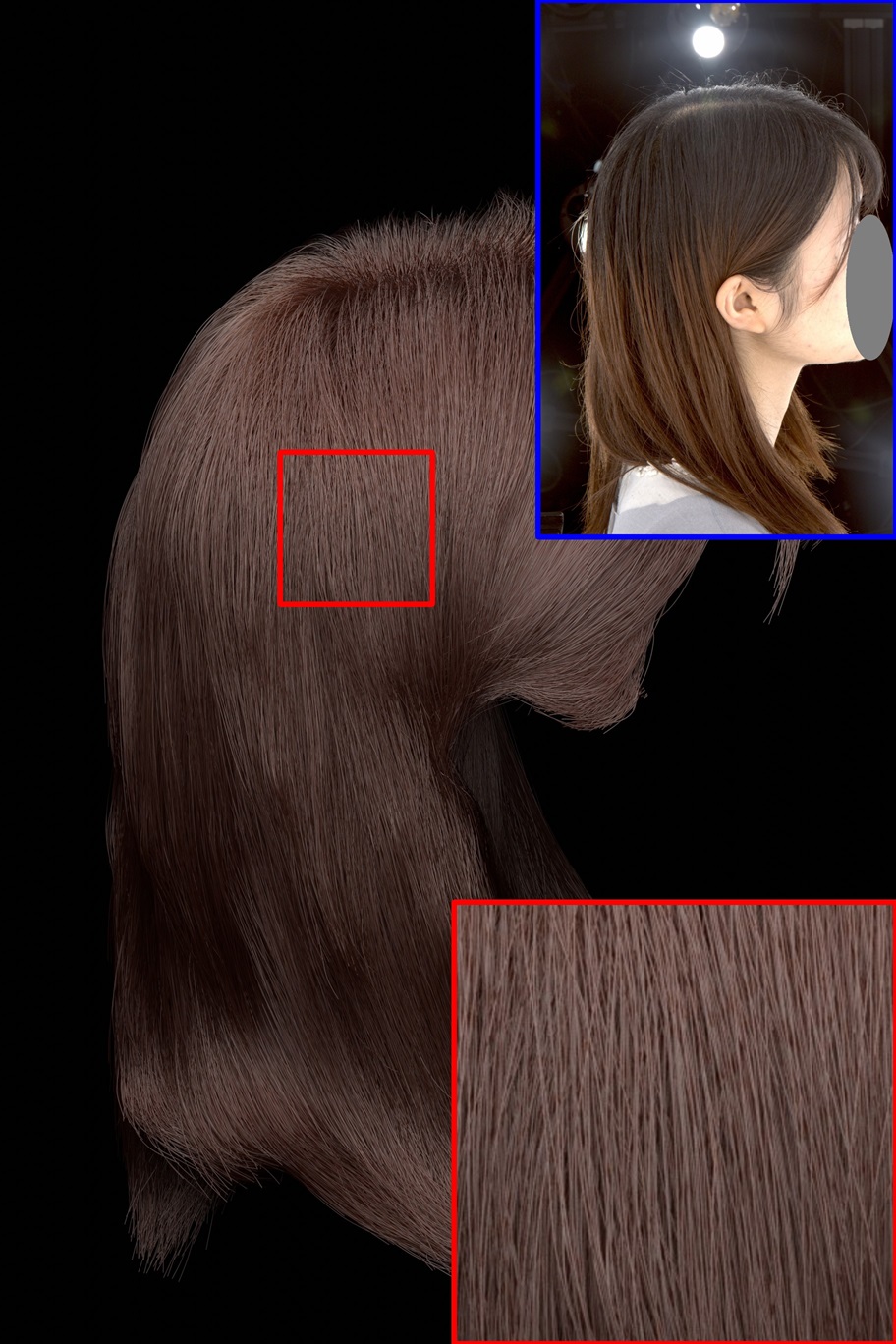}} &
\raisebox{-.0\height}{\includegraphics[width=.13\textwidth]{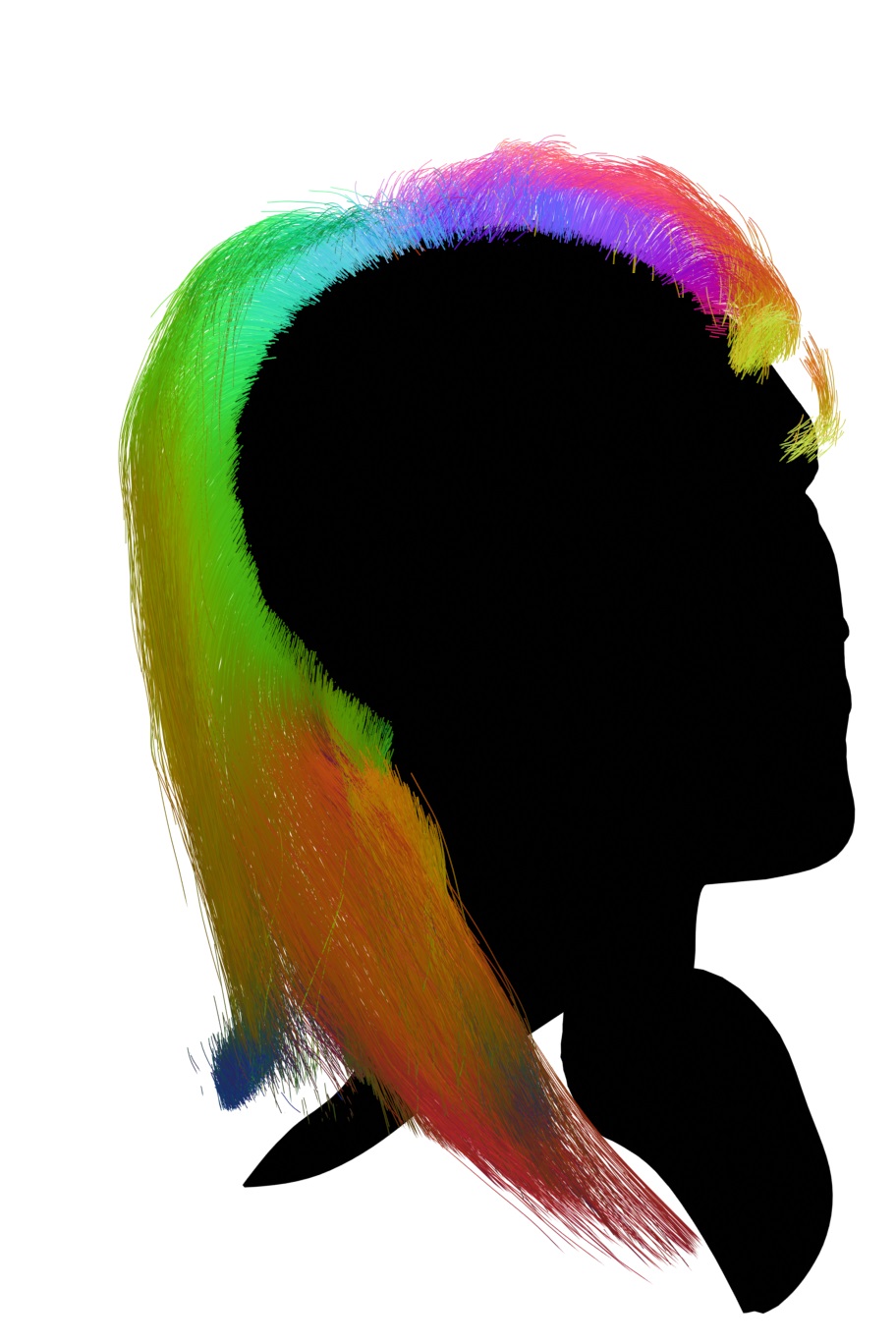}}
\end{tabular}
\caption{Left and middle: Re-rendering comparison.
The camera image is put on the top for hairstyle reference.
The images are ray-traced by Blender Cycles, using the common hair material and lighting.
Incorrect orientation of NeuralHaircut made artifacts at meeting points of opposite hair flow, which ours does not have.
Right: Volumetric slice of the middle view with certain near and far planes.
NeuralHaircut's hair is concentrated on the surface and scanty inside, whereas our method fills inside uniformly. }
\label{fig:exp_rendering}
\end{figure}

\begin{figure}[htbp]\centering
\begin{tabular}{lccc}
\raisebox{+5mm}{\rotatebox{90}{\textbf{NeuralHaircut}}}  &
\includegraphics[width=3cm]{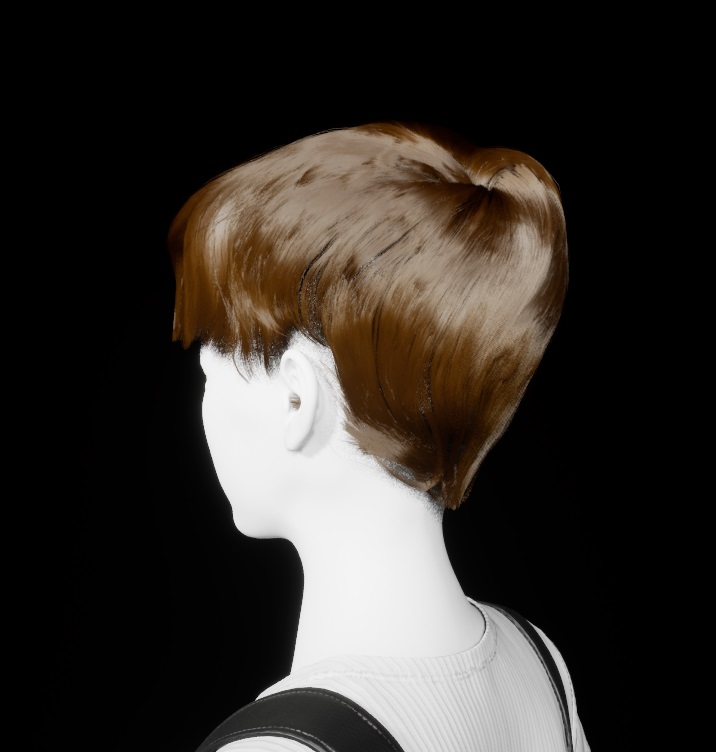} &
\includegraphics[width=3cm]{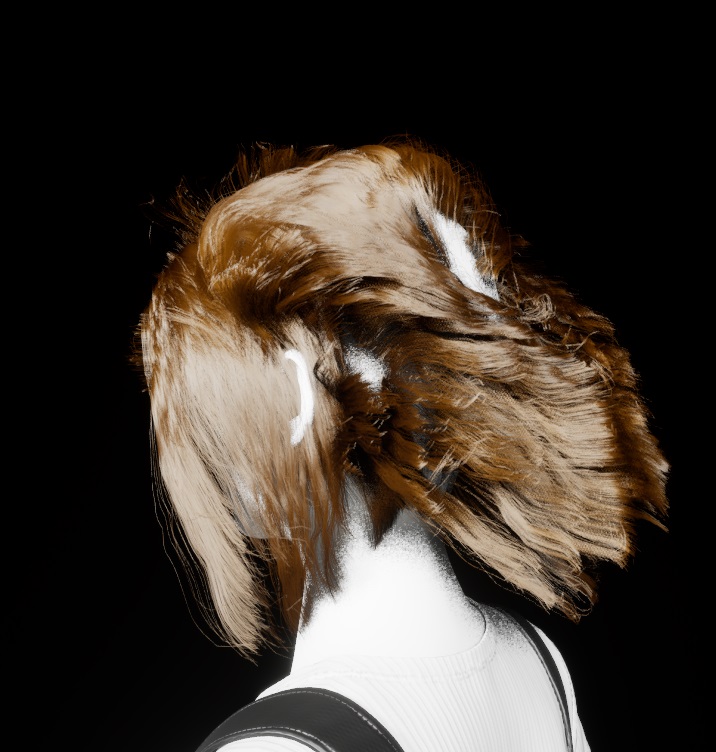} \\
\raisebox{+10mm}{\rotatebox{90}{\textbf{Ours}}}  &
\includegraphics[width=3cm]{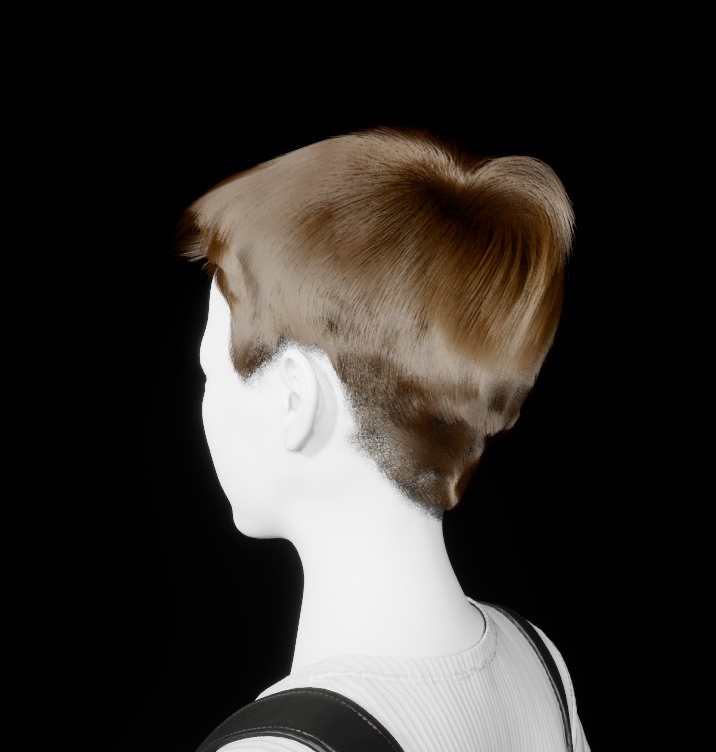} &
\includegraphics[width=3cm]{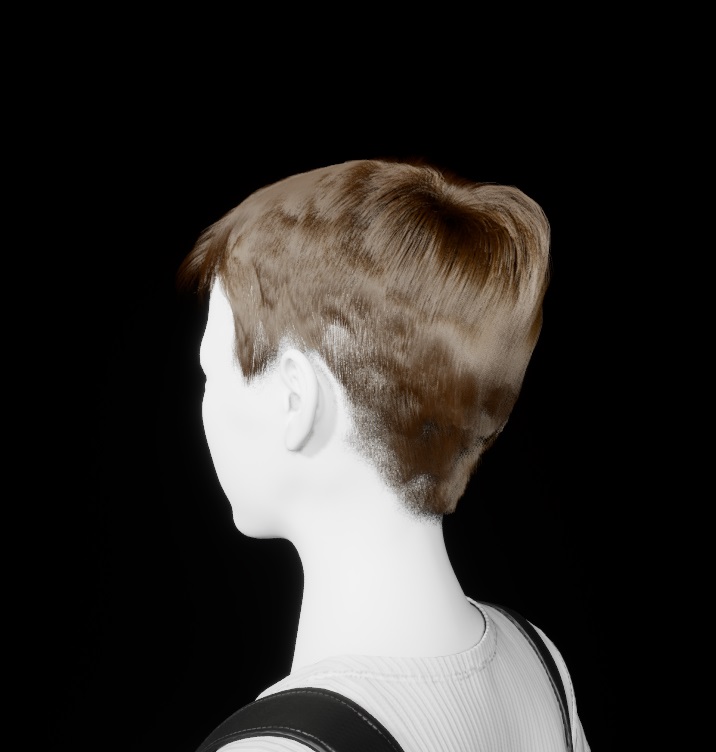} \\
& Initial state & After simulation
\end{tabular}
\caption{Physics simulation with gravity and certain hair stiffness applied to the same subject in \figurename~\ref{fig:exp_lc}.
The left image is the initial state given by reconstruction, and the right one is after simulation.
NeuralHaircut's hairs are oriented from bottom to top, so the simulated result is severely affected by unnatural sagging.
Thanks to the correct direction, our hair behaves naturally.}
\label{fig:exp_simu}
\end{figure}

We also conducted comparisons on a well-calibrated studio setup.
The 58 images taken by cameras positioned on a hemisphere under uniform illumination are used.
The results are shown in \figurename~\ref{fig:exp_lc}.
Ours reconstructs better strands than the existing methods in terms of direction.
\figurename~\ref{fig:exp_lc2} displays our other results that the challenging hairstyles are successfully handled.
Re-rendering comparison with reconstructed strands is shown in \figurename~\ref{fig:exp_rendering}.
Our strands are shaded photorealistic, indicating that our method's output is portable among rendering engines.
Furthermore, the density distribution of our hair is more reasonable than NeuralHaircut's.
\figurename~\ref{fig:exp_simu} visualizes the comparison of physics simulation starting from the reconstructed strands.
Although NeuralHaircut suffers from incorrect directions, ours exhibits reasonable behavior.

\footnotetext{The images of this dataset are only used for testing and comparison with existing methods and not used for algorithm improvement or training.}

\begin{figure}[htbp]
\begin{tabular}{ccc}\centering
\adjincludegraphics[width=.15\textwidth, trim={{.25\width} {.3\height} 0 {.1\height}},clip]{{\dataroot}HairPaper2023/figures/f7e930d8a9ff2091_img_0054_process.jpg} &
\includegraphics[height=.11\textwidth]{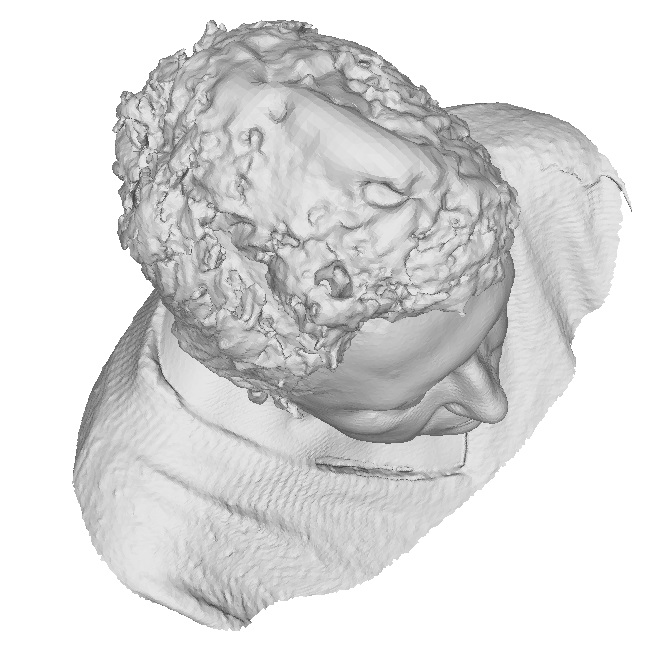} &
\adjincludegraphics[width=.15\textwidth, trim={{.25\width} {.3\height} 0 {.1\height}},clip]{{\dataroot}HairPaper2023/datasets/result_rendered/h3ds/f7e930d8a9ff2091/h3ds_f7e930d8a9ff2091_ours__tangent0000am_DrFmt_54.jpg} \\
\footnotesize \textbf{Image} &\footnotesize \textbf{Raw mesh} & \footnotesize \textbf{Reconstruction}
\end{tabular}
\caption{A limitation of our method. Raw mesh with inaccurate top geometry degenerates hair flow.}
\label{fig:limitation}
\end{figure}

\subsection{Limitations}
Our method is affected by the input raw mesh quality.
If the raw mesh shape has a notable difference from the subject, our strand reconstruction deteriorates, as shown in \figurename~\ref{fig:limitation}.
In addition, the assumption of a smooth flow makes it challenging to handle discontinuous hairstyles such as braids.
Moreover, protruding strands can disrupt the overall flow, limiting our accuracy, particularly for highly curly or spiky hair.
Finally, as shown in \figurename~\ref{fig:exp_rendering}, shaded hair color deviates from the actual one because our method does not recover material and lighting.

\section{Conclusion}
\label{sec:conlusion}
In this paper, we have introduced Dr.Hair, a novel method for reconstructing detailed human hair strands from multi-view images.
Our approach recovers consistent surface orientations, estimates the internal flow using differential equations, and performs optimization based on differentiable rendering, leveraging the hierarchical relationship between guide and child strands.
Our method's effectiveness has been demonstrated through both qualitative and quantitative evaluations.
Our method is capable of reconstructing a wide variety of hairstyles grown from a scalp without relying on priors trained on synthetic datasets, which are typically created through labor-intensive manual work.
Moreover, our proposed method outperforms existing methods in terms of processing speed.
We believe that our method can significantly contribute to the development of a cost-effective, photorealistic human digitization system.

{
    \small
    \bibliographystyle{ieeenat_fullname}
    \bibliography{main}
}

\clearpage
\maketitlesupplementary
\section{Experimental settings}
\label{sec:settings}

\subsection{Synthetic data: Cem Yuksel's hair models \cite{Yuksel2009}}
We used wStraight, wCurly, and wWavy models, all of which have 50,000 strands.
The head model, accompanied by the hair models, was attached for rendering by Blender Cycles.
A white uniform environment map was used for illumination.
Camera parameters were set the same as the real studio data discussed later.
\subsection{Real data}
\subsubsection*{H3DS \cite{ramon2021h3d}}
H3DS is a real-world multi-view dataset for head reconstruction.
GT Head meshes were scanned by a laser scanner, and independently captured multi-view images surrounding the subject in 360\textdegree~were registrated against the mesh.
About 70 images are provided per subject.
Top views are not well captured in both the mesh and the images.
Moreover, some views are affected by strong flash lighting that deteriorates image quality.
For NeuralHaircut, by following the official implementation setting, 32 clean views manually annotated by the dataset authors were used.
For the other methods, all views were inputted.
\subsubsection*{Monocular hand-held video \cite{Sklyarova_2023_ICCV}}
A subject asked to be as static as possible on a chair was captured in circular motion by a smartphone.
Subsampled 60 frames are provided.
Camera parameters were estimated with COLMAP \cite{schoenberger2016sfm}.
\subsubsection*{Studio data}
Original 58 images were captured in 1824x2736 pixels by DSLRs with hardware synchronized shooting.
The cameras were evenly put on a hemisphere, and similarly positioned LEDs were illuminated for uniform lighting.
Camera parameters and a raw mesh were estimated by MetaShape \cite{metashape}.
The images were resized to a height of 684 pixels for LPMVS and Strand Integration and 512 pixels for NeuralHaircut and ours.
\subsection{Existing methods' settings}
\subsubsection*{LPMVS \cite{Nam_2019_CVPR} and Strand Integration \cite{maeda2023refinement}}
Default values were used for most parameters.
Reasonable values were set to scene-dependent minimum and maximum depth according to the distance between the camera and the subjects.
\subsubsection*{NeuralHaircut \cite{Sklyarova_2023_ICCV}}
We followed the instructions to run the official implementation, including some manual processes.
50,000 strands were sampled for visualization and quantitative evaluation while 1,900 strands were used for training as in the default setting.

\begin{figure*}[htbp]\centering
\begin{tabular}{ccccc}
\includegraphics[width=.17\textwidth]{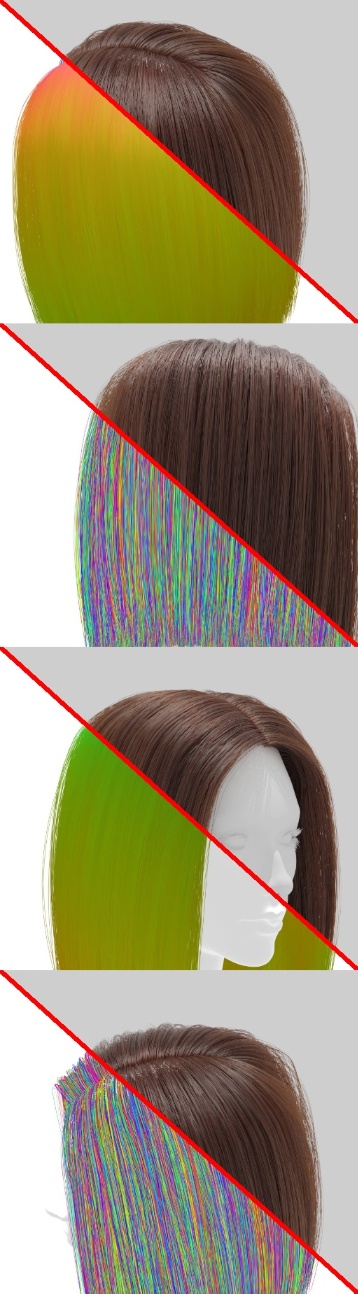} &
\includegraphics[width=.17\textwidth]{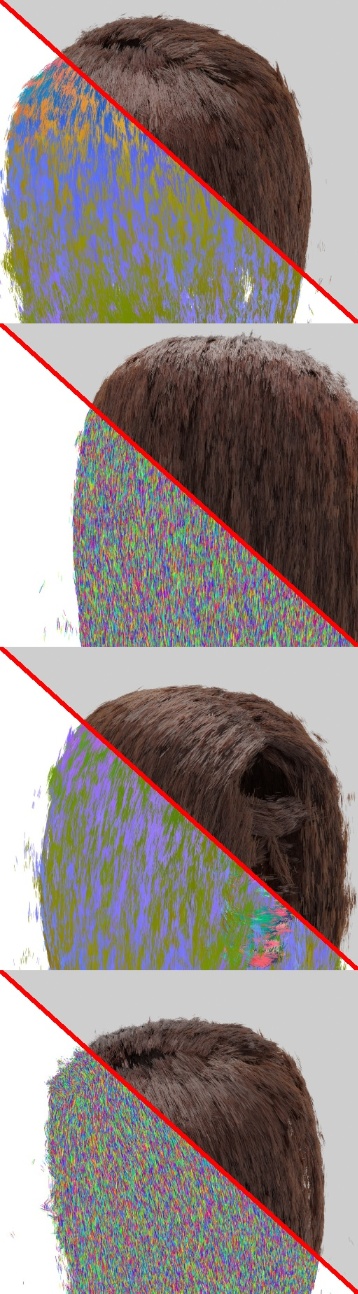} &
\includegraphics[width=.17\textwidth]{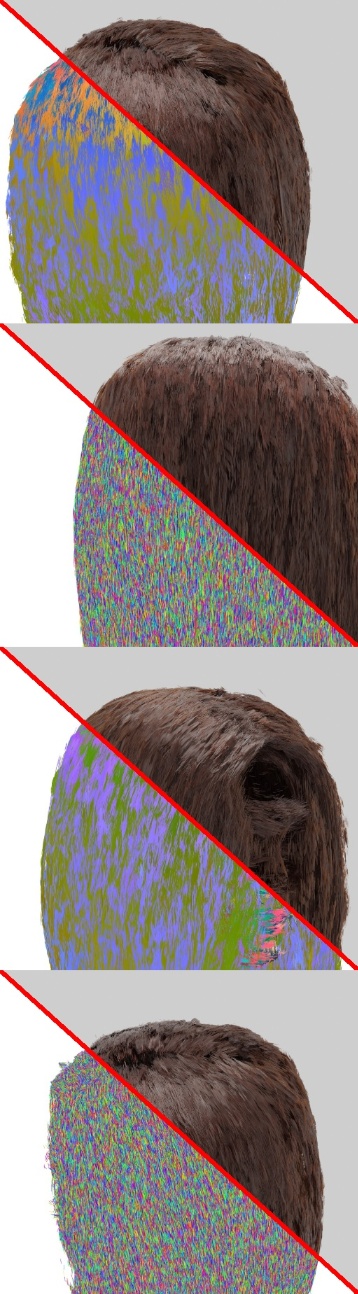} &
\includegraphics[width=.17\textwidth]{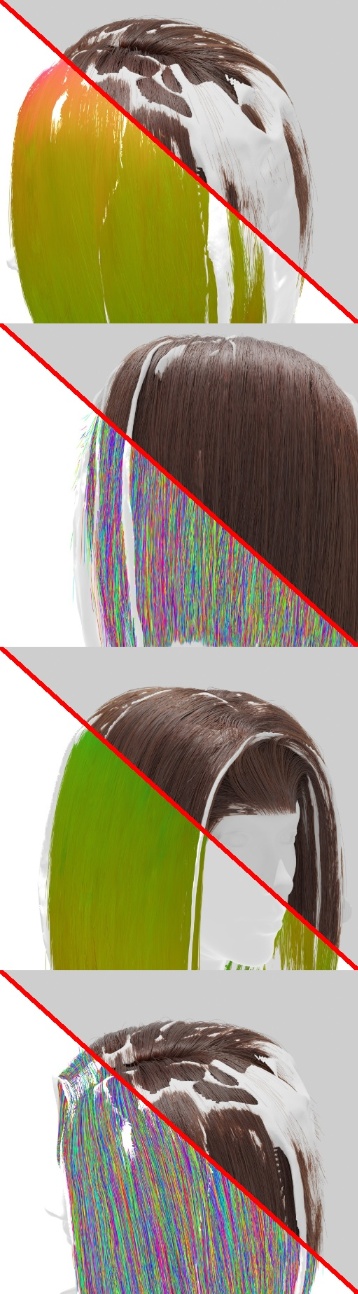} &
\includegraphics[width=.17\textwidth]{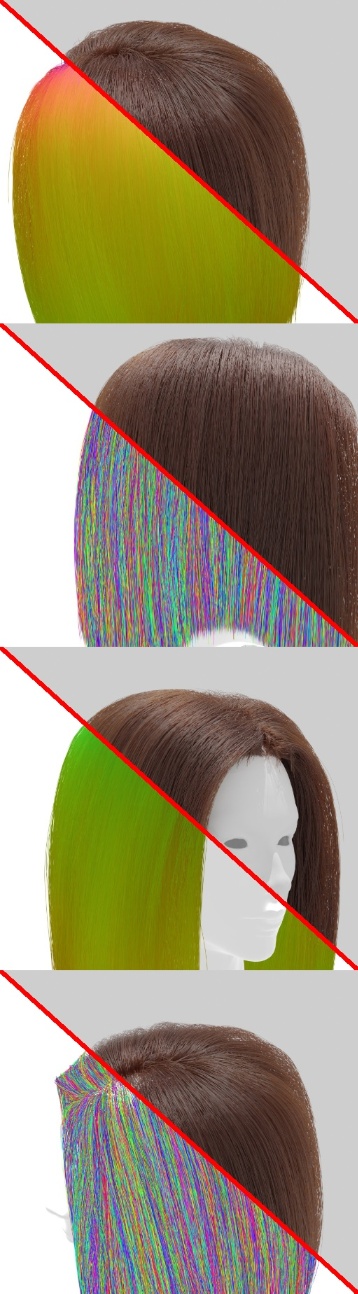} \\
\textbf{GT}  & \textbf{LPMVS} & \textbf{Strand Integration} & \textbf{NeuralHaircut} & \textbf{Ours} \\
\end{tabular}
\begin{tabular}{cccccc}
\includegraphics[width=.14\textwidth]{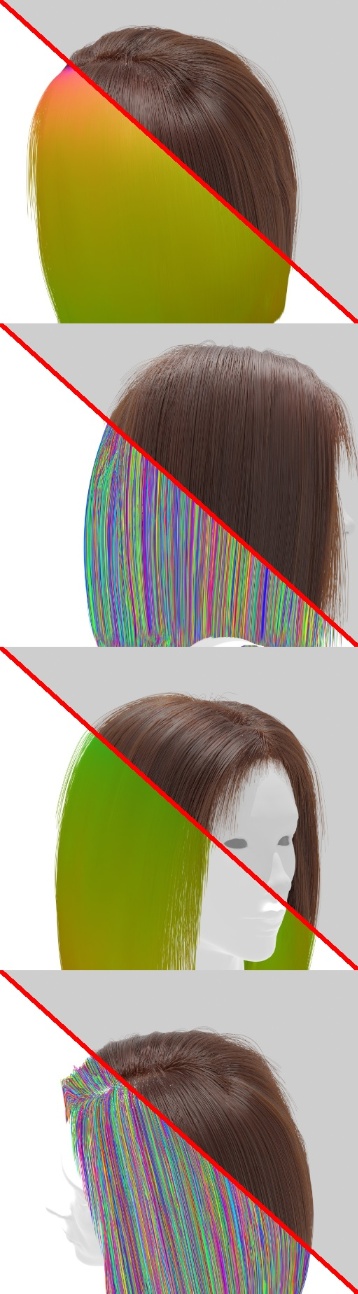} &
\includegraphics[width=.14\textwidth]{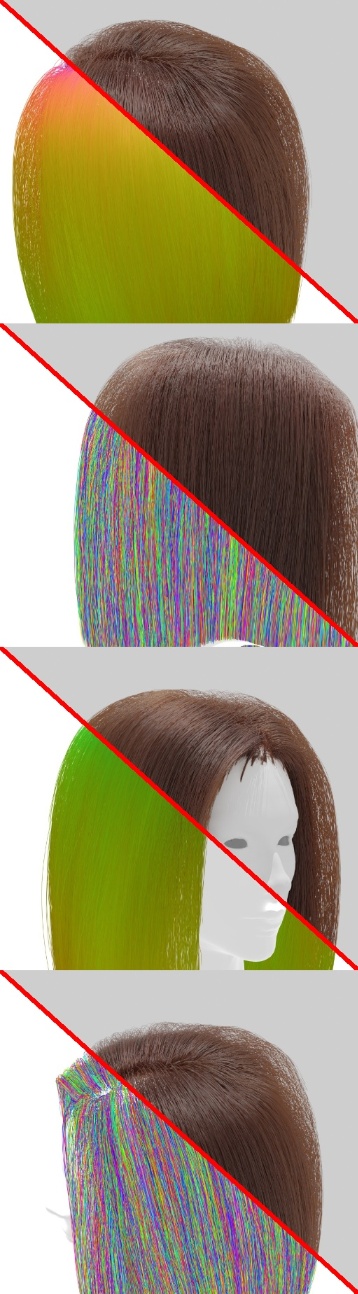} &
\includegraphics[width=.14\textwidth]{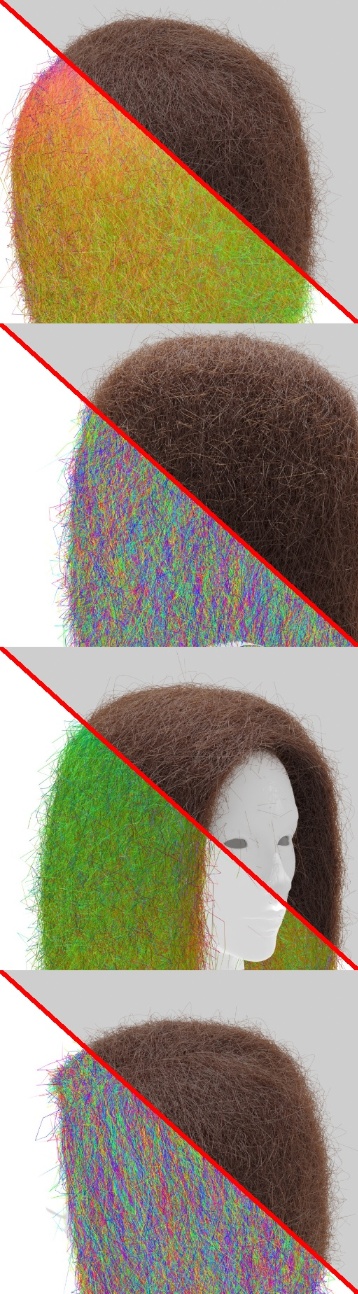} &
\includegraphics[width=.14\textwidth]{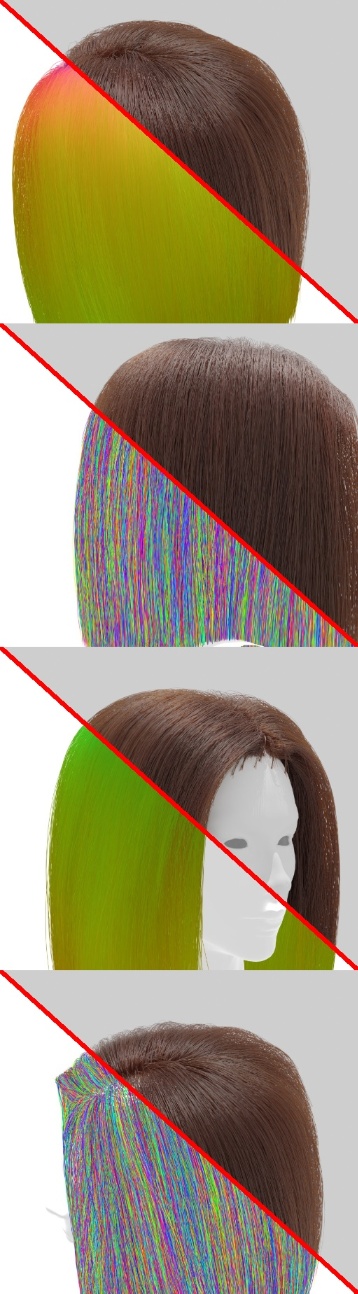} &
\includegraphics[width=.14\textwidth]{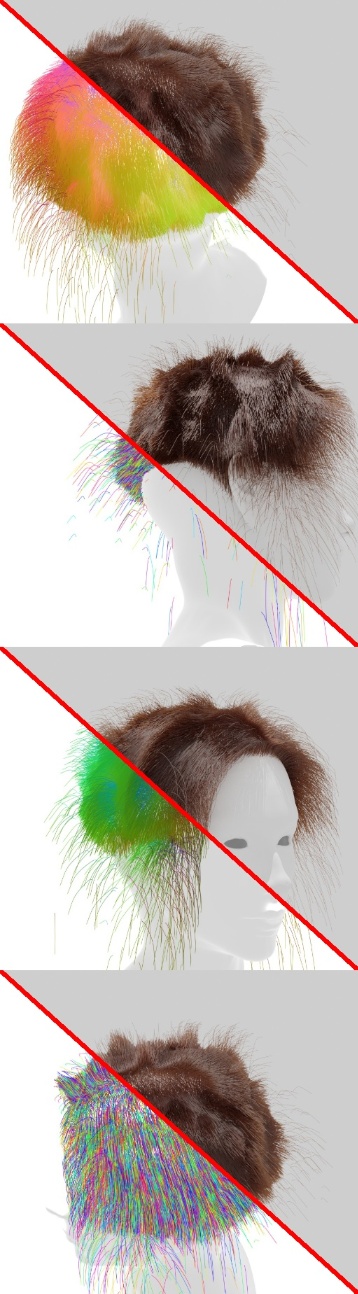} &
\includegraphics[width=.14\textwidth]{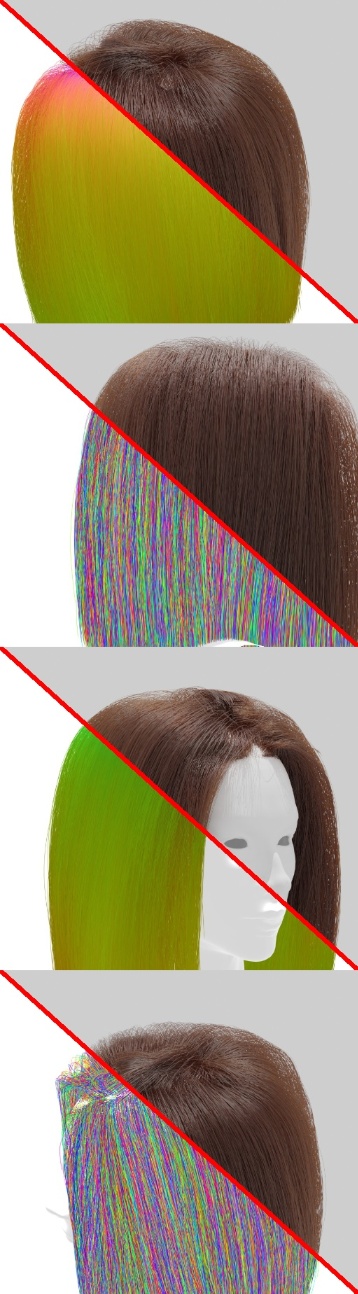} \\
\textbf{w/o DR} & \textbf{w/o guide opt.} & \textbf{w/o reparam.} & \textbf{w/o $\mathscr{N}$} & \textbf{w/o strand init.} & \textbf{w/o global opt.}\\
\end{tabular}
\vspace*{-2mm}
\caption{Qualitative evaluation on synthetic Straight Hair, corresponding to Table 2 of the main paper and \tablename~\ref{tab:exp_cg_180} of this material.
Four views per method are displayed with different strand visualization: Blender Cycles shading, 3D orientation, and random color.
The upper row shows a comparison with existing methods, and the lower row displays the ablation study.}
\label{fig:exp_cg_straight}
\end{figure*}

\begin{figure*}[htbp]\centering
\begin{tabular}{ccccc}
\includegraphics[width=.17\textwidth]{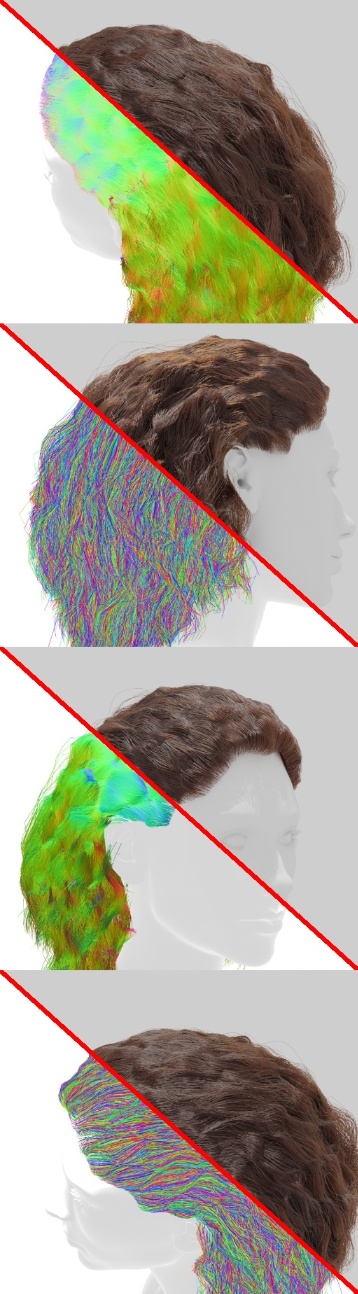} &
\includegraphics[width=.17\textwidth]{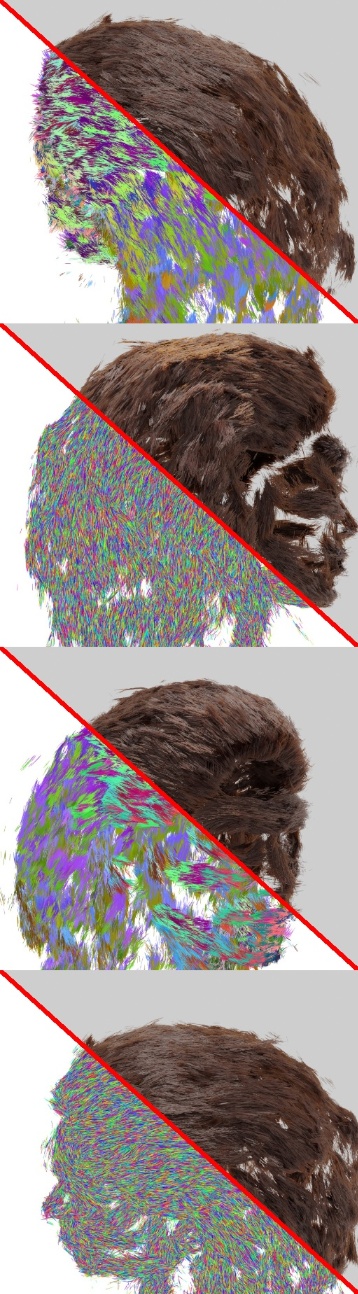} &
\includegraphics[width=.17\textwidth]{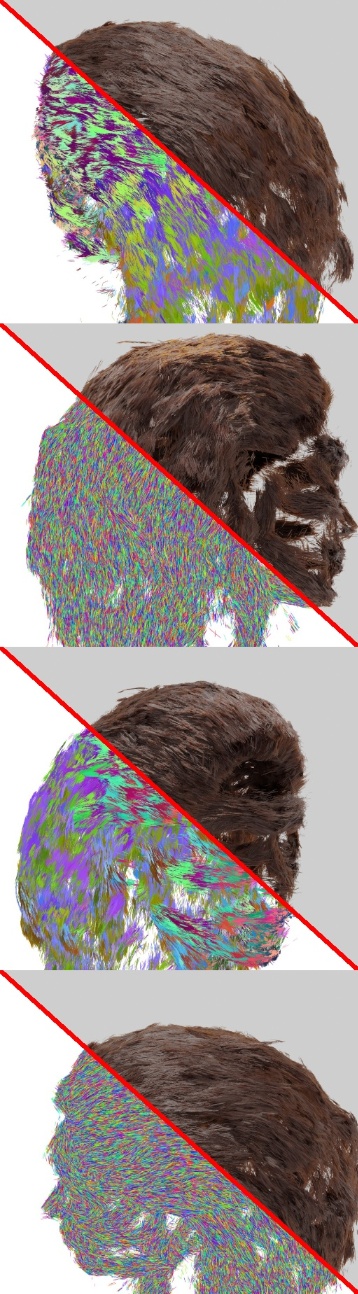} &
\includegraphics[width=.17\textwidth]{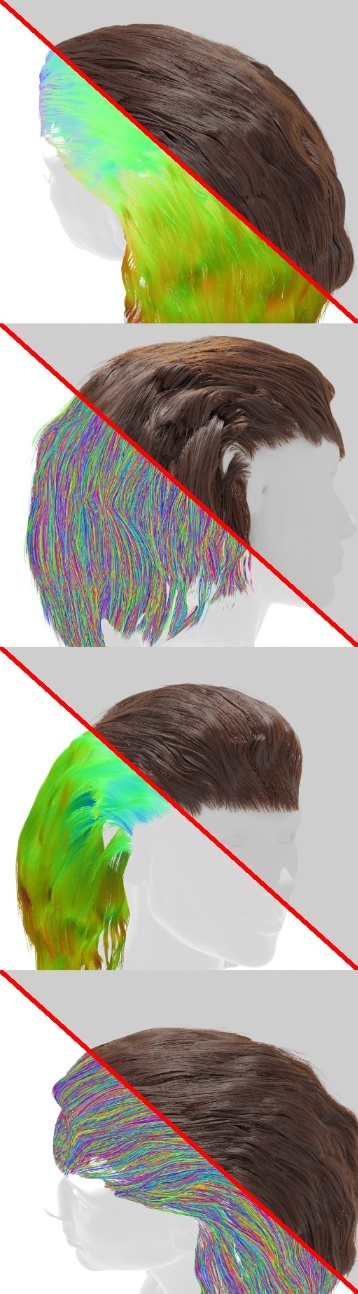} &
\includegraphics[width=.17\textwidth]{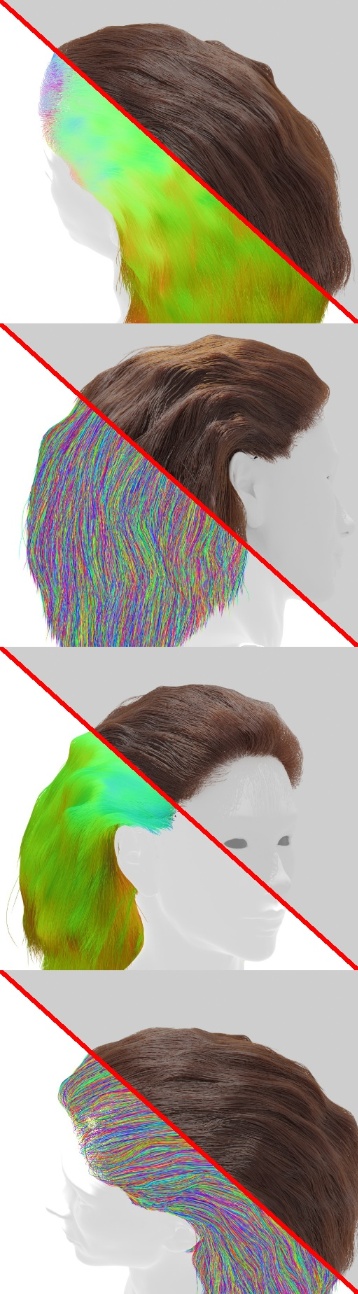} \\
\textbf{GT}  & \textbf{LPMVS} & \textbf{Strand Integration} & \textbf{NeuralHaircut} & \textbf{Ours} \\
\end{tabular}
\begin{tabular}{cccccc}
\includegraphics[width=.14\textwidth]{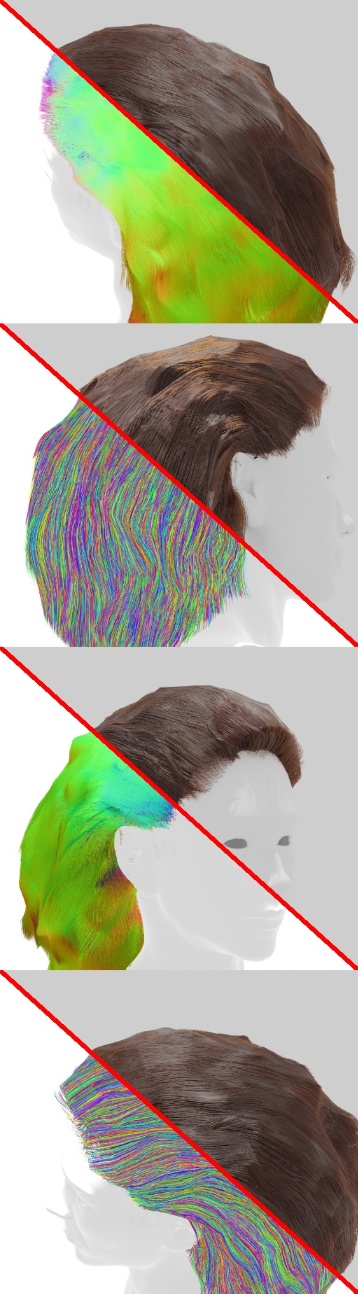} &
\includegraphics[width=.14\textwidth]{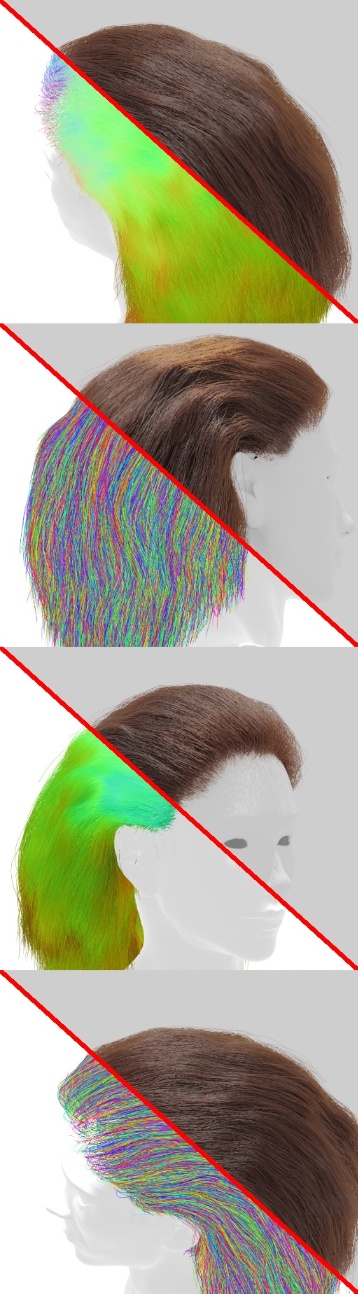} &
\includegraphics[width=.14\textwidth]{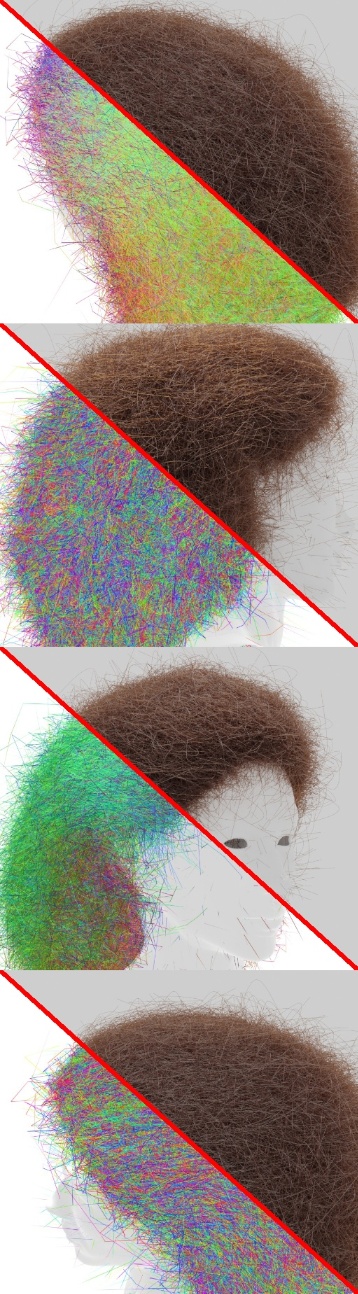} &
\includegraphics[width=.14\textwidth]{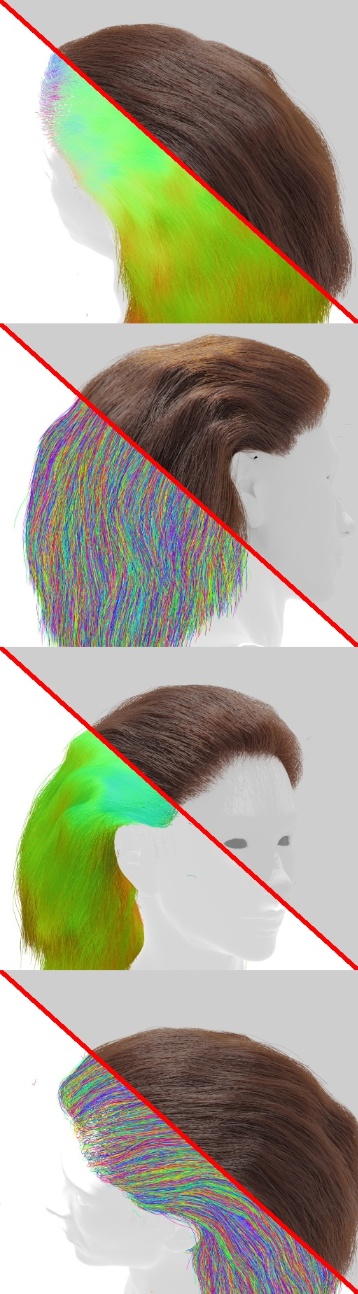} &
\includegraphics[width=.14\textwidth]{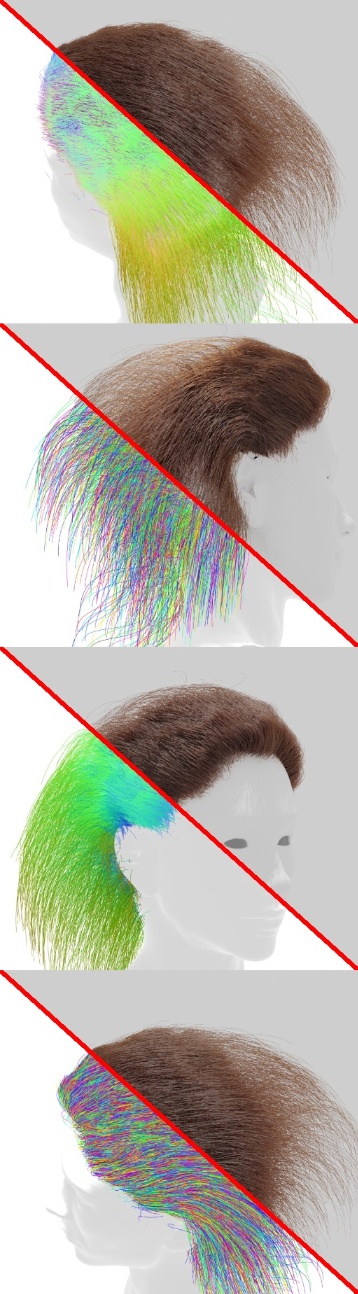} &
\includegraphics[width=.14\textwidth]{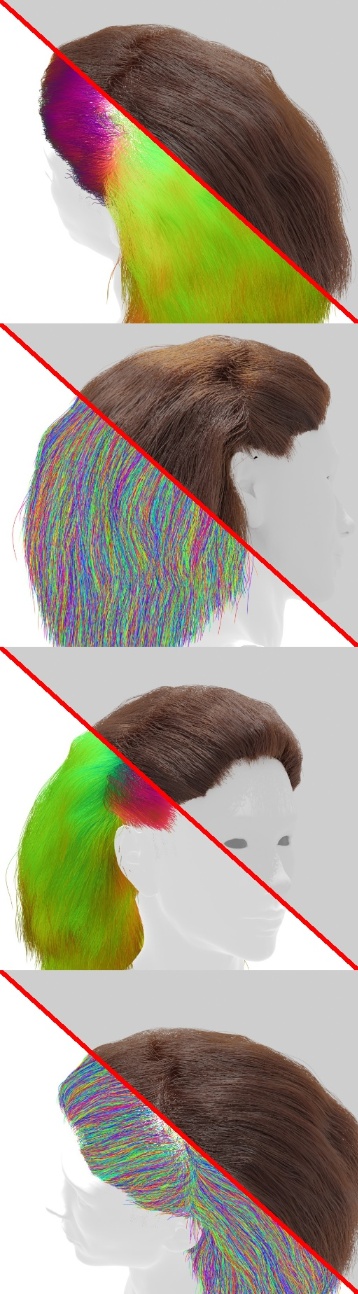} \\
\textbf{w/o DR} & \textbf{w/o guide opt.} & \textbf{w/o reparam.} & \textbf{w/o $\mathscr{N}$} & \textbf{w/o strand init.} & \textbf{w/o global opt.}\\
\end{tabular}
\vspace*{-2mm}
\caption{Qualitative evaluation on synthetic Curly Hair, corresponding to Table 2 of the main paper and \tablename~\ref{tab:exp_cg_180} of this material.
Four views per method are displayed with different strand visualization: Blender Cycles shading, 3D orientation, and random color.
The upper row shows a comparison with existing methods, and the lower row displays the ablation study.}
\label{fig:exp_cg_curly}
\end{figure*}

\begin{figure*}[htbp]\centering
\includegraphics[width=0.93\textwidth]{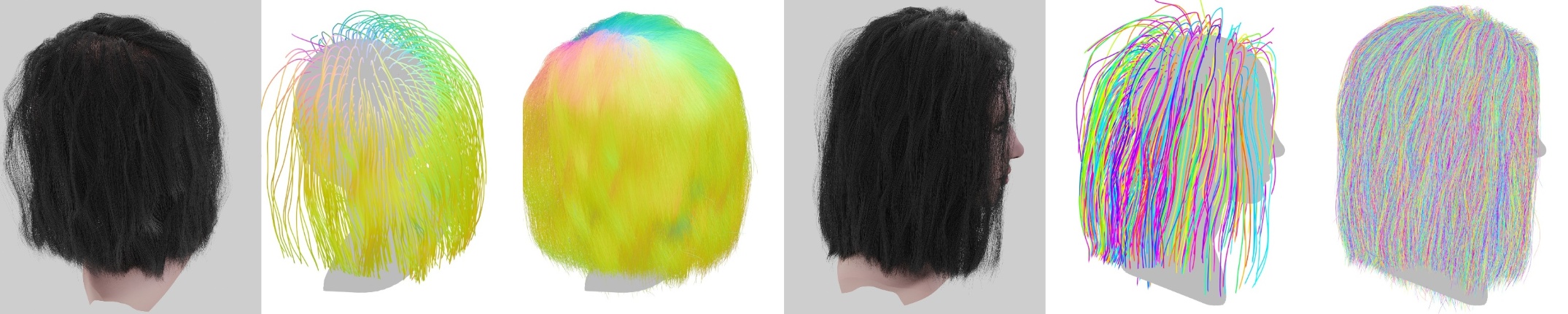}\\
\includegraphics[width=0.93\textwidth]{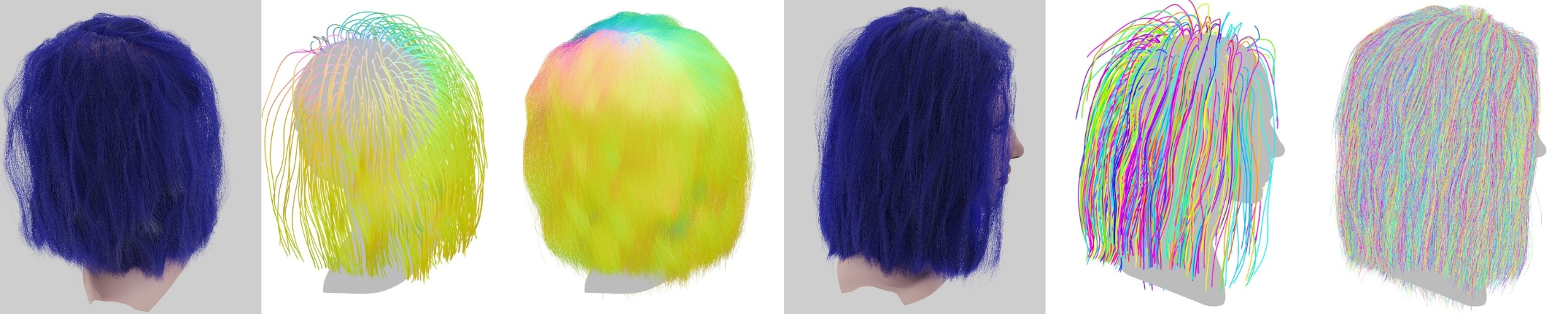}\\
\includegraphics[width=0.93\textwidth]{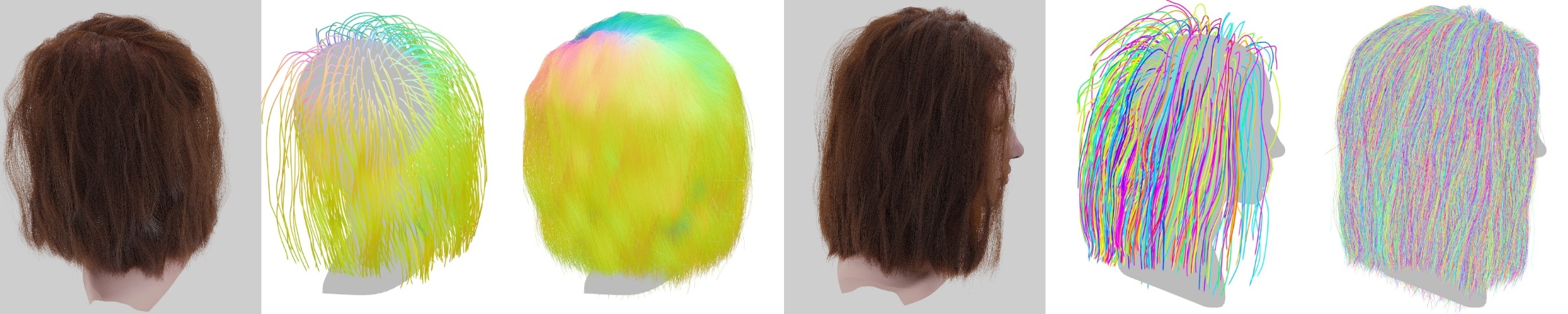}\\
\includegraphics[width=0.93\textwidth]{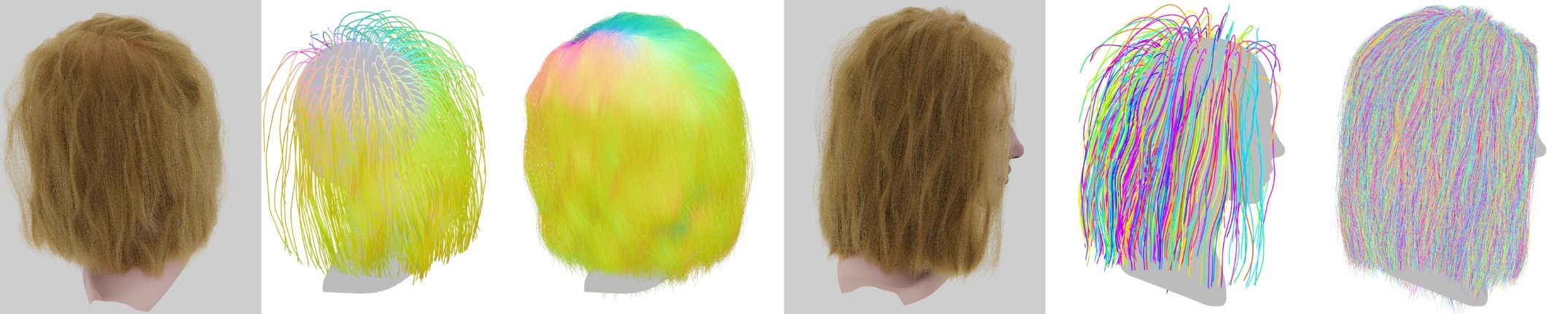}\\
\includegraphics[width=0.93\textwidth]{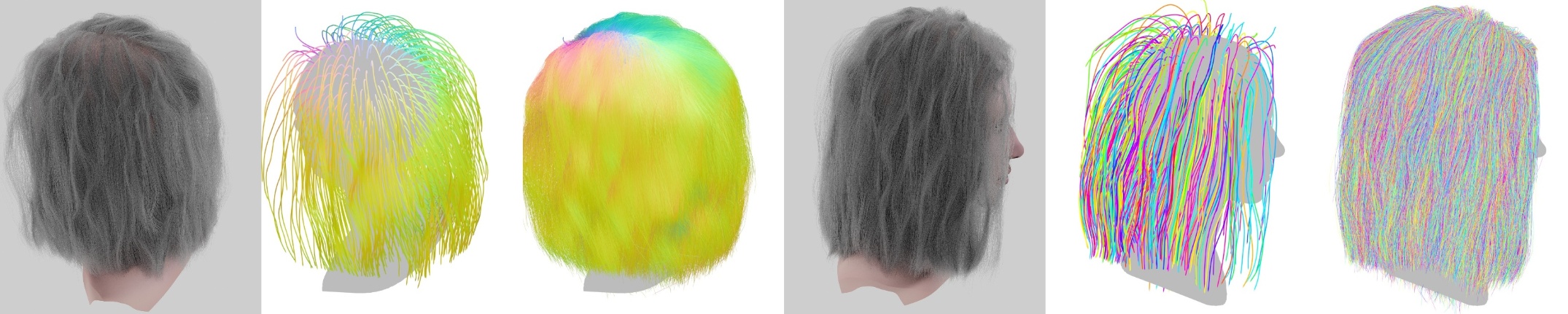}\\
\includegraphics[width=0.93\textwidth]{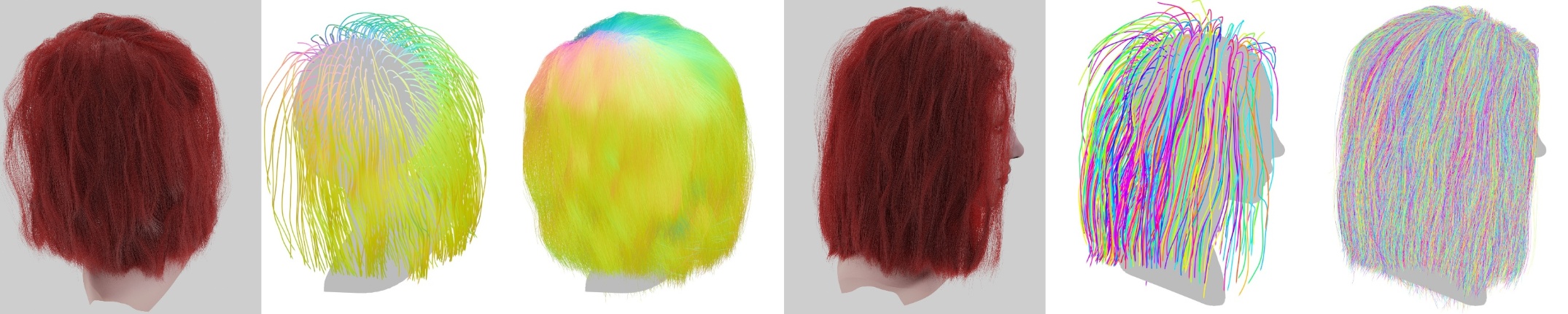}
\vspace*{-2mm}
\caption{Additional results on synthetic Wavy hair with different colors.
From top to bottom, black, blue, brown, gold, gray, and red colors are tested.
From right to left, top view GT, top view guide with 3D orientation, top view child with 3D orientation,
side view GT, side view guide with random color, and side view child with random color are shown.
The results are almost same and accurate, which indicate our method is less sensitive to the color of the input hairs. }
\label{fig:exp_cg_color}
\end{figure*}

\begin{table*}[htbp]\centering
\caption{Quantitative comparison with existing methods and ablation study on synthetic data tolerating \textbf{180\textdegree~ambiguity}, which have been used in the previous studies.
\textbf{P}, \textbf{R}, and \textbf{F1} denote precision, recall, and F1 score, respectively.
Higher is better.
The lower rows describe the values of our full pipeline and ours without individual modules.
w/o DR: DR optimization is not applied, and the initialized strands are evaluated.
w/o guide opt.: Child strands are optimized from the beginning of the DR step.
w/o reparam.: Reparameterization is disabled.
w/o $\mathscr{N}$: Only $\mathscr{N}$ is abandoned in the reparameterization.
w/o strand init.: Strands are initialized by straight lines parallel to the normal of the scalp.
w/o global opt.: Only gravity heuristic is applied to the initial 3D orientation, and 180\textdegree~ambiguity is accepted on the other steps.
}
\vspace*{-5mm}
\label{tab:exp_cg_180}
\small
\begin{tabular} {lwc{3.5mm}wc{3.5mm}wc{3.5mm}|wc{3.5mm}wc{3.5mm}wc{3.5mm}|wc{3.5mm}wc{3.5mm}wc{3.5mm}||wc{3.5mm}wc{3.5mm}wc{3.5mm}|wc{3.5mm}wc{3.5mm}wc{3.5mm}|wc{3.5mm}wc{3.5mm}wc{3.5mm}}\\
{}  & \multicolumn{9}{c||}{\textbf{Straight Hair}} & \multicolumn{9}{c}{\textbf{Curly Hair}}\\
\multicolumn{1}{r}{Threshold}  & \multicolumn{3}{c|}{1mm/10°} & \multicolumn{3}{c|}{2mm/20°} & \multicolumn{3}{c||}{3mm/30°} & \multicolumn{3}{c|}{1mm/10°} & \multicolumn{3}{c|}{2mm/20°} & \multicolumn{3}{c}{3mm/30°}\\
\multicolumn{1}{r}{Measure} & \textbf{P} & \textbf{R} & \textbf{F1} & \textbf{P} & \textbf{R} & \textbf{F1} & \textbf{P} & \textbf{R} & \textbf{F1} & \textbf{P} & \textbf{R} & \textbf{F1} & \textbf{P} & \textbf{R} & \textbf{F1} & \textbf{P} & \textbf{R} & \textbf{F1}\\
\hline
LPMVS \cite{Nam_2019_CVPR} & 61.0 & 37.1 & 46.1 & 81.1 & 62.8 & 70.8 & 87.5 & 76.0 & 81.4 & 36.9 & 8.1 & 13.3 & 65.6 & 18.6 & 28.9 & 74.0 & 28.1 & 40.7\\
Strand Integration \cite{maeda2023refinement} & \textbf{68.3} & 42.0 & 52.0 & 86.9 & 62.2 & 72.5 & 91.6 & 72.6 & 81.0 & \textbf{38.7} & 8.8 & 14.3 & 68.2 & 18.8 & 29.4 & 76.2 & 26.6 & 39.4\\
NeuralHaircut \cite{Sklyarova_2023_ICCV} & 50.3 & 15.0 & 23.1 & 76.4 & 29.3 & 42.4 & 85.9 & 38.6 & 53.3 & 21.0 & 3.9 & 6.6 & 58.6 & 14.7 & 23.6 & 80.8 & 28.1 & 41.7\\
\hline
Ours & 60.3 & 46.4 & 52.5 & 88.2 & 84.3 & 86.2 & \textbf{94.5} & 93.6 & \textbf{94.1} & 38.4 & 23.6 & \textbf{29.2} & \textbf{79.2} & 61.1 & \textbf{69.0} & \textbf{90.1} & 81.2 & \textbf{85.4}\\
Ours (w/o DR) & 65.4 & 41.8 & 51.0 & \textbf{88.6} & 78.8 & 83.4 & 93.3 & 88.6 & 90.9 & 22.3 & 15.9 & 18.6 & 60.0 & 56.8 & 58.4 & 77.5 & 83.0 & 80.1\\
Ours (w/o guide opt.) & 61.1 & 46.8 & \textbf{53.0} & 86.8 & 86.1 & \textbf{86.5} & 92.7 & 95.0 & 93.9 & 36.9 & 22.9 & 28.3 & 77.5 & 60.7 & 68.1 & 88.8 & 81.0 & 84.7\\
Ours (w/o reparam.) & 8.2 & 43.2 & 13.8 & 24.9 & \textbf{97.4} & 39.7 & 40.5 & \textbf{99.9} & 57.6 & 8.1 & \textbf{33.5} & 13.1 & 30.1 & \textbf{95.5} & 45.8 & 52.1 & \textbf{99.9} & 68.5\\
Ours (w/o $\mathscr{N}$) & 59.5 & \textbf{46.9} & 52.5 & 86.6 & 85.4 & 86.0 & 92.8 & 94.3 & 93.5 & 36.5 & 23.2 & 28.4 & 76.4 & 60.8 & 67.7 & 87.6 & 80.9 & 84.1\\
Ours (w/o strand init.) & 5.6 & 1.0 & 1.7 & 24.3 & 7.1 & 11.0 & 40.2 & 19.1 & 25.8 & 10.1 & 5.1 & 6.8 & 27.6 & 24.8 & 26.2 & 45.0 & 50.2 & 47.5\\
Ours (w/o global opt.) & 58.1 & 45.7 & 51.2 & 85.5 & 85.8 & 85.7 & 91.8 & 94.8 & 93.3 & 28.5 & 15.4 & 20.0 & 68.9 & 51.1 & 58.7 & 82.6 & 74.8 & 78.5\\
\end{tabular}
\end{table*}

\section{Additional Results}
We show additional results on synthetic and real data.

\subsection{Additional comparison and ablations on synthetic data}
Qualitative comparison with existing methods is shown in the upper rows of \figurename~\ref{fig:exp_cg_straight} and \figurename~\ref{fig:exp_cg_curly}.
These results correspond to the quantitative comparison in the main paper, Table 2.
The frontal scalp alignment of NeuralHaircut is not accurate, and for Straight Hair, NeuralHaircut confuses hair with the head.
The results of LPMVS and Strand Integration are almost similar, showing many short strands, inconsistent 3D orientation, and no distinction between head and hair.
Our full pipeline shows better precision for both cases.

The lower rows of \figurename~\ref{fig:exp_cg_straight} and \figurename~\ref{fig:exp_cg_curly} visualize the ablation study.
Even in \textbf{w/o DR}, the outline is well estimated, but it leaves room for improvement in fine details.
In \textbf{w/o guide opt.}, child strands become too smoother.
In \textbf{w/o reparam.}, hair moves freely to improve recall value but causes noised shapes due to a lack of regularization.
In \textbf{w/o $\mathscr{N}$}, individual strands move freely.
In the case of curly hair, the strands are easily entangled in close observation and quickly become stuck in the local minima.
The same behavior was observed for straight hair, but the uniformity of the hair flow had less negative impact on the numerical evaluation.
In \textbf{w/o strand init.}, because initial strands are far from actual, hair growing stops in the middle of a stretch.
Note that increasing the learning rate may improve hair growth, but shape collapse may also happen.
In \textbf{w/o global opt.}, the boundary condition becomes heuristic, and all 3D orientations on surface are treated as downward facing, resulting in partially wrong, opposite guide hair flow.
Even if the DR process allows 180\textdegree~ambiguity, it will never be the correct orientation because the initial value will settle to the local minima of the direction it is facing.
Turning off the individual modules causes reasonable degradation, which indicates that the effectiveness of each component of our pipeline is validated.

Next, \figurename~\ref{fig:exp_cg_color} illuminates the robustness against hair color.
Hair color mainly affects the former part of our pipeline, such as raw mesh reconstruction and 2D/3D orientation estimation.
Six colors with the same hair geometry were tested, and our method reconstructed similar, accurate strands for all colors, which indicates that our method can handle various hair colors.

In Table 2 of the main paper, a quantitative comparison was performed in 360\textdegree~range to evaluate absolute hair flow with synthetic data \cite{Yuksel2009}.
To align with the criteria used in the previous papers \cite{Nam_2019_CVPR, rosu2022neuralstrands, Sklyarova_2023_ICCV}, we show the evaluation tolerating 180\textdegree~ambiguity in \tablename~\ref{tab:exp_cg_180}.
Note that only evaluation metrics were updated, and the same geometries shown in \figurename~\ref{fig:exp_cg_straight} and \figurename~\ref{fig:exp_cg_curly} were used.
The values of LPMVS and Strand Integration become better because they are not aware of 180\textdegree~ambiguity.
NeuralHaircut keeps the most values because it estimates the correct absolute hair direction in this case.
Ours is still best in most values.

\begin{figure*}[htb]\centering
\includegraphics[width=0.9\textwidth]{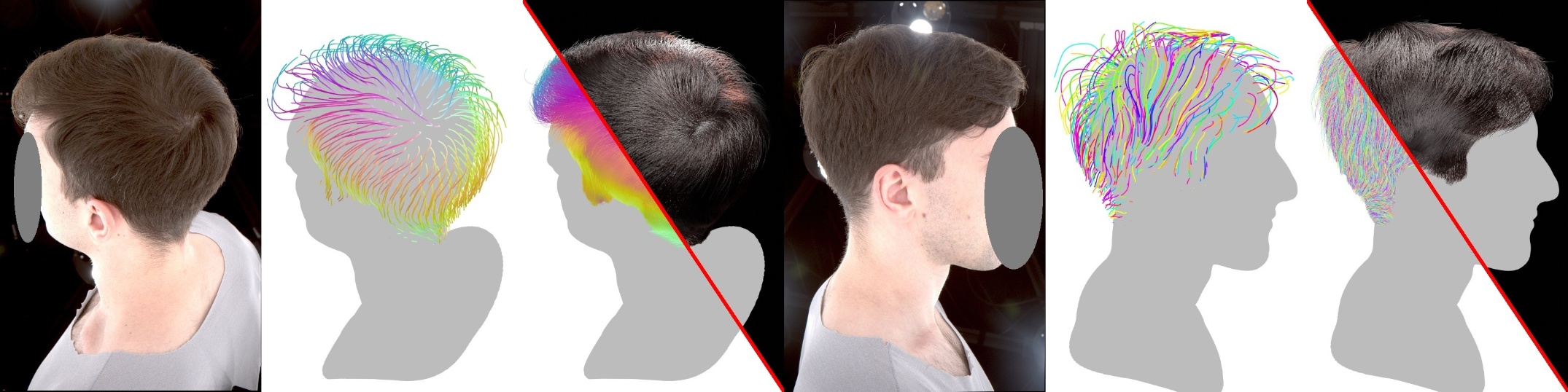} \\
\includegraphics[width=0.9\textwidth]{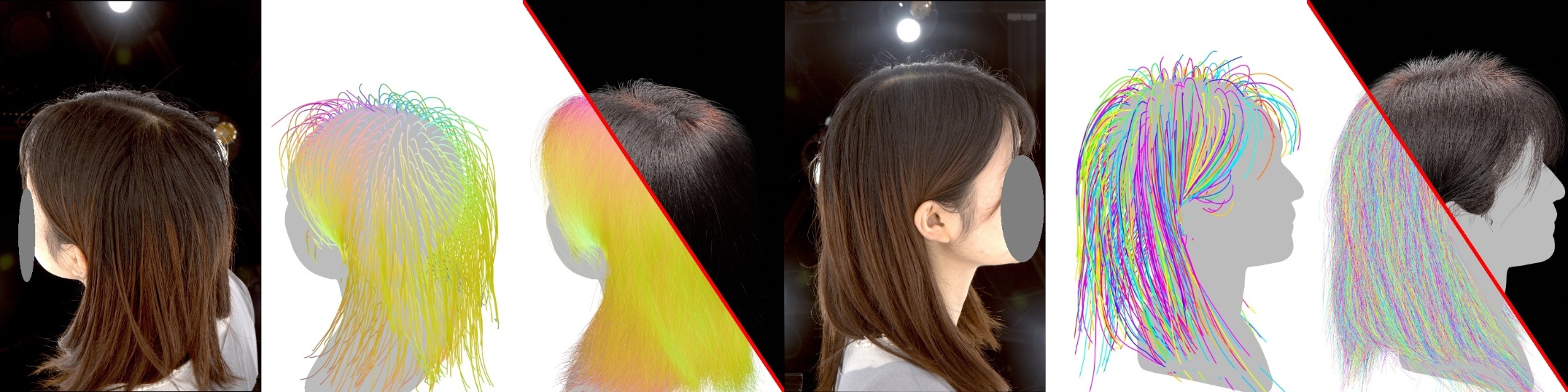} \\
\includegraphics[width=0.9\textwidth]{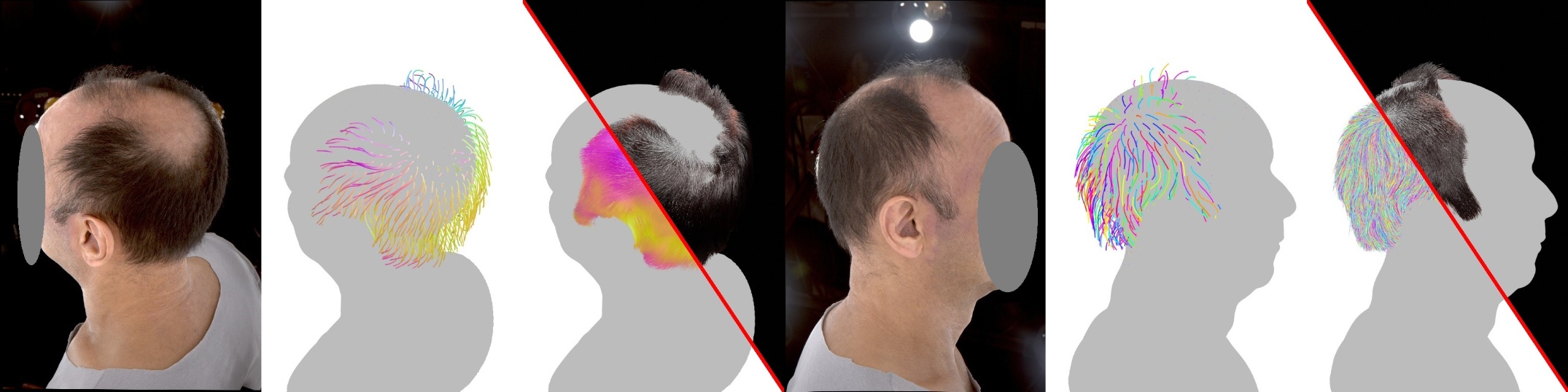} \\
\includegraphics[width=0.9\textwidth]{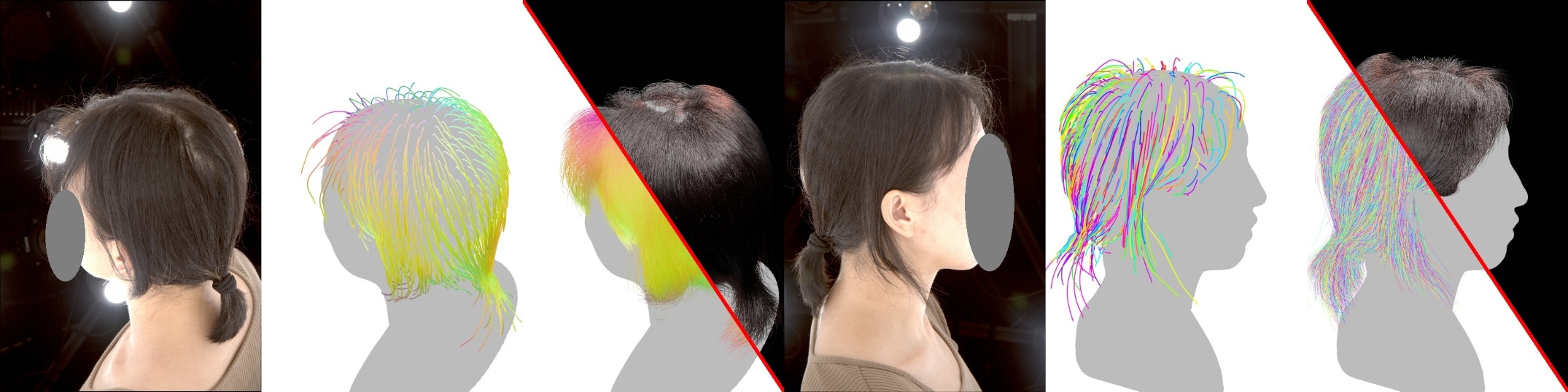} \\
\vspace*{-2mm}
\caption{Additional results on studio data 1 / 2.
From left to right, top view image, top view guide with 3D orientation, top view child with Blender Cycles shading and 3D orientation,
side view image, side view guide with random color, and side view child with Blender Cycles shading and random color are shown.
From top to bottom, the subjects with short hair, long hair, half bald head, and short-tied hair are displayed.
Our method enables realistic reconstruction for a wide range of hairstyles in the wild.}
\label{fig:exp_lc3}
\end{figure*}

\begin{figure*}[htb]\centering
\includegraphics[width=0.9\textwidth]{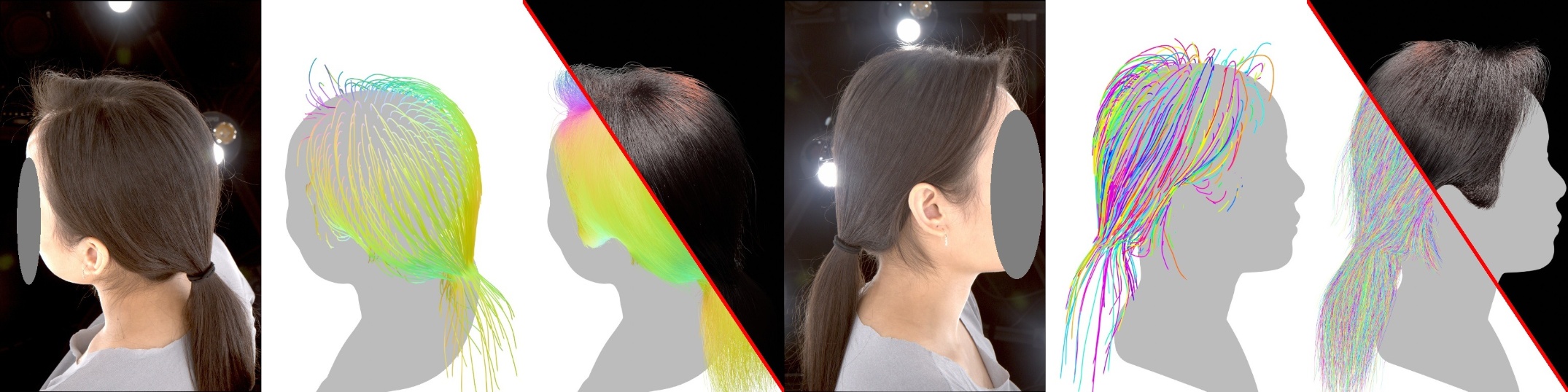} \\
\includegraphics[width=0.9\textwidth]{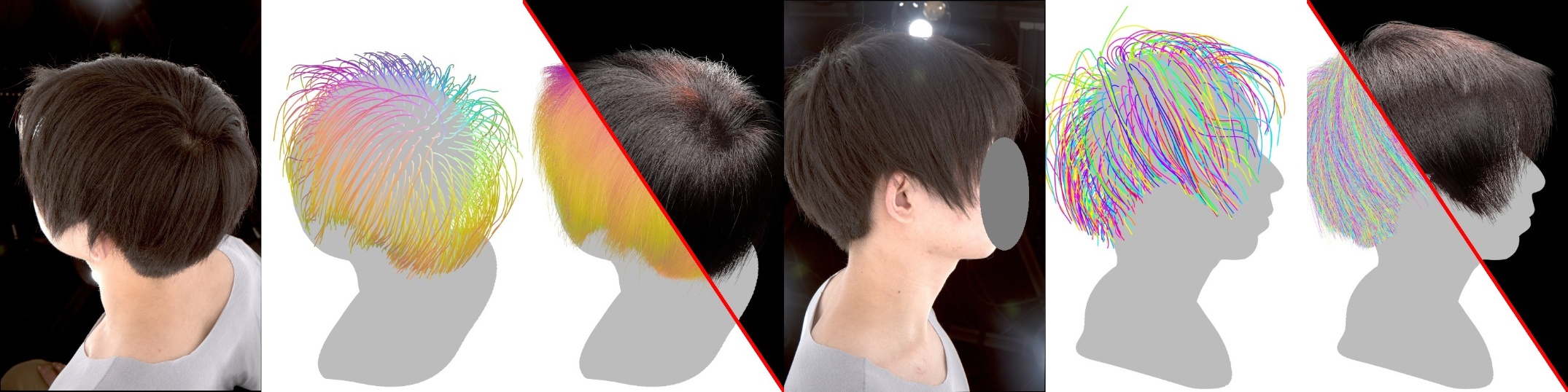} \\
\includegraphics[width=0.9\textwidth]{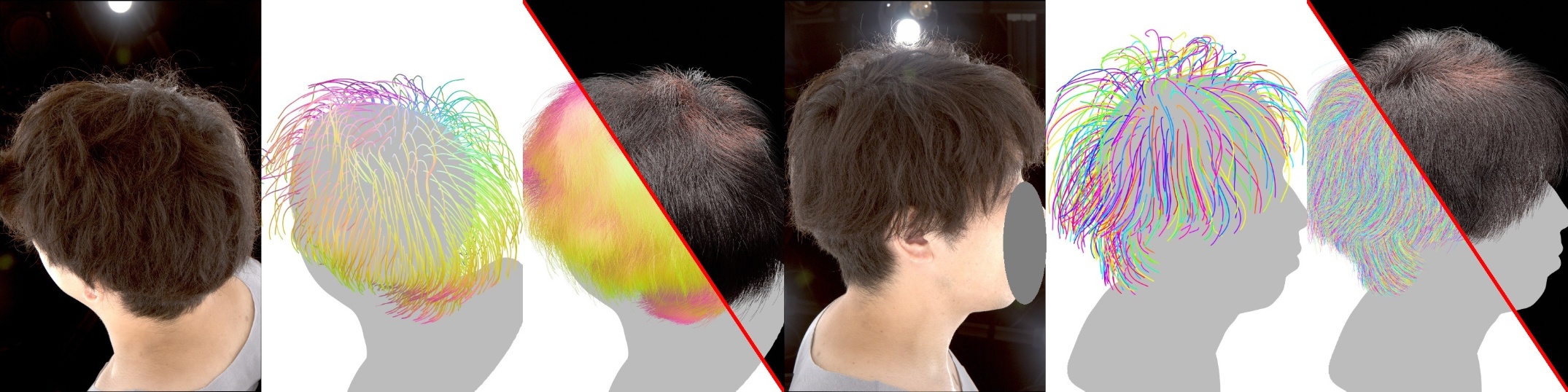} \\
\includegraphics[width=0.9\textwidth]{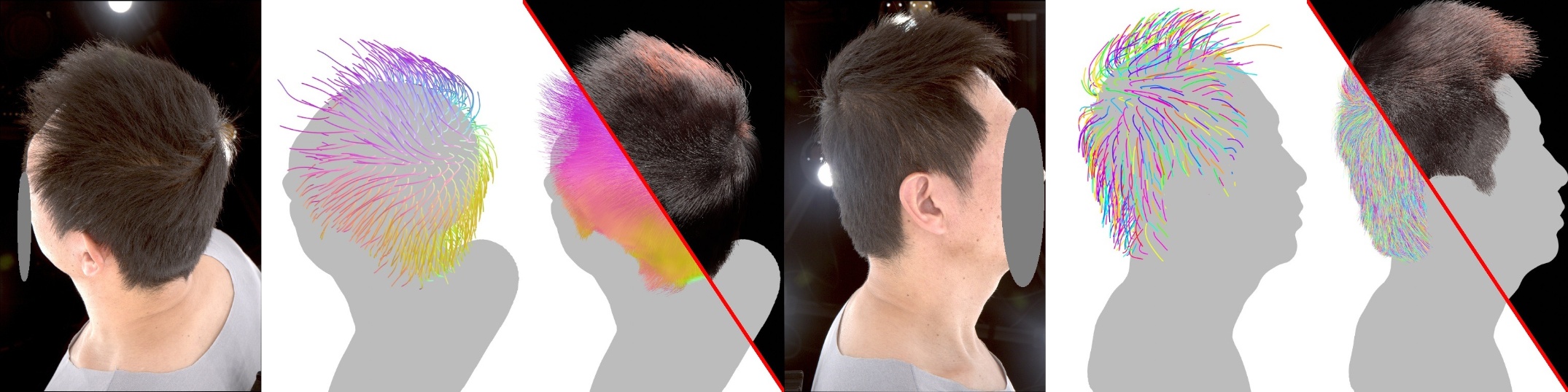} \\
\vspace*{-2mm}
\caption{Additional results on studio data 2 / 2.
From left to right, top view image, top view guide with 3D orientation, top view child with Blender Cycles shading and 3D orientation,
side view image, side view guide with random color, and side view child with Blender Cycles shading and random color are shown.
From top to bottom, the subjects with long-tied hair, short straight hair, wavy hair, stiff hair that is not squished by gravity are displayed.
Our method enables realistic reconstruction for a wide range of hairstyles in the wild.}
\label{fig:exp_lc4}
\end{figure*}

\begin{figure*}[htb]
\begin{tabular}{ccccccc}
\raisebox{+0mm}{\rotatebox{90}{\scriptsize{\textbf{NeuralHaircut}}}} &
\includegraphics[width=.135\textwidth]{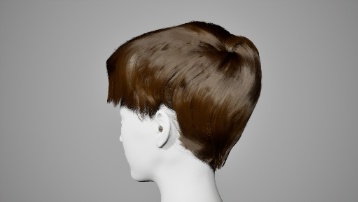} &
\includegraphics[width=.135\textwidth]{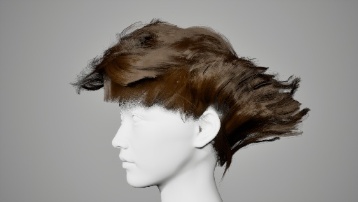} &
\includegraphics[width=.135\textwidth]{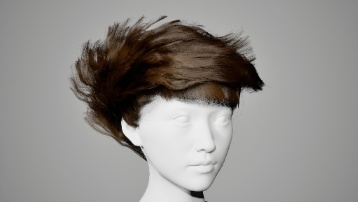} &
\includegraphics[width=.135\textwidth]{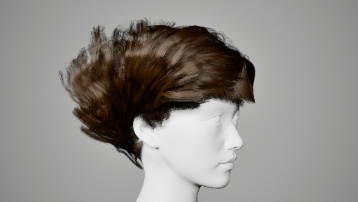} &
\includegraphics[width=.135\textwidth]{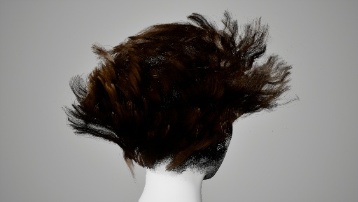} &
\includegraphics[width=.135\textwidth]{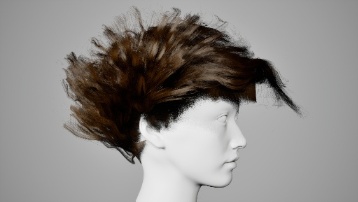} \\
\raisebox{+3mm}{\rotatebox{90}{\scriptsize{\textbf{Ours}}}} &
\includegraphics[width=.135\textwidth]{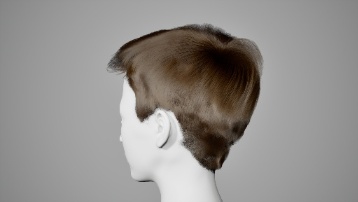} &
\includegraphics[width=.135\textwidth]{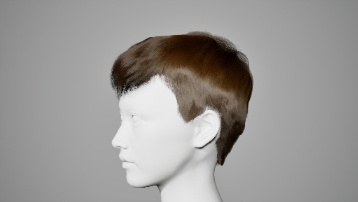} &
\includegraphics[width=.135\textwidth]{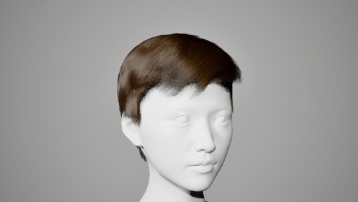} &
\includegraphics[width=.135\textwidth]{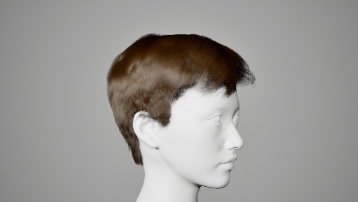} &
\includegraphics[width=.135\textwidth]{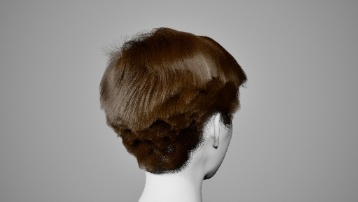} &
\includegraphics[width=.135\textwidth]{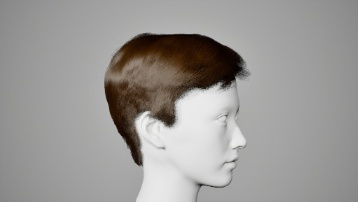} \\
\raisebox{+0mm}{\rotatebox{90}{\scriptsize{\textbf{NeuralHaircut}}}} &
\includegraphics[width=.135\textwidth]{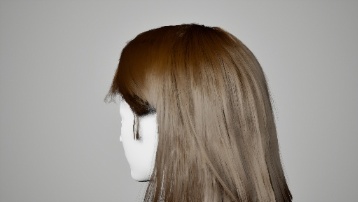} &
\includegraphics[width=.135\textwidth]{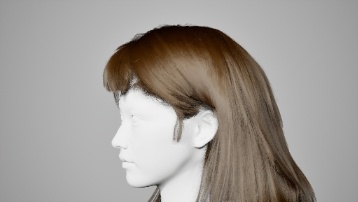} &
\includegraphics[width=.135\textwidth]{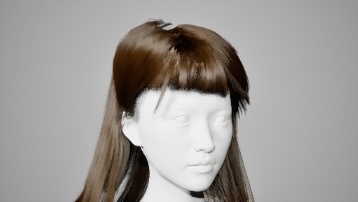} &
\includegraphics[width=.135\textwidth]{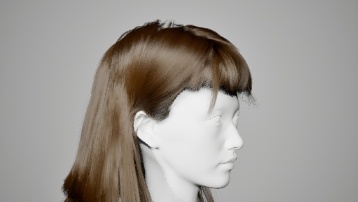} &
\includegraphics[width=.135\textwidth]{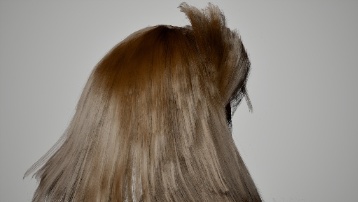} &
\includegraphics[width=.135\textwidth]{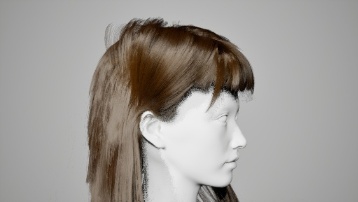} \\
\raisebox{+3mm}{\rotatebox{90}{\scriptsize{\textbf{Ours}}}} &
\includegraphics[width=.135\textwidth]{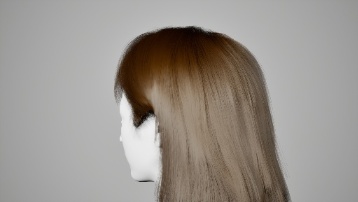} &
\includegraphics[width=.135\textwidth]{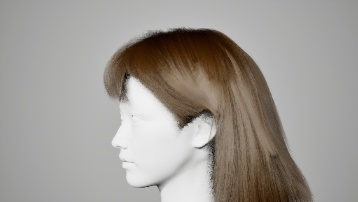} &
\includegraphics[width=.135\textwidth]{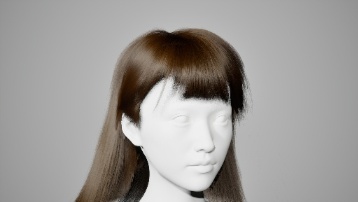} &
\includegraphics[width=.135\textwidth]{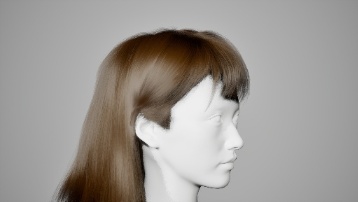} &
\includegraphics[width=.135\textwidth]{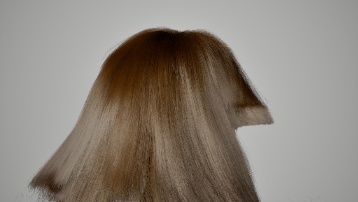} &
\includegraphics[width=.135\textwidth]{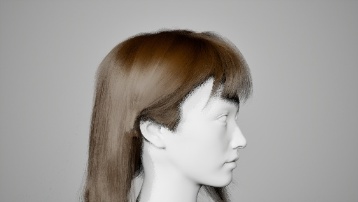} \\
\raisebox{+0mm}{\rotatebox{90}{\scriptsize{\textbf{NeuralHaircut}}}} &
\includegraphics[width=.135\textwidth]{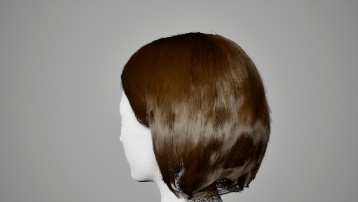} &
\includegraphics[width=.135\textwidth]{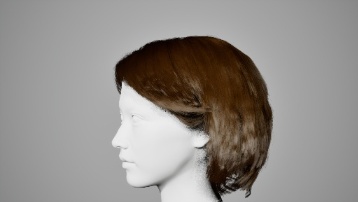} &
\includegraphics[width=.135\textwidth]{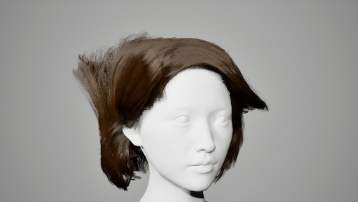} &
\includegraphics[width=.135\textwidth]{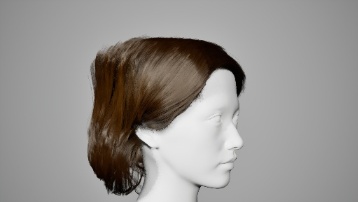} &
\includegraphics[width=.135\textwidth]{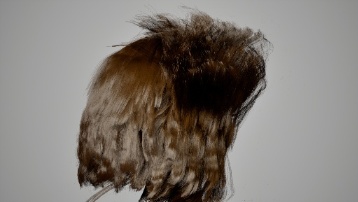} &
\includegraphics[width=.135\textwidth]{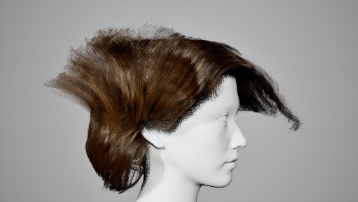} \\
\raisebox{+3mm}{\rotatebox{90}{\scriptsize{\textbf{Ours}}}} &
\includegraphics[width=.135\textwidth]{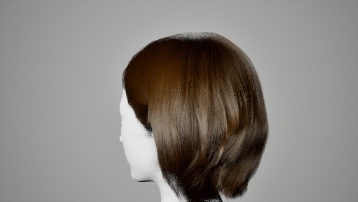} &
\includegraphics[width=.135\textwidth]{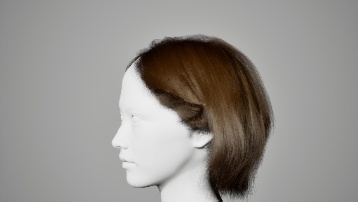} &
\includegraphics[width=.135\textwidth]{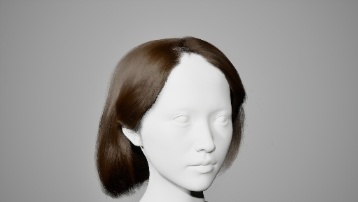} &
\includegraphics[width=.135\textwidth]{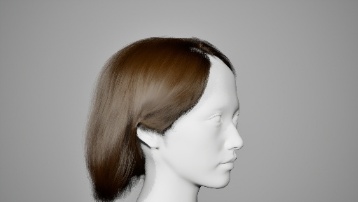} &
\includegraphics[width=.135\textwidth]{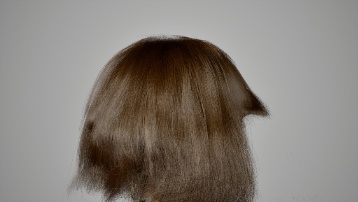} &
\includegraphics[width=.135\textwidth]{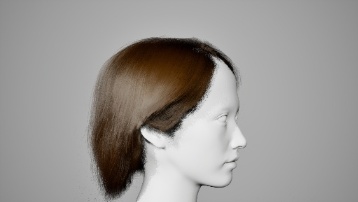} \\
&\textbf{Start}  & \textbf{$\Rightarrow$} & \textbf{$\Rightarrow$} & \textbf{$\Rightarrow$} & \textbf{$\Rightarrow$} & \textbf{End} \\
\end{tabular}
\caption{Comparison of physics simulation with head motion.
Full sequences are available in the supplemental video.
Starting from the reconstructed strands, gravity, hair stiffness and head rotation are applied to NeuralHaircut and ours.
On the top two rows, the subjects of studio data in Figures 1 and 10 of the main paper are shown.
The bottom row shows the subject of H3DS in the top row of Figure 7 of the main paper.
Thanks to the correct hair growing direction, our hair shows more natural behavior under strong head movement.
The original scalp shapes and hair root positions are kept while the head model is replaced for privacy protection.}
\label{fig:exp_physics_motion}
\end{figure*}

\begin{figure*}[htb]\centering
\includegraphics[width=0.93\textwidth]{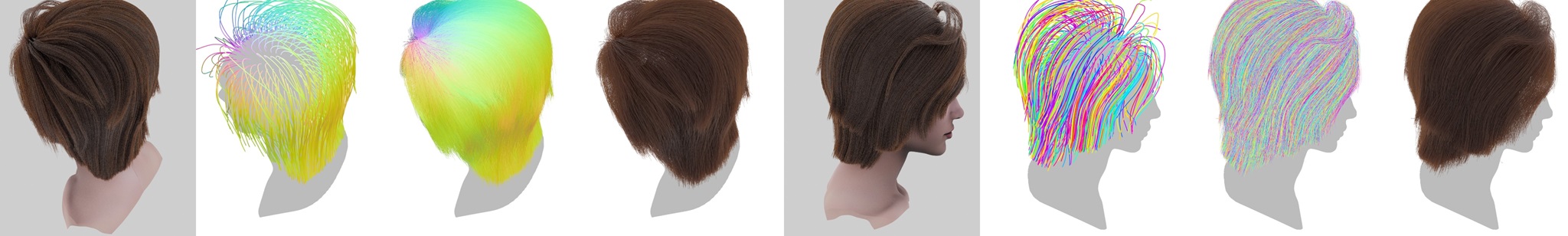}\\
\includegraphics[width=0.93\textwidth]{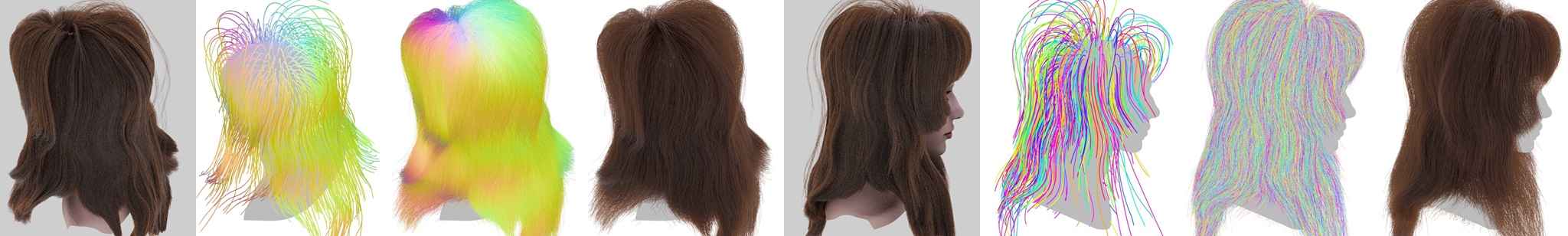}\\
\includegraphics[width=0.93\textwidth]{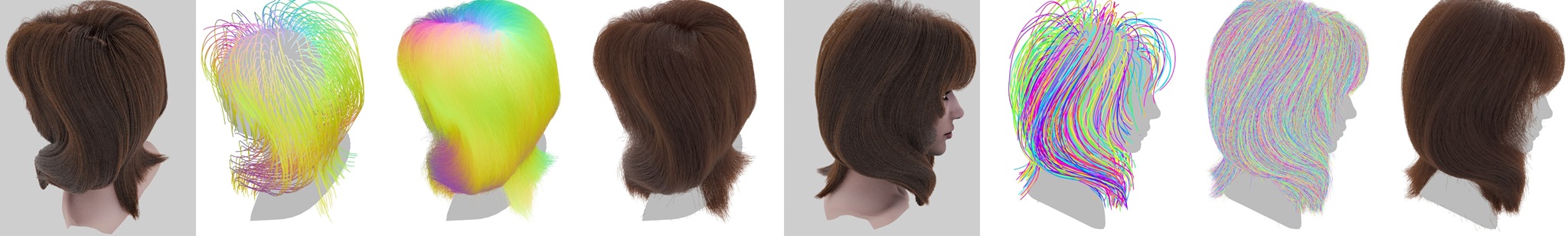}\\
\caption{Additional results on USC-HairSalon \cite{hu2015database}.
From right to left, top view GT, top view guide with 3D orientation, top view child with 3D orientation, top view shaded child,
side view GT, side view guide with random color, side view child with random color, and side view shaded child are shown.
Our method successfully handles various hairstyles.}
\label{fig:exp_usc}
\end{figure*}

\begin{figure*}[htb]\centering
\begin{tabular}{cccc}
\centering
\includegraphics[width=.175\textwidth]{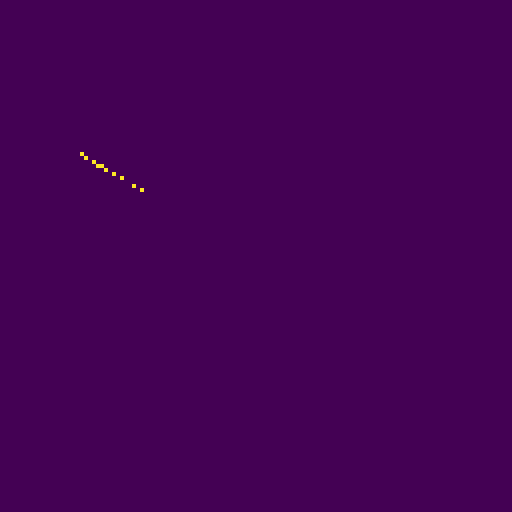} &
\includegraphics[width=.175\textwidth]{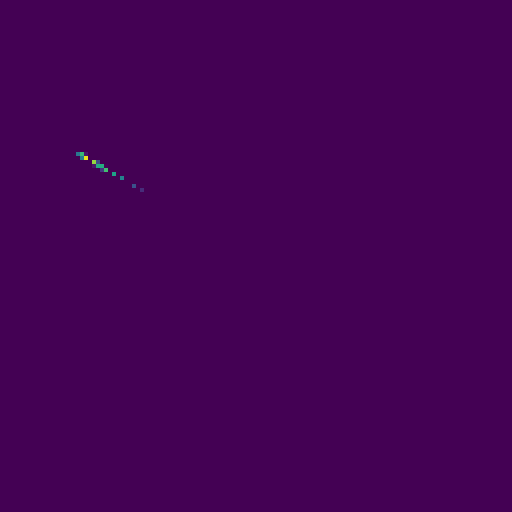} &
\includegraphics[width=.175\textwidth]{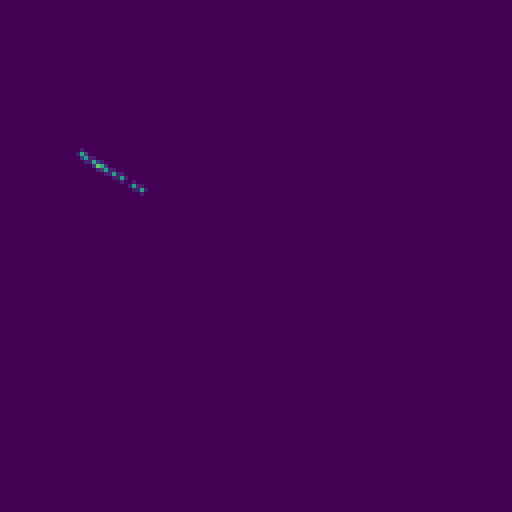} &
\includegraphics[width=.175\textwidth]{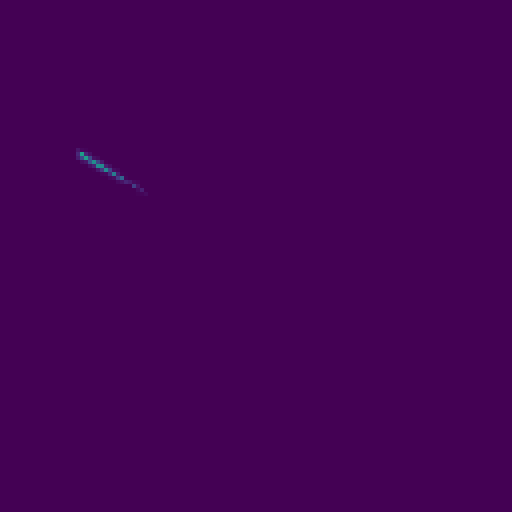} \\
\includegraphics[width=.175\textwidth]{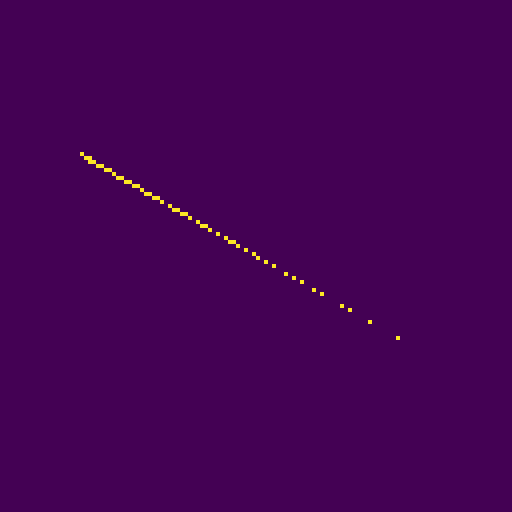} &
\includegraphics[width=.175\textwidth]{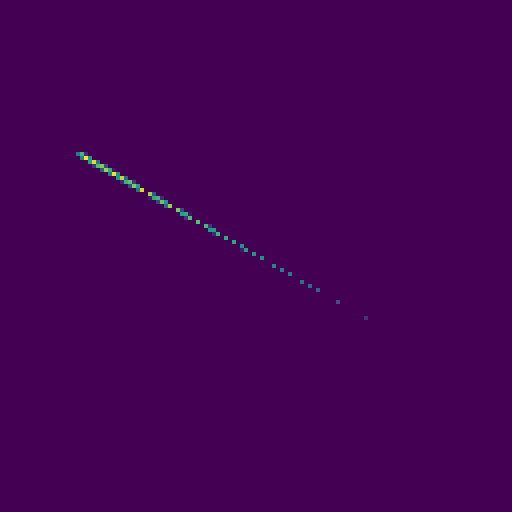} &
\includegraphics[width=.175\textwidth]{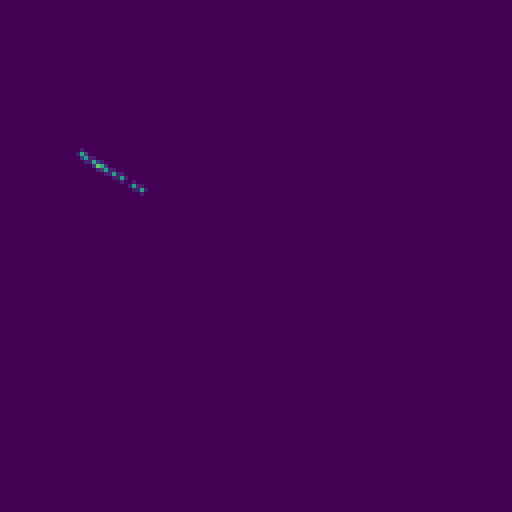} &
\includegraphics[width=.175\textwidth]{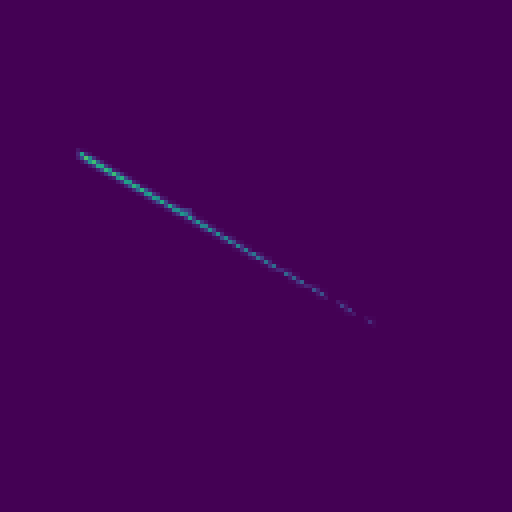} \\
\multicolumn{4}{c}{(a) 1.0 pixel width} \\
\includegraphics[width=.175\textwidth]{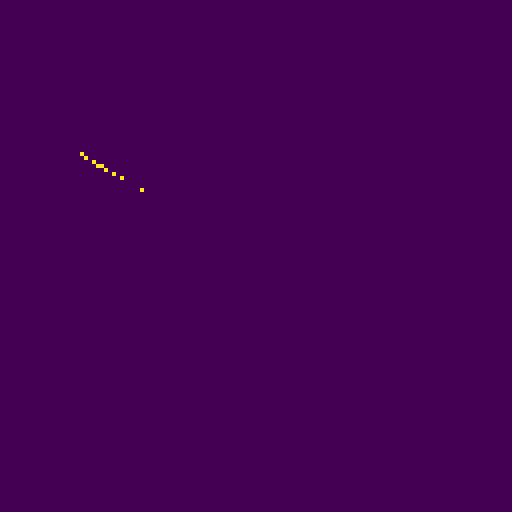} &
\includegraphics[width=.175\textwidth]{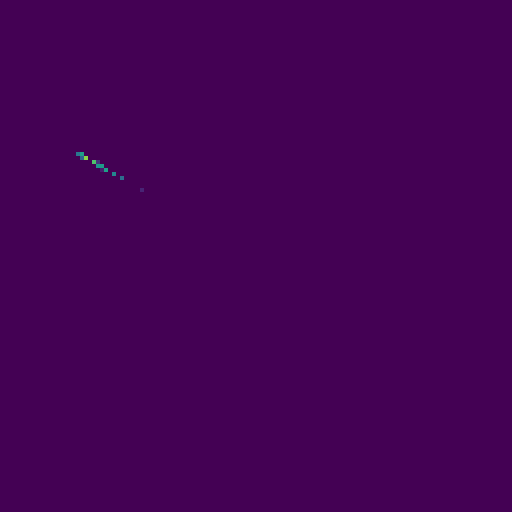} &
\includegraphics[width=.175\textwidth]{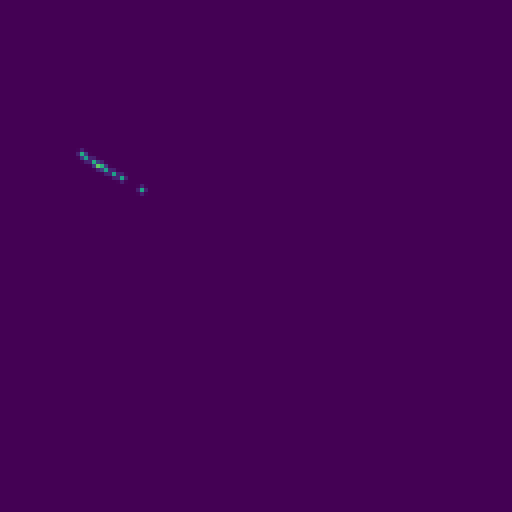} &
\includegraphics[width=.175\textwidth]{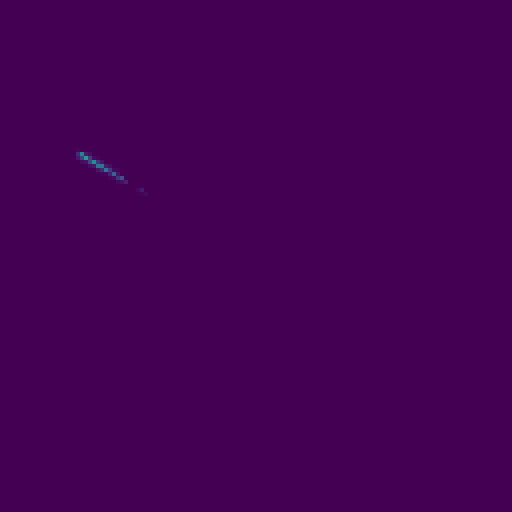} \\
\includegraphics[width=.175\textwidth]{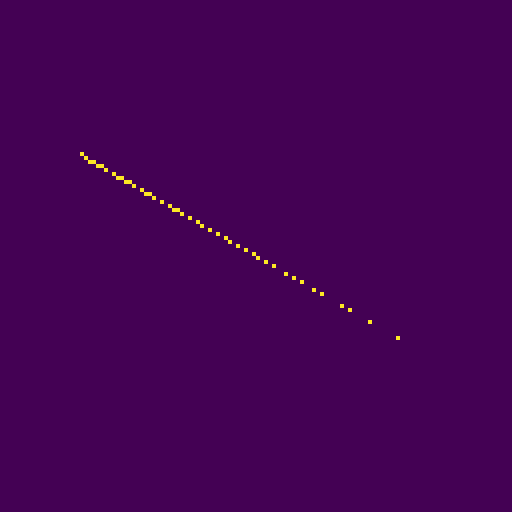} &
\includegraphics[width=.175\textwidth]{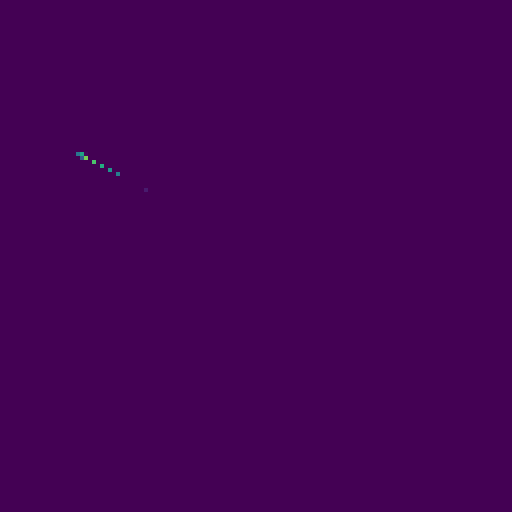} &
\includegraphics[width=.175\textwidth]{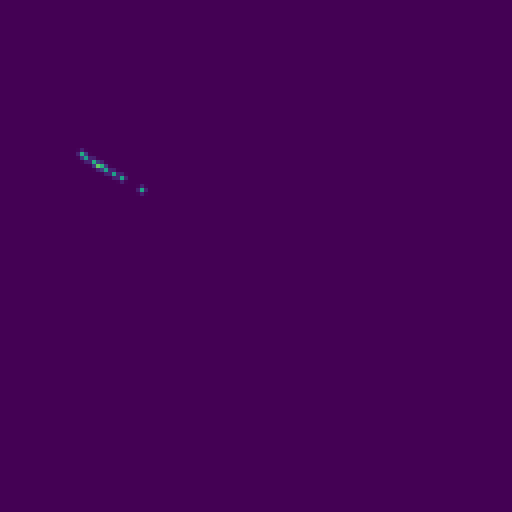} &
\includegraphics[width=.175\textwidth]{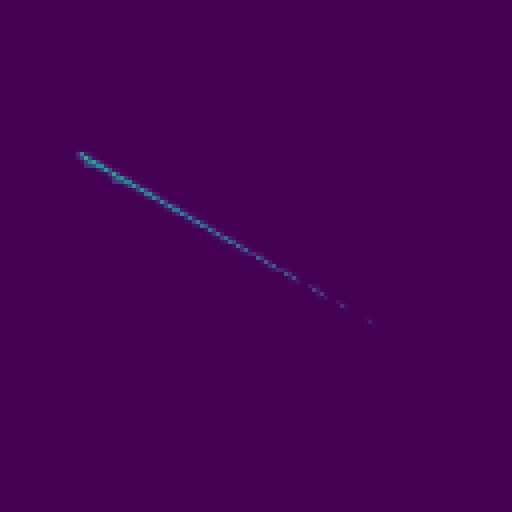} \\
\multicolumn{4}{c}{(b) 0.8 pixel width} \\
\includegraphics[width=.175\textwidth]{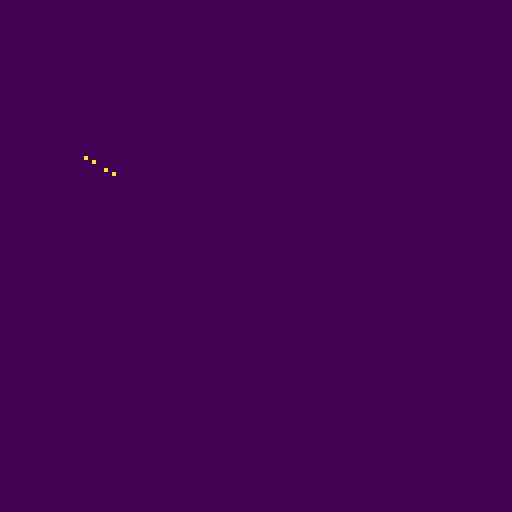} &
\includegraphics[width=.175\textwidth]{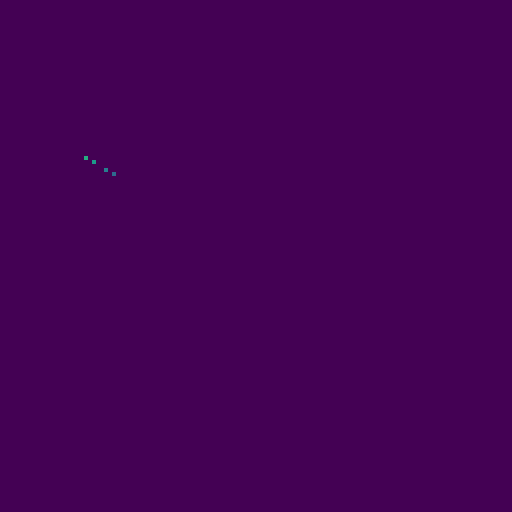} &
\includegraphics[width=.175\textwidth]{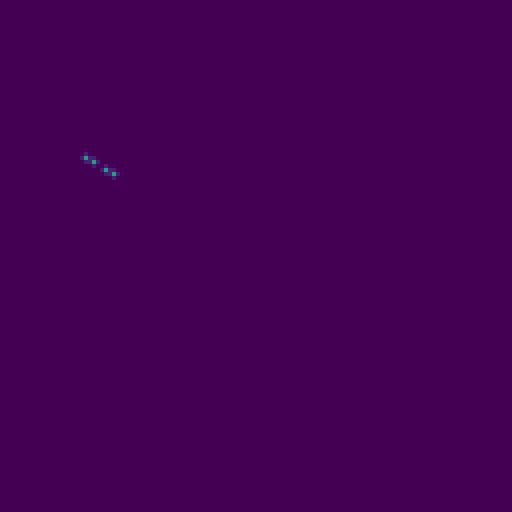} &
\includegraphics[width=.175\textwidth]{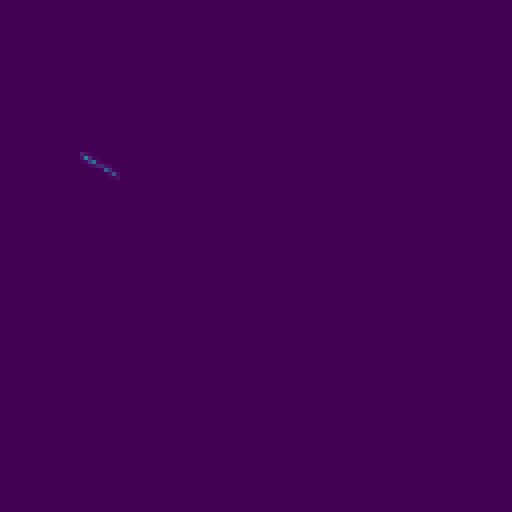} \\
\includegraphics[width=.175\textwidth]{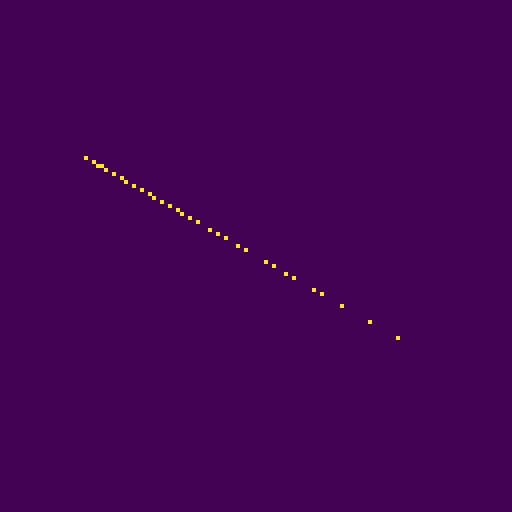} &
\includegraphics[width=.175\textwidth]{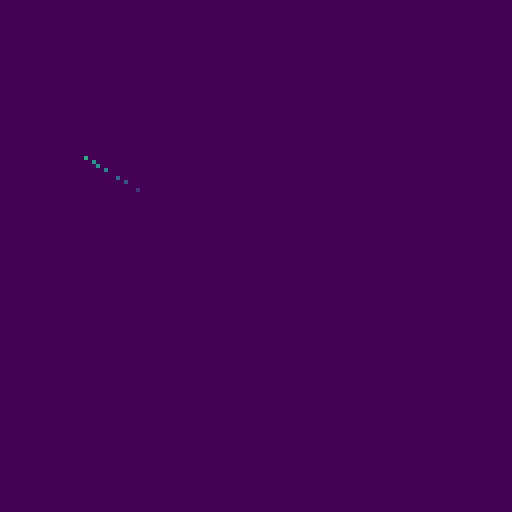} &
\includegraphics[width=.175\textwidth]{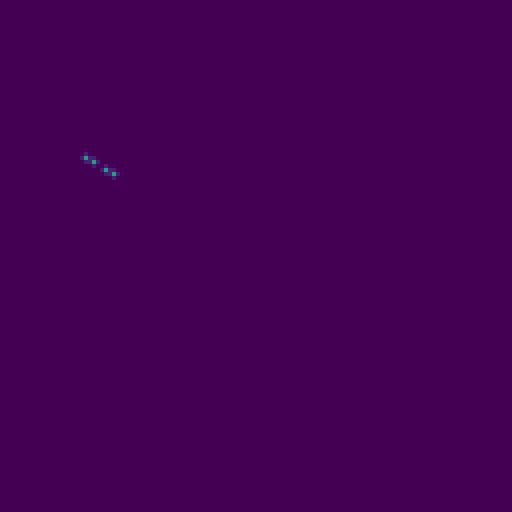} &
\includegraphics[width=.175\textwidth]{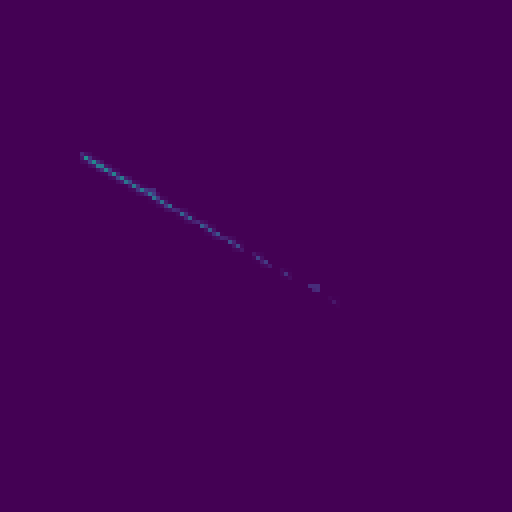} \\
\multicolumn{4}{c}{(c) 0.6 pixel width} \\
\textbf{GT} init / target & \textbf{nvdiffrast} \cite{Laine2020diffrast} init / final & \textbf{splatting} \cite{cole2021differentiable} init / final& \textbf{Ours} init / final
\end{tabular}
\caption{Qualitative validation of our AA. Digital zoom is recommended.
The fixed root vertex is on the top left, and the tip vertex to be optimized is placed on the lower left of the root.
Initial and final strands of various pixel widths are visualized.
Our broader gradient compared to other AAs demonstrates that the strands grow even when the width is much narrower than one pixel.
}
\label{fig:aa_validation_fig}
\end{figure*}

\begin{figure*}[htb]\centering
\begin{minipage}[t]{0.30\textwidth}
\centering
\includegraphics[width=1.\textwidth]{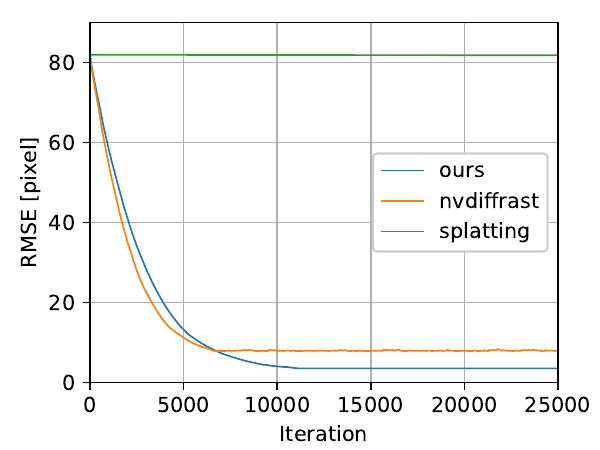}
\subcaption{1.0 pixel width}
\end{minipage}
\begin{minipage}[t]{0.30\textwidth}
\centering
\includegraphics[width=1.\textwidth]{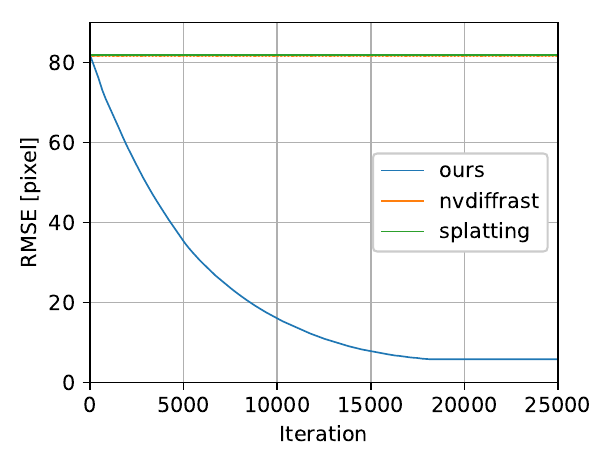}
\subcaption{0.8 pixel width}
\end{minipage}
\begin{minipage}[t]{0.30\textwidth}
\centering
\includegraphics[width=1.\textwidth]{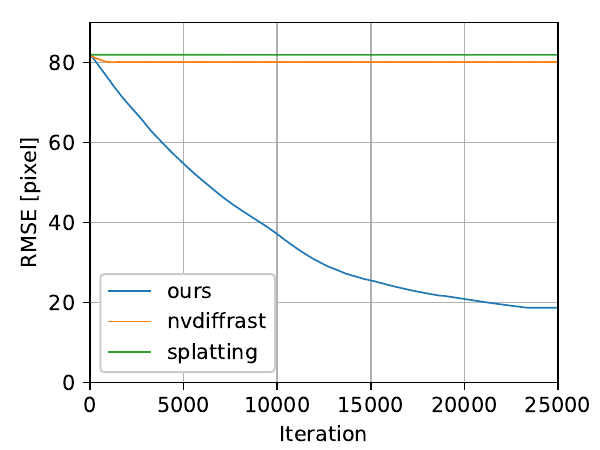}
\subcaption{0.6 pixel width}
\end{minipage}
\caption{Quantitative validation of our AA.
Error curves of various pixel widths are shown.
The proposed method reduces errors in the long term, whereas the existing method is stuck in the early stages of optimization.}
\label{fig:aa_validation_graph}
\end{figure*}

\begin{figure}[htb]\centering
\begin{minipage}[t]{0.9\linewidth}
\adjincludegraphics[width=1.\linewidth, trim={0 {.1\height} 0 {.1\height}},clip]{{\dataroot}HairPaper2023/figures/cvpr_drawing/gt.jpg}
\subcaption{GT}
\adjincludegraphics[width=1.\linewidth, trim={0 {.1\height} 0 {.1\height}},clip]{{\dataroot}HairPaper2023/figures/cvpr_drawing/start.jpg}
\subcaption{Initial strands (10\% length of the GT)}
\adjincludegraphics[width=1.\linewidth, trim={0 {.1\height} 0 {.1\height}},clip]{{\dataroot}HairPaper2023/figures/cvpr_drawing/intermediate.jpg}
\subcaption{Intermediate strands}
\adjincludegraphics[width=1.\linewidth, trim={0 {.1\height} 0 {.1\height}},clip]{{\dataroot}HairPaper2023/figures/cvpr_drawing/end.jpg}
\subcaption{Optimized strands}
\end{minipage}
\caption{``CVPR'' drawing by DR-based hair growing.
The entire sequence is available in the supplementary video.
The images were ray-traced by Blender Cycles after each letter was optimized independently.
}
\label{fig:cvpr_drawing}
\end{figure}

\begin{figure}[htb]\centering
\begin{minipage}[t]{0.45\linewidth}
\centering
\includegraphics[width=2.3cm]{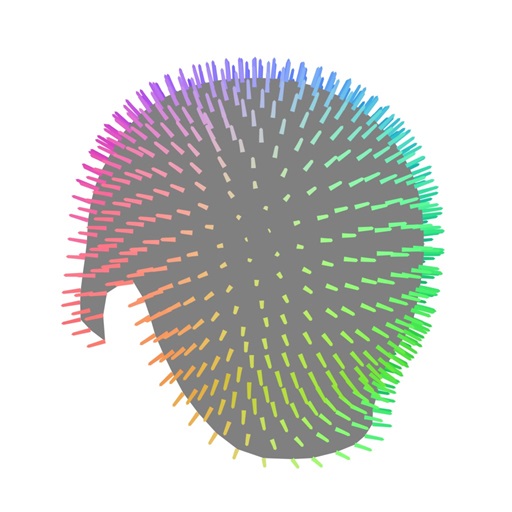}
\subcaption{Normals at scalp, $n_s(p_s)$}
\end{minipage}
\begin{minipage}[t]{0.45\linewidth}
\centering
\includegraphics[width=2.3cm]{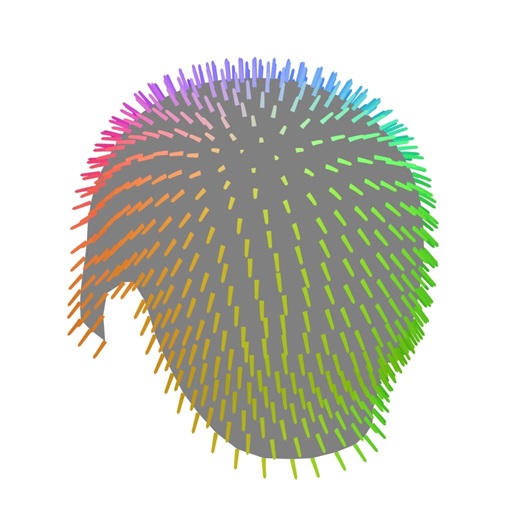}
\subcaption{Our $S(p_s)$}
\end{minipage}
\vspace*{-3mm}
\caption{Comparison of scalp normal and our $S(p_s)$.
(a) Normals at the scalp, $n_s(p_s)$. Growing directions at the side and back are different from real humans.
(b) Our hair orientation at scalp $S(p_s)$. Strands can grow more naturally at the side and back. }
\label{fig:scalp_dir}
\end{figure}

\subsection{Additional results on studio data}
To demonstrate our method's robustness on various real hairstyles, \figurename~\ref{fig:exp_lc3} and \figurename~\ref{fig:exp_lc4} show more results on studio data.
In addition to final child hair, guide hair is also shown for each subject.
The robustness of our framework against diverse natural hairstyles is proven.
\subsection{Additional physics simulation results}
In \figurename~\ref{fig:exp_physics_motion}, the physics simulations with head movement were compared in time series.
The full sequence is available in the supplementary video.
The natural behavior of our strands indicates that our method is capable of reconstructing simulation-ready hair strands.
\subsection{Additional results on USC-HairSalon \cite{hu2015database}}
To address robustness against artistic hairstyles, we conducted experiments using USC-HairSalon \cite{hu2015database}.
A total of 58 synthetic images were generated following the same procedure as Cem Yuksel's hair models as input.
\figurename~\ref{fig:exp_usc} illustrates that our approach adeptly reconstructs artistic hairstyles.

\section{Anti-aliasing validation}
\subsection{Comparison with existing AAs}
\label{sec:aa}
Our anti-aliasing (AA) for line segments is validated on a toy problem that grows a strand by DR.
The toy problem is to fit a minimum line segment with two vertices, one at the root and one at the tip, into the target image, where the silhouette of a long strand is depicted.
The line segment is initialized with the length in 20\% of the target strand.
Its root is fixed, and the tip position is optimized.
To validate gradient quality itself, we used a simple optimizer, stochastic gradient descent without momentum.
The learning rate was set to 1.0, and DR optimization with an L2 silhouette loss was performed in 25,000 iterations.
Per iteration, the line segment is converted to a triangle, as shown in Figure 5 of the main paper.
The triangle is rasterized in 128x128 pixels, and then each AA is applied.

In the experiments of the main paper, the width of our strand is set to 0.2 mm, which is often thinner than one pixel.
So, in this validation, we tested root thickness in 1.0, 0.8, and 0.6 pixels.
Thin width is prone to cause jumping pixels by the nature of rasterization.

\figurename~\ref{fig:aa_validation_graph} shows the quantitative comparison with nvdiffrast \cite{Laine2020diffrast} and splatting \cite{cole2021differentiable}, which are AAs for meshes.
Similar to ours, nvdiffrast is based on geometric AA, but its gradient generation is selective.
Splatting is another approach that propagates gradients through the weighted sum of neighbor pixels with differentiable screen space position interpolation.
Our AA generates gradients for all line edges and propagates them to vertex positions via pixel-to-edge distance.
Ours can reduce loss monotonically, while the other AAs show difficulty in handling tiny geometry.
The qualitative comparison is displayed in \figurename~\ref{fig:aa_validation_fig}.
Our AA generates a smooth gradient even with a very thin geometry, which leads to successful line segment alignment.
\subsection{``CVPR'' drawing by hair growing}
Although the proposed pipeline utilizes the AA for fine-tuning following initialization, it possesses sufficient capability for growth.
The depiction of the ``CVPR'' drawing with strands is presented in \figurename~\ref{fig:cvpr_drawing}.
Each GT letter is constructed using one, two, two, and three bundles for ``C'', ``V'', ``P'', and ``R'' respectively, wherein a bundle comprises 500 strands with 50 segments functioning as child hair, each capable of independent movement.
Target images were rendered from random views of GT hairs on a per-letter basis.
GT hairs were shortened to 10\% of their original length as starting values for optimization utilizing DR.
Each letter was optimized independently by Adam optimizer.
The loss comprised $L_{m}$, $L_{o}$, $R_{root}$, $R_{c}$, alongside regularizer for equalizing segment lengths.
For reparameterization, $k$NN with $k=10$ was performed for $\mathscr{N}$ at 50\% of the original length to address artifacts occurring at junctions.
The ``CVPR'' drawing demonstrates that our AA can optimize complex shapes effectively.

\section{Implementation details}
\subsection{Scalp fitting and hair region extraction}
\label{sec:scalp_fitting}
We describe the details of scalp fitting and hair region extraction, corresponding to \textbf{3.1. Initialization} of the main paper.
We project semantic segmentation \cite{liu2022cdgnet} onto a raw mesh from each view while extracting the hair region by vertex-wise voting.
Facial landmarks \cite{insightface} are also projected.
Subsequently, 3D correspondences between the raw mesh and the head template model are established, and similarity transform is estimated using Umeyama's method \cite{umeyama1991least}.
Non-rigid registration by deforming vertices is then carried out.
During the non-rigid registration process, as the raw mesh included hairs, but some scalp regions were not visible, we only considered regions other than the hair, such as the ears, face, and neck.
More specifically, we rendered depth images from each view and optimized the vertex positions with reparameterization \cite{Nicolet2021Large} via differentiable rendering to minimize an L1 depth loss within the facial area and an L2 3D landmark loss.
At this stage, the scalp area might extend beyond the hair region in the raw mesh.
To address this, we performed a post-process to push the scalp area into the raw mesh.
Based on the same non-rigid registration framework, an L1 silhouette loss between the head model's scalp area and the raw mesh's hair region is minimized with a regularization term to keep facial depth values.
The final scalp mesh is obtained from the scalp area of the head mesh through linear interpolation.
\subsection{Detailed description of the scalp boundary condition}
To clarify $S(p_s)$ in the Equation 2 of the main paper, \figurename~\ref{fig:scalp_dir} illustrates the contrast between the scalp normal $n_s(p_s)$ and $S(p_s)$.
Our $S(p_s)$ reflects the natural directions of scalp pores.
\subsection{Detailed description of the motivation for reparameterization}
We will explain our motivation of \textbf{3.2. Hierarchical Strand Optimization \textbar ~ Reparameterization} of the main paper in detail.
Our hair is represented as a set of thin geometries of less than one pixel, and the visibility in screen-space is stochastic due to the nature of hardware rasterization.
The outermost strands are not always rasterized, and the inner strands may show through.
Na\"ive DR optimization can easily collapse hairstyles since it makes only the visible division points of strands move at each iteration as w/o reparam. in \figurename~\ref{fig:exp_cg_straight} and \figurename~\ref{fig:exp_cg_curly}.
On the other hand, reparameterization, countermeasures for sparse gradients in geometry, has been studied in the context of meshes \cite{Nicolet2021Large}.
We, therefore, proposed the reparameterization for line segments, where each Laplacian element has the following regularizing effect:
$\mathscr{F}$: If a small part of a strand is visible, the whole strand moves smoothly based on the visible part;
$\mathscr{N}$: Always, even if a strand is not visible at all, the strand moves smoothly based on the other visible strands in the neighborhood.

\subsection{Hair mask for DR loss}
At \textbf{3.2. Hierarchical Strand Optimization \textbar ~ Guide/Child Hair Optimization} of the main paper, we generate hair masks for $L_{m}$ through the hair region of the raw mesh in a similar manner to NeuralStrands \cite{rosu2022neuralstrands}.
First, silhouettes are rendered with the hair region mesh onto each view.
Because a multi-view voting scheme estimates the hair region mesh, our silhouette extraction is more tolerant of severe failures than applying 2D silhouette extraction to the input images individually.
Then, the tri-map is made by erosion and dilation.
KNN Matting \cite{chen2013knn} with the tri-map and an input color image is finally applied to generate an alpha hair mask.

\subsection{Module-level performance measurement}
We report relative time spent on each module in \tablename~\ref{tab:time_module}.

\begin{table}[htb]\centering
\caption{Time consumption ratio per module}
\label{tab:time_module}
\begin{tabular}{lc}
\hline
Raw mesh reconstruction   & 4\%    \\
Scalp fitting             & 20\%   \\
3D Orientation estimation & 18\% \\
Strand initialization     & 5\%    \\
Hair mask generation      & 15\%   \\
Strand optimization by DR & 38\% \\ \hline
\end{tabular}
\end{table}

\end{document}